\documentclass[journal]{IEEEtran}

\usepackage{cite}
\usepackage{amsmath}
\usepackage{algorithmic}
\usepackage{array}
\usepackage{bm}
\usepackage{bbm}
\usepackage{amssymb}
\usepackage{amsthm}
\usepackage{mathrsfs}
\usepackage{empheq}	
\usepackage[mathcal]{euscript}
\usepackage{xcolor}
\usepackage{floatrow}

\newtheorem{theorem}{Theorem}
\newtheorem{lemma}{Lemma}

\newtheorem{proposition}{Proposition}
\newtheorem{property}{Property}

\newtheorem*{mcdiar}{McDiarmid's Inequality}

\renewcommand{\P}{\mathbb{P}}
\newcommand{\E}{\mathbb{E}}

\newcommand{\T}{^\top}

\DeclareMathOperator*{\argmax}{arg\,max}
\DeclareMathOperator*{\argmin}{arg\,min}
\newcommand*{\QED}{\hfill\ensuremath{\blacksquare}}
\newcommand*{\medcap}{\mathbin{\scalebox{1.5}{\ensuremath{\cap}}}}
\newcommand*{\medcup}{\mathbin{\scalebox{1.5}{\ensuremath{\cup}}}}%
\newcommand{\ropt}{{\sf R}^o}
\ifCLASSINFOpdf
   \usepackage[pdftex]{graphicx}
\else
\fi
\ifCLASSOPTIONcompsoc
  \usepackage[caption=false,font=normalsize,labelfont=sf,textfont=sf]{subfig}
\else
  \usepackage[caption=false,font=footnotesize]{subfig}
\fi

\allowdisplaybreaks

\begin{document}

\title{Learning from Heterogeneous Data Based on Social Interactions over Graphs}

\author{Virginia~Bordignon, Stefan~Vlaski,
        Vincenzo~Matta
        and~Ali~H.~Sayed%
\thanks{Virginia Bordignon and Ali H. Sayed are with the School of Engineering, EPFL, CH1015 Lausanne, Switzerland (e-mails: \{virginia.bordignon, ali.sayed\}@epfl.ch).}%
\thanks{Stefan Vlaski is with the Department of Electrical and Electronic Engineering, Imperial College London, London SW7 2BT, UK, (e-mail: s.vlaski@imperial.ac.uk). This work was performed while he was a post-doctoral researcher at the EPFL Adaptive Systems Laboratory.}
\thanks{Vincenzo Matta is with the Department of Information and Electrical Engineering and Applied Mathematics (DIEM), University of Salerno, 84084 Fisciano (SA), Italy, and also with the National Inter-University Consortium for Telecommunications (CNIT), Italy (e-mail: vmatta@unisa.it).}
\thanks{An early version with partial results from this paper was presented at ICASSP 2021 \cite{bordignon2020network}.
This work was supported in part by grant 205121-184999 from the Swiss National Science Foundation.}
}

\maketitle

\begin{abstract}
This work proposes a decentralized architecture, where individual agents aim at solving a classification problem while observing streaming features of different dimensions and arising from possibly different distributions. In the context of social learning, several useful strategies have been developed, which solve decision making problems through local cooperation across distributed agents and allow them to learn from streaming data. However, traditional social learning strategies rely on the fundamental assumption that each agent has significant prior knowledge of the underlying distribution of the observations. In this work we overcome this issue by introducing a machine learning framework that exploits social interactions over a graph, leading to a fully data-driven solution to the distributed classification problem. In the proposed social machine learning (SML) strategy, two phases are present: in the training phase, classifiers are independently trained to generate a belief over a set of hypotheses using a finite number of training samples; in the prediction phase, classifiers evaluate streaming unlabeled observations and share their instantaneous beliefs with neighboring classifiers. We show that the SML strategy enables the agents to learn consistently under this highly-heterogeneous setting and allows the network to continue learning even during the prediction phase when it is deciding on unlabeled samples. The prediction decisions are used to continually improve performance thereafter in a manner that is markedly different from most existing static classification schemes where, following training, the decisions on unlabeled data are not re-used to improve future performance.
\end{abstract}

\IEEEpeerreviewmaketitle

\section{Introduction}
\IEEEPARstart{S}{ocial} {learning strategies allow for the classification of unlabeled streaming observations by a network of agents~\cite{jadbabaie2012non,zhao2012learning,krishnamurthy2013social,nedic2017fast,salami2017social,lalitha2018social,matta2020interplay,bordignon2021adaptive}. These agents are heterogeneous in two main aspects: first, they may be observing different (possibly non-overlapping) sets of attributes of the same observed scene; second, their statistical models need not be the same, e.g., agents may be observing the same attribute from different perspectives, { which allows for a {\em distribution diversity} across agents.} In a collaborative solution, neighboring agents share beliefs about their observations and diffuse this information across the network to arrive at a conclusion on the true underlying class explaining the common observed scene.

Under a Bayesian social learning paradigm, the agents would need to build the full posterior distribution of the data by exploiting sophisticated dependencies across their individual models. From a design perspective, this approach is seldom viable since it would require exploiting mutual dependencies across a (possibly large) set of agents. From a behavioral perspective, it has been shown~\cite{molavi2018theory} that in many distributed systems, agents learn in a non-Bayesian way in the following sense. They first manage to construct their belief according to a {\em local} Bayesian update rule, and then these locally updated beliefs are shared among neighbors to guarantee percolation of information across the network. In this work, we focus accordingly on the non-Bayesian paradigm. Several non-Bayesian social learning strategies exist in the literature. A common feature is that they allow for correct asymptotic learning of the truth~\cite{jadbabaie2012non,zhao2012learning,nedic2017fast,lalitha2018social,bordignon2021adaptive}. This desirable learning property requires however prior knowledge of the true probability distributions characterizing agents' observations (usually referred to as \emph{likelihoods}), which are in general not available in real-world applications. In practice, these models are only approximate, oftentimes the result of a previous \emph{training} stage, where, from limited data, a parameterized model is learned. 

This issue is recognized in the works~\cite{hare2020non,hare2021general}, where the authors propose a framework for incorporating uncertainty into non-Bayesian social learning. While \cite{hare2020non} focuses only on sets of Gaussian distributions, their proposed strategy in \cite{hare2021general} broadens the approach, 
but requires nonetheless the selection of some suitable family of distributions that is amenable to closed-form mathematical manipulations, such as determination of a conjugate prior.
While relevant for numerically generated data, in practical applications there is generally little \emph{a priori} evidence regarding the structure of likelihoods, e.g., in the distributed classification of images or videos.

In this work, we propose the Social Machine Learning (SML) strategy: a decentralized algorithm for combining the outputs of a heterogeneous network of classifiers over space and time, based on the social learning algorithms proposed in \cite{lalitha2018social,matta2020interplay,bordignon2020adaptation,bordignon2021adaptive}. Our approach has the advantage of allowing the use of a fairly general class of distributions, which is relevant in practical machine learning tasks. The SML strategy consists of two independent phases: a \emph{training phase}, in which the classifiers are independently trained given a finite set of labeled data samples, and a \emph{prediction phase}, in which the trained classifiers are deployed in a local collaborative structure while observing streaming unlabeled samples. An early version of this strategy was presented in~\cite{bordignon2020network}. 

In our proposed strategy, agents (or classifiers) cooperate with neighbors to overcome \emph{local} spatial limitations. They also aggregate their \emph{instantaneous} opinions from streaming observations, strengthening their decision-making capabilities over time. These two aspects, i.e., information aggregation over space and time, are common topics of research in the fields of ensemble~\cite{dietterich2000ensemble} and multi-view learning~\cite{zhao2017multi}. 

Popular examples of ensemble approaches are bagging \cite{breiman1996bagging} and boosting \cite{freund1997decision}, in which classifiers combine weighted decisions across \emph{space}. However, such combination takes place in a {\em centralized} manner, namely, it is assumed that all agents communicate their decisions to a fusion center. This mechanism is fundamentally different from the {\em fully decentralized} setting addressed here, where only local cooperation between neighboring agents is permitted, and each individual agent is eventually able to learn the correct class. Moreover, both bagging and boosting methods do not address the streaming data case, i.e., they do not leverage the temporal quality of the online observations. 

In multi-view learning, multiple views of the same data are available, which are jointly used to improve generalization performance. Multi-view co-training approaches~\cite{blum1998combining} are notably suitable for semi-supervised learning, where a substantial number of unlabeled samples are available. In these approaches, distinct classifiers are trained on different views, and one classifier's predictions on new unlabeled examples are used to enlarge the labeled training set of the other. The procedure is repeated over the unlabeled samples, improving their accuracy over successive iterations. However, multi-view learning does not address the {\em distributed} and {\em streaming-data} aspects. 
Regarding the former aspect, in multi-view learning the classifiers are not spatially distributed or, if they are, it is simply assumed that they can share their beliefs without any constraints (i.e., as if they were co-located). Regarding the latter aspect, multi-view learning does not assume that streaming data are available for prediction.

The main contributions of this work are as follows. We propose a social machine learning strategy that inherits the following qualities from social learning: the ability to combine classifiers with different dimensions and statistical models, to adapt in view of changing real-time measurements, while providing asymptotic performance guarantees, in addition to continuous performance improvement during the prediction phase.
We show that with high probability, consistent learning occurs despite the imperfectly trained models. We also show that  poorly trained classifiers can leverage the networked setup to improve their performance, and that agents can use temporal memory to improve their generalization performance, in the form of classification accuracy. This property is distinct from existing static prediction phases for traditional classifiers, where classification decisions are instantaneous and do not take advantage from aggregation of increasing amounts of data that become available as time elapses.  We illustrate these results using an application to the MNIST~\cite{lecun2010mnist} dataset for image classification in a distributed setup. A comparison with AdaBoost shows how, contrary to AdaBoost, SML enables increased accuracy over time. We furthermore include a discussion in Appendix~\ref{ap:compare} comparing SML with the uncertain-likelihood approach introduced in~\cite{hare2021general}.

\emph{Notation:} We use boldface fonts to denote random variables, and normal fonts for their realizations. $\mathbb{E}$ and $\mathbb{P}$ denote expectation and probability operators, respectively. When necessary, suitable subscripts will be appended to denote the specific random variables the operators refer to. 
\section{Background}
\subsection{Inference Problem}\label{sec:ip}
We consider a network of $K$ agents or classifiers, indexed by $k\in\{1,2,\dots,K\}$, trying to identify the true state of nature $\gamma_0$ out of a binary set of hypotheses or classes $\Gamma=\{-1,+1\}$. The true state characterizes the \emph{scene} all agents are observing. To make a decision on the true state, each agent relies on the observation of streaming private data, which relate to the observed scene. Data are qualified as private due to the implicit assumption that raw observations cannot be shared among agents in order to, for example, minimize communication costs or preserve secrecy. 

More specifically, each agent $k$ observes at each instant $i$ the feature vector $\bm{h}_{k,i}\in\mathcal{H}_k$. The feature vectors are assumed to be independent and identically distributed (i.i.d.) over time. Moreover, the features $\bm{h}_{k,i}$ at agent $k$ given the state $\gamma_0$ form a sequence of i.i.d. random vectors distributed according to some conditional distribution (or likelihood):
\begin{equation}
\bm{h}_{k,i}\sim L_k(h|\gamma_0),\quad h\in\mathcal{H}_k, \gamma_0\in\Gamma.
\end{equation}
Notably, the model allows the features to be dependent across agents (i.e., over space). The feature set $\mathcal{H}_k$ is particular to agent $k$, allowing agents to be observing different features from the same scene; in particular, the dimension of ${\cal H}_k$ can be different across the agents. For example, an agent might be observing RGB video frames while another might be receiving infrared imagery taken both from the same street scene. Another source of heterogeneity is the likelihood model $L_k(h|\gamma)$, which differs across agents and reflects their individual perceptions. Within the previous street scene example, agents might observe frames captured under different, possibly non-overlapping, fields of view.

We can treat the true state of nature as a random variable $\bm{\gamma}_0$ and furthermore establish that the pair $(\bm{h}_{k,i},\bm{\gamma}_0)$ is distributed according to the following joint distribution:
\begin{equation}
(\bm{h}_{k,i}, \bm{\gamma}_0)  \sim p_k(h, \gamma) = L_k(h|\gamma)p_k(\gamma),\label{eq:testjoint}
\end{equation}
with $h\in\mathcal{H}_k,~\gamma\in\Gamma$, for every $i=1,2,\dots$ due to the i.i.d. assumption over time. Here, the notation $p_k(\gamma)$ corresponds to the prior distribution at agent $k$ for $\bm{\gamma}_0$ over the discrete set of hypotheses $\Gamma$. 

If the likelihood and prior distributions are perfectly known to agent $k$, different strategies can be deployed to enable truth learning. In a noncooperative framework, where each agent has enough information to solve the problem on their own, we can resort to the \emph{Bayes classifier}. Alternatively, to leverage the data spread across different agents of the network, a cooperative strategy can be used, such as one of the existing \emph{social learning} methods. We discuss each of the two strategies in more detail in the next paragraphs.

\subsection{Bayes Classifier}
When the Kullback-Leibler (KL) divergence~\cite{cover1999elements} between likelihoods $L_k(h|+1)$ and $L_k(h|-1)$ is strictly positive, we say that agent $k$ possesses {\em informative} likelihoods and can thus distinguish classes $+1$ and $-1$. Therefore, if the likelihood and prior distributions are known to agent $k$ and its likelihoods are informative, the agent can employ the Bayes classifier to solve the following maximum-a-posteriori (MAP) problem given an observed sequence of features $\{\bm{h}_{k,j} \}$ with $j=1,2,\dots,i$:
\begin{equation}
\bm{\gamma}^{\text{Bayes}}_{k,i} = \argmax_{\gamma\in\Gamma} p_k(\gamma| \bm{h}_{k,1}, \bm{h}_{k,2}, \dots, \bm{h}_{k,i}),\label{eq:bayespost}
\end{equation}
where $p_k(\gamma| \bm{h}_{k,1}, \bm{h}_{k,2}, \dots, \bm{h}_{k,i})$ indicates the posterior probability of the event $\{\bm{\gamma}=\gamma\}$ given the sequence $\{\bm{h}_{k,j} \}$ with $j=1,2,\dots,i$.
Using Bayes' rule and the property of conditional independence of the features over time, and taking the logarithm of the posterior probability, we obtain
\begin{align}
&\log p_k(\gamma|\bm{h}_{k,1},\bm{h}_{k,2},\dots, \bm{h}_{k,i})\nonumber\\&= \sum_{j=1}^i\log L_{k}(\bm{h}_{k,j}|\gamma)+\log p_{k}(\gamma)-\log p_k(\bm{h}_{k,1}, \dots, \bm{h}_{k,i}),\label{eq:bayeslikeli}
\end{align}
where the joint distribution over the feature vectors, namely $p_k(\bm{h}_{k,1}, \dots, \bm{h}_{k,i})$, does not depend on $\gamma$. For binary labels, solving \eqref{eq:bayespost} is equivalent {to producing a binary decision} $\bm{\gamma}_{k,i}^{\sf Bayes}$ using the following test:
\begin{equation}
\sum_{j=1}^i	\log  \frac{L_k(\bm{h}_{k,j}|+1)}{L_k(\bm{h}_{k,j}|-1)}+	\log  \frac{p_k(+1)}{p_k(-1)} \mathop{\gtrless}_{-1}^{+1} 0. \label{eq:bayesineq}
\end{equation}
Proposition~\ref{prop:bayes} characterizes the consistency of Bayes classifier, also known as recursive Bayesian forecasting given streaming observations. It is a known result~\cite{epstein2010non} and is therefore stated without proof.
\begin{proposition}[\textbf{Bayes Classifier for a Sequence of Observations}]\label{prop:bayes} Assume that $p_k(\gamma)>0$ for $\gamma \in \Gamma$, and that the KL divergences between likelihoods satisfy:
	\begin{align}
		&0<{\sf D}_{\sf KL}\left[L_k(+1)||L_k(-1) \right]< \infty\label{eq:kl1},\\
		&0<{\sf D}_{\sf KL}\left[L_k(-1)||L_k(+1) \right]< \infty\label{eq:kl2},
	\end{align}
where ${\sf D}_{\sf KL}\left[L_k(\gamma)||L_k(\gamma') \right]$ denotes the KL divergence between likelihoods $L_k(h|\gamma)$ and $L_k(h|\gamma')$.
Then the Bayes classifier $\bm{\gamma}_{k,i}^{\sf Bayes}$ learns the truth with probability 1 as $i$ goes to infinity.\QED
\end{proposition}
While this result ensures consistent learning, it requires each individual agent to have informative likelihoods and thus to be able to distinguish both hypotheses. This restriction motivates the pursuit of \emph{collaborative} social learning schemes, where agents exchange information to resolve ambiguities arising from incomplete information.

\subsection{Social Learning}\label{sec:SL}
In a multi-agent setup, a network of $K$ agents is modeled as a strongly-connected graph, i.e., where there is a path in both directions linking any two agents and at least one self-loop (a diagram can be seen in Fig.~\ref{fig:diag_net}). The graph possesses a left-stochastic combination matrix $A$, whose elements $a_{\ell k}$ satisfy:
\begin{equation}
\sum_{\ell=1}^Ka_{\ell k}=1,\quad a_{\ell k}\geq 0, \quad a_{\ell k}= 0 \text{ if }\ell \notin \mathcal{N}_k,
\end{equation}
where $\mathcal{N}_k$ indicates the set of neighbors of $k$ (including $k$ itself). Each $a_{\ell k}$ is a combination weight scaling information arriving at agent $k$ from agent $\ell$. From the strong-connectivity assumption, $A$ is a primitive matrix and we can therefore define its Perron eigenvector $\pi$ as~\cite{sayed2014adaptation}:
\begin{equation}
A\pi = \pi, \quad \sum_{k=1}^K\pi_{k}=1,\quad \pi_{ k}>0 \text{ for } k=1,2,\dots,K.\label{eq:perron}
\end{equation}
If the $K-$agent network is collectively observing streaming features $\{\bm{h}_{1,i},\bm{h}_{2,i},\dots, \bm{h}_{K,i}\}$, it is beneficial for agents to cooperate in order to solve the inference problem.  One intuitive benefit of cooperation comes from the fact that the network is observing $K$ times more data than each individual alone. Another fundamental benefit is that network cooperation can enable successful learning even when the decision problem is {\em not identifiable} for the individual agents, but is {\em globally} identifiable at the network level. Among the existing cooperative protocols, decentralized processing is particularly appealing for not only improving learning performance, but also for increasing robustness and ensuring data privacy~\cite{sayed2014adaptation}. Moreover, in a decentralized setup, agents share processed information, i.e., they do not share raw data, with their immediate neighbors. 

\begin{figure}[t]
	\centering
	\includegraphics[width=\textwidth]{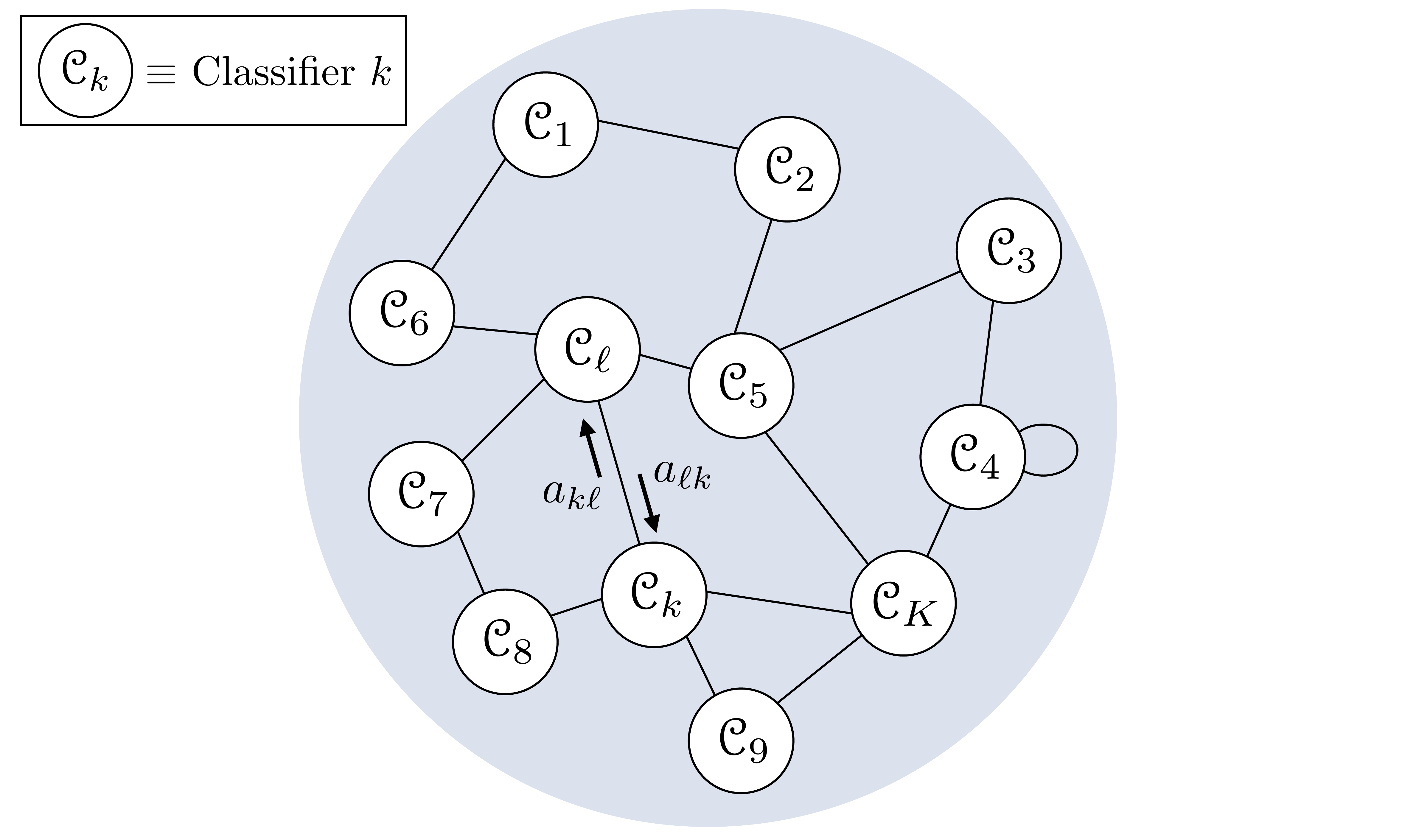}
	\caption{Diagram of the network of classifiers.}\label{fig:diag_net}
\end{figure}

Inspired by the advantages of decentralized processing, a large family of non-Bayesian social learning works has been proposed with the purpose of solving the aforementioned inference problem~\cite{jadbabaie2012non,zhao2012learning,nedic2017fast,lalitha2018social,salami2017social, bordignon2021adaptive}.  Resorting to different update and combination mechanisms, many of them have the feature of yielding asymptotic truth learning~\cite{jadbabaie2012non,zhao2012learning,nedic2017fast,salami2017social,lalitha2018social}. In the following development, we detail two of these strategies.

\paragraph{Traditional Social Learning}
We first consider the social learning (SL) strategy discussed in~\cite{molavi2018theory,nedic2017fast,lalitha2018social,matta2020interplay}. At every instant $i$, each agent $k$ updates its \emph{belief} (opinion) $\bm{\varphi}_{k,i}(\gamma)$ given its private observations and its neighbors' beliefs. The belief vector $\bm{\varphi}_{k,i}$ is a probability mass function over the set of hypotheses $\Gamma$, where each element $\bm{\varphi}_{k,i}(\gamma)$ represents the confidence that hypothesis $\gamma$ corresponds to the true state.

The belief is updated according to a two-step protocol:
\begin{align}
\bm{\psi}_{k,i}(\gamma)&=\frac{\bm{\varphi}_{k,i-1}(\gamma)L_k(\bm{h}_{k,i}|\gamma)}{\sum_{\gamma'\in\Gamma}\bm{\varphi}_{k,i-1}(\gamma')L_k(\bm{h}_{k,i}|\gamma')},\label{eq:bayest}\\
\bm{\varphi}_{k,i}(\gamma)&=\frac{\exp\left\{\sum_{\ell\in\mathcal{N}_k}a_{\ell k}\log \bm{\psi}_{\ell,i}(\gamma)\right\}}{\sum_{\gamma'\in\Gamma}\exp\left\{\sum_{\ell\in\mathcal{N}_k}a_{\ell k}\log \bm{\psi}_{\ell,i}(\gamma')\right\}},\label{eq:combt}
\end{align}
where in the first step (Eq.~\eqref{eq:bayest}) agent $k$ updates its \emph{intermediate belief} $\bm{\psi}_{k,i}$ using the observed feature vector $\bm{h}_{k,i}$. Then in the second step (Eq.~\eqref{eq:combt}), agents share their intermediate beliefs with neighboring agents and update their beliefs using a geometric averaging rule. 

An equivalent linear way of representing \eqref{eq:bayest} and \eqref{eq:combt} is in the form of the \emph{diffusion strategy}~\cite{sayed2014adaptation,sayed2014adaptive}:
\begin{align}
\bm{\eta}_{k,i}& =  \bm{\lambda}_{k,i-1} +c_k(\bm{h}_{k,i}),\label{eq:diffstatt1}\\
\bm{\lambda}_{k,i}&=\sum_{\ell\in\mathcal{N}_k} a_{\ell k}\bm{\eta}_{\ell,i},
\label{eq:diffstatt}
\end{align}
in terms of the following scalar quantities:
\begin{align}
&\bm{\lambda}_{k,i}\triangleq\log \frac{\bm{\varphi}_{k,i}(+1)}{\bm{\varphi}_{k,i}(-1)},\quad \bm{\eta}_{k,i}\triangleq\log \frac{\bm{\psi}_{k,i}(+1)}{\bm{\psi}_{k,i}(-1)},\label{eq:def_lc1}\\
 & c_k(\bm{h}_{k,i})\triangleq \log \frac{L_k(\bm{h}_{k,i}|+1)}{L_k(\bm{h}_{k,i}|-1)}.\label{eq:def_lc}
\end{align}
Equations \eqref{eq:diffstatt1} and \eqref{eq:diffstatt} can be joined into a single equation as:
\begin{equation}
\bm{\lambda}_{k,i}=\sum_{\ell\in\mathcal{N}_k}a_{\ell k}\Big(\bm{\lambda}_{\ell,i-1}+  c_\ell(\bm{h}_{\ell,i})\Big).\label{eq:SLrecursion}
\end{equation}
The representation in~\eqref{eq:SLrecursion} highlights that, in the decision variable $\bm{\lambda}_{k,i}$,  agent $k$ aggregates past (in the form of $\bm{\lambda}_{\ell,i-1}$) and present (in the form of $c_{\ell}(\bm{h}_{\ell,i})$) information relative to its neighbors $\ell\in\mathcal{N}_k$.

Previous works~\cite{lalitha2018social,nedic2017fast,matta2020interplay,molavi2018theory} show that, under the strategy in \eqref{eq:bayest} and \eqref{eq:combt}, the belief component $\bm{\varphi}_{k,i}(\gamma_0)$ converges almost surely to 1 as $i\rightarrow \infty$, i.e., the belief is maximized at the true hypothesis. Specifically, by developing the recursion in \eqref{eq:SLrecursion} it is possible to show that, as $i\rightarrow\infty$, the belief function is maximized at the true hypothesis, provided that a weighted combination (through the Perron eigenvector weights) of the detection statistics $c_{\ell}(\bm{h}_{\ell,i})$ has positive expectation under hypothesis $+1$ and negative expectation under $-1$, namely,\footnote{
		The sufficient condition for consistent learning in \eqref{eq:consist} can be reached by following similar arguments as in~\cite[Appendix A]{matta2020interplay}. Developing the recursion in \eqref{eq:SLrecursion} and dividing by $i$, we can conclude that 
		\begin{equation}
		\frac{1}{i}\bm{\lambda}_{k,i}\stackrel{\textnormal{a.s.}}{\longrightarrow}\sum_{\ell=1}^K\pi_{\ell}\E_{L_\ell(\gamma_0)}c_{\ell}(\bm{h}_{\ell,i}).
		\end{equation}
		If $\sum_{\ell=1}^K\pi_{\ell}\E_{L_\ell(\gamma_0)}c_{\ell}(\bm{h}_{\ell,i})>0$, then $\bm{\lambda}_{k,i}$ goes to $+\infty$, and thus agents decide for class $+1$. Otherwise if $\sum_{\ell=1}^K\pi_{\ell}\E_{L_\ell(\gamma_0)}c_{\ell}(\bm{h}_{\ell,i})<0$, then $\bm{\lambda}_{k,i}$ goes to $-\infty$, and thus agents decide for class $-1$ .}
\begin{equation}
\boxed{\sum_{\ell=1}^{K}\pi_\ell \E_{L_\ell(+1)} c_\ell(\bm{h}_{\ell,i})> 0,~~ \sum_{\ell=1}^{K}\pi_\ell \E_{L_\ell(-1)}c_\ell(\bm{h}_{\ell,i})< 0} \label{eq:consist}
\end{equation}
where $\E_{L_\ell(\gamma)}$ indicates that the expectation is computed with respect to the distribution $L_\ell(h|\gamma)$. We remark that the condition for consistency in \eqref{eq:consist} would apply to general detection statistics $c_{\ell}(\cdot)$, and not only to log-likelihood ratios as in \eqref{eq:def_lc}. For example, detection statistics different from \eqref{eq:def_lc} may arise because the agents compute {\em mismatched} log-likelihood ratios due to imperfect knowledge.
This observation is particularly relevant in our work since, when we will examine the social machine learning setting (where the likelihoods are unknown) we will need to work with general detection statistics learned from a training set. On the other hand, for the specific case where the likelihoods are known and \eqref{eq:def_lc} is employed, the conditions in \eqref{eq:consist} are satisfied whenever the network satisfies the \emph{global identifiability} assumption~\cite{lalitha2018social}, i.e., at least one agent in the network is able to distinguish the hypotheses. In this case, for at least one agent $k$, it follows that
\begin{align}
\E_{L_k(+1)}c_k(\bm{h}_{k,i})&= {\sf D}_{\sf KL}[L_k(+1)||L_k(-1)]>0\label{eq:ident}\\
\E_{L_k(-1)}c_k(\bm{h}_{k,i})&= -{\sf D}_{\sf KL}[L_k(-1)||L_k(+1)]<0.\label{eq:ident1}
\end{align}
From the positivity of the Perron eigenvector $\pi$, \eqref{eq:ident} and \eqref{eq:ident1} imply that the conditions in \eqref{eq:consist} are satisfied.

Therefore, the strategy in \eqref{eq:bayest} and \eqref{eq:combt} allows agents to learn the truth asymptotically, as $i$ tends to infinity, with probability 1~\cite{nedic2017fast,lalitha2018social}. Almost sure consistent learning is a desirable feature in any inference algorithm. This remarkable accuracy in learning the truth comes however at the expense of a reduced ability by the algorithm to \emph{adapt} to drifting non-stationary conditions, e.g., to changes in the true state $\gamma_0$, to missing observations, and to changes in the statistical characteristics of the data. As discussed in~\cite{bordignon2021adaptive}, agents behave stubbornly in face of environment changes, displaying an unreasonably large time to adapt to new conditions. To address non-stationary applications, we introduce a second SL algorithm.

\paragraph{Adaptive Social Learning}
In a real-time application, we expect the environment conditions to change with time, and the learning strategy should be able to track the drifting conditions within a reasonable response time.  In~\cite{bordignon2020adaptation,bordignon2021adaptive}, an {\em adaptive social learning} strategy was proposed to overcome the lack of adaptation in traditional social learning. 

In one of the formulations proposed in~\cite{bordignon2021adaptive}, the first step of the update rule in \eqref{eq:bayes} is replaced by an \emph{adaptive} update as follows:
\begin{align}
 \bm{\psi}_{k,i}(\gamma)&=\frac{\bm{\varphi}_{k,i-1}^{1-\delta}(\gamma)L_k(\bm{h}_{k,i}|\gamma)}{\sum_{\gamma'\in\Gamma}\bm{\varphi}_{k,i-1}^{1-\delta}(\gamma')L_k(\bm{h}_{k,i}|\gamma')},\label{eq:bayes}
\end{align}
where $0<\delta\ll 1$ is a small step-size (or learning) parameter\footnote{In~\cite{bordignon2021adaptive}, it is shown that the update in \eqref{eq:bayes} yields a learning performance that is similar to the alternative update:
		\begin{equation}
\bm{\psi}_{k,i}(\gamma)=\frac{\bm{\varphi}_{k,i-1}^{1-\delta}(\gamma)L_k^\delta(\bm{h}_{k,i}|\gamma)}{\sum_{\gamma'\in\Gamma}\bm{\varphi}_{k,i-1}^{1-\delta}(\gamma')L_k^{\delta}(\bm{h}_{k,i}|\gamma')},
		\end{equation} 
in which $\delta$ also weights the new information contained in $L_k(\bm{h}_{k,i}|\gamma)$.}. The combination step, in which the intermediate beliefs are shared across neighbors, is the same as the one in~\eqref{eq:combt}. The introduction of a step-size to the local update in \eqref{eq:bayes} infuses the algorithm with the ability to adapt in face of non-stationary conditions with an \emph{adaptation time} that scales as $\mathcal{O}(1/\delta)$~\cite{bordignon2021adaptive}. In the limit case, when $\delta \rightarrow 0$, we recover the Bayesian update in \eqref{eq:bayest}.
 
Similarly to \eqref{eq:diffstatt1}--\eqref{eq:diffstatt}, we can represent \eqref{eq:combt} and \eqref{eq:bayes} in the form of an \emph{adaptive diffusion strategy}~\cite{matta2018estimation}:
\begin{align}
\bm{\eta}_{k,i}&=(1-\delta) \bm{\lambda}_{k,i-1}+c_k(\bm{h}_{k,i}),
\\
\bm{\lambda}_{k,i}&=\sum_{\ell\in\mathcal{N}_k}a_{\ell k} \bm{\eta}_{\ell,i},
\label{eq:diffstat}
\end{align}
which yields:
\begin{equation}
\bm{\lambda}_{k,i}=\sum_{\ell\in\mathcal{N}_k}a_{\ell k}\Big((1-\delta)\bm{\lambda}_{\ell,i-1}+  c_\ell(\bm{h}_{\ell,i})\Big).\label{eq:ASLrecursion}
\end{equation}
From \eqref{eq:ASLrecursion}, we see clearly that the step-size $\delta$ attenuates the influence of past data, embodied by $\bm{\lambda}_{\ell,i-1}$. As long as $\delta$ is strictly greater than zero and smaller than one, the recursion in \eqref{eq:ASLrecursion} can be shown to be stable, i.e., $\bm{\lambda}_{k,i}$ does not degenerate to $\pm \infty$ as $i\rightarrow\infty$. This non-degenerate behavior is the reason why the adaptive social learning algorithm can quickly recover from a previous state when faced with changes in the environment~\cite{bordignon2021adaptive}. 

The price for this improved adaptation is reflected on the learning accuracy. In contrast with the almost sure convergence found in traditional social learning, consistent learning now occurs asymptotically (as $i\rightarrow \infty$) with high probability in the regime of small step-sizes (as $\delta\rightarrow 0$)~\cite{bordignon2020adaptation,bordignon2021adaptive}. The same sufficient condition for attaining consistent truth learning enunciated in \eqref{eq:consist}, applies for the adaptive social learning algorithm as well, even for general detection statistics $c_{\ell}(\cdot)$~\cite{matta2018estimation}.

Similarly to most social learning algorithms in the literature, the family of likelihoods $L_{k}(h|\gamma)$ with $\gamma\in\Gamma$ is assumed to be {\em perfectly known} to each agent $k$. As we can see, in both of the aforementioned social learning strategies, one important statistic diffused across agents and over time is the log-ratio of likelihoods, $c_{\ell}(\cdot)$ as in \eqref{eq:def_lc}, which is classically employed as the basic building block to design other types of distributed detection strategies~\cite{viswanathan1997distributed,blum1997distributed}. In real-world applications, these likelihood models are generally unavailable. Instead, they are obtained as the result of a prior training step in which (parameterized) models are trained using a finite set of data examples.

In view of this practical limitation of social learning, we propose in this work a two-phase learning strategy, which we refer to as Social Machine Learning (SML). In this strategy, the likelihood models are assumed to be \emph{unknown}.

\section{Social Machine Learning}
The SML strategy is designed as a two-step approach. In the \emph{training phase}, the classifiers are trained individually given private finite datasets. In the \emph{prediction phase}, classifiers are deployed in a cooperative social learning structure. In Fig.~\ref{fig:diag}, we show a diagram depicting the SML approach. These distinct \emph{learning phases} are detailed in the following sections.

To avoid confusion, random variables related to the training phase are topped with a symbol $\sim$. Furthermore, data samples pertaining to the training datasets are indexed by $n$, whereas, in the prediction phase, they are indexed by $i$. For example, $\widetilde{\bm{h}}_{k,n}\in\mathcal{H}_k$ represents the $n-$th feature vector available in the training dataset of agent $k$, whereas  $\bm{h}_{k,i}\in\mathcal{H}_k$ represents the feature vector observed by agent $k$ at instant $i$ during the prediction phase. We assume that the random variables are independent between different learning phases.

\begin{figure}[t]
	\centering
	\includegraphics[width=\textwidth]{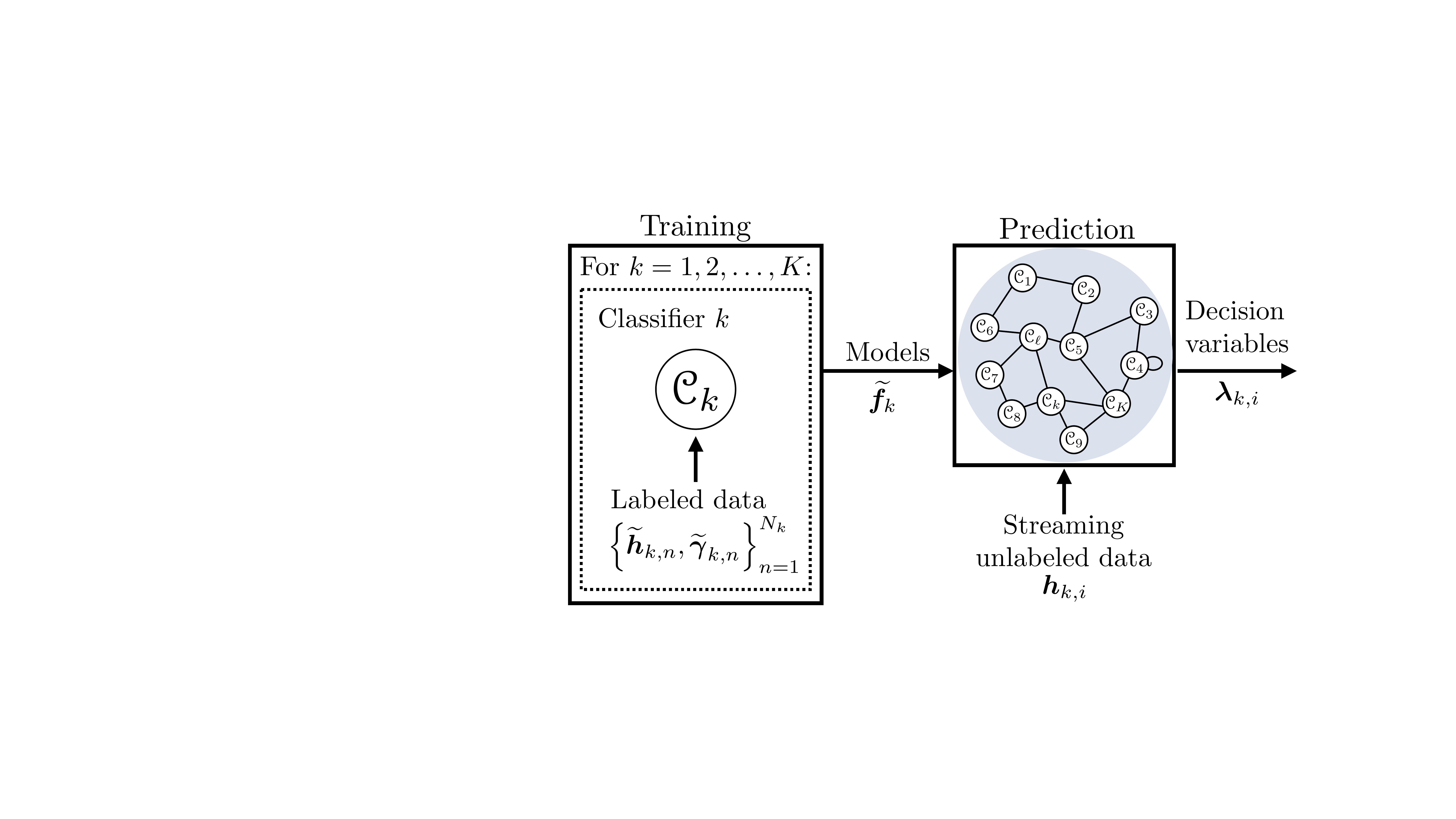}
	\caption{Social Machine Learning (SML) diagram.}\label{fig:diag}
\end{figure}

\subsection{Training Phase} 
During training, each agent $k$ has access to $N_k$ examples consisting of pairs $\{\widetilde{\bm{h}}_{k,n},\widetilde{\bm{\gamma}}_{k,n}\}_{n=1}^{N_k}$. We assume that the training set is balanced so that both classes are sufficiently explored, namely, we assume that, {\em during training}, labels $\widetilde{\bm{\gamma}}_{k,n}$ are uniformly distributed over $\Gamma=\{-1,+1\}$ for all agents.
This is a standard technical assumption that will be useful to obtain readable bounds for the social machine learning strategy. 
However, we remark that the assumption of a balanced dataset during the training phase does not impose any constraint on the behavior of the true hypothesis during the {\em prediction} phase. In particular, the data observed during prediction are all coming from a certain true state of nature $\gamma_0$, and we will establish that the SML strategy achieves vanishing error regardless of the particular hypothesis being in force.

The pair $(\widetilde{\bm{h}}_{k,n},\widetilde{\bm{\gamma}}_{k,n})$ is distributed according to the joint distribution:
\begin{equation}
	(\widetilde{\bm{h}}_{k,n},\widetilde{\bm{\gamma}}_{k,n})\sim \widetilde{p}_k(h,\gamma)=L_k(h|\gamma)\widetilde{p}_k(\gamma),\label{eq:jointd}
	\end{equation}
where $h\in\mathcal{H}_k,~\gamma\in\Gamma$ and $\widetilde{p}_k(\gamma)=1/2$ for $\gamma\in\Gamma$. Note that, given a label $\gamma$, the corresponding feature vector $\widetilde{\bm{h}}_{k,n}$ is distributed according to 
\begin{equation}
\widetilde{\bm{h}}_{k,n}\sim L_k(h|\gamma), \quad h\in\mathcal{H}_k, \gamma \in\Gamma.
\end{equation}
Using these training samples, we wish to deploy a fully data-driven solution inspired by the social learning algorithms presented in Sec.~\ref{sec:SL}. To accomplish this, we first need to approximate the key unknown function $c_k$ used in \eqref{eq:SLrecursion} and \eqref{eq:ASLrecursion} by a quantity resulting from some statistical learning method. The choice of method and the characteristics of the problem at hand, e.g., training samples and class of models considered, will heavily determine the quality of the resulting approximation. These decisive factors and their impact will be examined later in this work.

Let us delve into the details of our training setup. First, note that, under the assumption of uniform priors during training, and using Bayes' rule, we can write:
\begin{equation}
c_k(h)=\log\frac{L_k(h|+1)}{L_k(h|-1)}=\log\frac{\widetilde{p}_k(+1|h)}{\widetilde{p}_k(-1|h)},\label{eq:postratio}
\end{equation}
where $\widetilde{p}_k(\gamma|h)$ represents the posterior probability of $\{ \widetilde{\bm{\gamma}}_{k,n}=\gamma\}$ given $\{\widetilde{\bm{h}}_{k,n}=h\}$ using the joint model seen in \eqref{eq:jointd}.
The significance of the log-ratio of posterior probabilities on the RHS of \eqref{eq:postratio} can be interpreted in an intuitive manner: the log-ratio is positive whenever class $+1$ is more likely to be the true state of nature given the observation of $h$ and negative when class $-1$ is more likely to explain the same data evidence. It is therefore reasonable that we seek an approximation for the log-ratio of posteriors in \eqref{eq:postratio} during the training phase.
	
One relevant machine learning paradigm to approximate the posterior distribution is the {\em discriminative} paradigm (which includes, e.g., logistic regression and neural networks), where the output of the classifier is in the form of {\em approximate} posterior probabilities for each class, namely, $\widehat{p}_k(+1|h)$ and $\widehat{p}_k(-1|h)$. In order to illustrate this paradigm, it is convenient to introduce the \emph{logit} statistic:
\begin{align}
\log\frac{\widehat{p}_k(+1|h)}{\widehat{p}_k(-1|h)}=\log\frac{\widehat{p}_k(+1|h)}{1-\widehat{p}_k(+1|h)}\triangleq f_k(h),\label{eq:logit}
\end{align}
where the function $f_k$ can be chosen from an admissible class $\mathcal{F}_k$, namely,
\begin{equation}
 f_k\in \mathcal{F}_{k}:\mathcal{H}_k\mapsto \mathbb{R}.
\end{equation}
The choice of the class $\mathcal{F}_k$ depends on the choice of classifier. For example, in linear logistic regression with $h\in\mathbb{R}^M$, $\mathcal{F}_{k}$ is parameterized by a vector $w\in\mathbb{R}^M$, and we have the linear logit function~\cite{mohri2018foundations}:
\begin{equation}
f_k(h;w)=w\T h.\label{eq:logregression}
\end{equation}
Another example is to consider MultiLayer Perceptrons (MLPs) with $L$ hidden layers and a softmax output layer, whose weight matrices are given by $\{W_{\ell}\}$ over layers $\ell=1,2,\dots,L$. In the binary classification case, the network outputs two approximate posterior quantities~\cite{bishop2006pattern}, namely $\widehat{p}_k(+1|h; W)$ and $\widehat{p}_k(-1|h;W)$, where $W$ represents the parameterization of the classifier w.r.t. matrices $\{W_{\ell}\}$. In this case, the logit function is given by the expression:
\begin{equation}
f_k(h;W)=\log\frac{\widehat{p}_k(+1|h; W)}{\widehat{p}_k(-1|h; W)},\label{eq:mlpcase}
\end{equation}
where the class of functions $\mathcal{F}_k$ is parameterized by matrices $\{W_{\ell}\}$ in a nonlinear manner. Note that the forthcoming analysis does not assume a specific model for the logit function, and applies instead to general classes $\mathcal{F}_k$.

The logit functions $f_k$ are trained by each classifier $k=1,2,\dots,K$ by finding the function $f_k$ within $\mathcal{F}_k$ that minimizes a suitable risk function $R_k(f_k)$. For example, in the already mentioned logistic regression and MLP cases, the training process results in optimal parameters $w$ and $W_{\ell}$ for $\ell=1,2,\dots,L$, respectively.

One common risk function adopted in binary classification is the logistic risk:
\begin{equation}
R_k(f_k)=\E_{\tilde{h}_k, \tilde{\gamma}_k}
\log
\left(1+e^{- \widetilde{\bm{\gamma}}_{k,n}f_k(\widetilde{\bm{h}}_{k,n})}\right),
\label{eq:exprisk}
\end{equation}
where $\E_{\tilde{h}_k, \tilde{\gamma}_k}$ corresponds to the expectation computed under the (unknown) joint distribution $\widetilde{p}_k(h,\gamma)$ seen in \eqref{eq:jointd}.
We remark that the logistic risk can be used either in association with the linear model in \eqref{eq:logregression} or with more complex structures such as neural networks with softmax output layers. The logistic risk can be shown to be equivalent in the binary case to the cross-entropy risk function~\cite{murphy2012machine}. 

	We define the \emph{target risk} at every agent and the weighted network average according to:
	\begin{equation}
	{\sf R}^o_k\triangleq \inf_{f_k\in\mathcal{F}_k} R_k(f_k), \qquad{\sf R}^o\triangleq \sum_{k =1 }^{K}\pi_k{\sf R}^o_k.\label{eq:targetrisk}
	\end{equation}
Unfortunately, in practice the expectation in \eqref{eq:exprisk} cannot be computed since the underlying feature/label distribution is unknown. 
The agents rely instead on a finite set of training samples to minimize an {\em empirical} risk:
\begin{equation}
\widetilde{\bm{f}}_{k}\triangleq \argmin_{f_k\in\mathcal{F}_k}\widetilde{\bm{R}}_{k}(f_k),\label{eq:emprisk}
\end{equation}
given by
\begin{equation}
\widetilde{\bm{R}}_{k}(f_k)=\frac{1}{N_k}\sum_{n=1}^{N_k}\log
\left(1+e^{- \widetilde{\bm{\gamma}}_{k,n}f_k(\widetilde{\bm{h}}_{k,n})}\right),\label{eq:emprisk1}
\end{equation}
which is computed over the training set. The resulting function $\widetilde{\bm{f}}_{k}$ can then be used by the agents to approximate the logit statistic in \eqref{eq:logit}. For future use, we also define the {\em network average} for the expected risk and the empirical risk expressions:
\begin{align}
R(f) \triangleq \sum_{k=1}^{K}\pi_k R_k(f_k),\qquad \widetilde{\bm{R}}(f) \triangleq \sum_{k=1}^{K}\pi_k \widetilde{\bm{R}}_k(f_k),\label{eq:netrisk}
\end{align}
where the argument $f$ represents the dependence of the risk expressions on the collection of functions $\{f_k\}$, i.e., $R(f)=R(f_1,f_2,\dots,f_K)$. This concise notation will be used whenever we are dealing with network-averaged quantities.

We will detail in the next section how the trained models can be deployed in the prediction phase, when agents are faced with streaming unlabeled feature vectors. 

\subsection{Prediction Phase}
In the prediction phase, agents find themselves in the setup described in Sec.~\ref{sec:ip}. They aim at solving the inference problem of determining the true state $\gamma_0\in\Gamma$, given streaming \emph{unlabeled} private features $\bm{h}_{k,i}$, $i=1,2,\dots$. The difference now is that they are equipped with the  trained models $\{\widetilde{\bm{f}}_{k}\}$, which are constructed so as to provide a reasonable approximation for the log-ratio of posterior probabilities (see Fig.~\ref{fig:diag} for an illustrative diagram of this process).

During prediction, agents deploy one of the SL algorithms enunciated in Sec.~\ref{sec:SL} using the following approximation for the function $c_k$:
\begin{equation}
\widetilde{\bm{c}}_{k}(h)=\widetilde{\bm{f}}_k(h)-	\widetilde{\bm{\mu}}_{k}(\widetilde{\bm{f}}_k),
\label{eq:choiceofc}
\end{equation}
where the second term on the RHS of \eqref{eq:choiceofc} is called the \emph{empirical training mean} and is defined for any function $f_k\in\mathcal{F}_k$ as:
\begin{equation}
\widetilde{\bm{\mu}}_{k}(f_k)=\frac{1}{N_k}\sum_{n=1}^{N_k}f_k(\widetilde{\bm{h}}_{k,n}),\label{eq:emptrainmean}
\end{equation}
i.e., it is defined as the average of function $f_k$ over the training samples. Discounting the empirical training mean in \eqref{eq:choiceofc} prevents the logit statistic from being biased towards one class or another. This is relevant considering that the decision of each agent is taken according to the rule 
\begin{equation}
\bm{\gamma}_{k,i}^{\sf SML}\triangleq\text{sign}(\bm{\lambda}_{k,i}),\label{eq:decsml}
\end{equation}
where $\text{sign}(x)=+1$, if $x\geq 0$ and $\text{sign}(x)=-1$ otherwise, i.e., the decision threshold is zero. Note that $\widetilde{\bm{c}}_{k}$ is a random function, whose randomness stems from the training phase.

Next, we illustrate how the debiasing operation used in \eqref{eq:choiceofc} helps preventing biased decisions, but first let us define for any function $f_k\in\mathcal{F}_k$ the following conditional means:
\begin{align}
\mu^+_k(f_k)\triangleq \E_{L_k(+1)}f_{k}(\bm{h}_{k,i}),\quad
\mu^-_k(f_k)\triangleq \E_{L_k(-1)}f_{k}(\bm{h}_{k,i}).\label{eq:mupm}
\end{align}
Assume that $f_k$ is fixed, i.e., $\widetilde{\bm{c}}_k(h)=f_k(h)-\widetilde{\bm{\mu}}_k(f_k)$, and that $N_k$ is sufficiently large. Then, the empirical mean $\widetilde{\bm{\mu}}_k(f_k)$ approximates the expected value of $f_k(\widetilde{\bm{h}}_{k,n})$ in the training phase, namely, $[\mu^+_k(f_k)+\mu^-_k(f_k)]/2$ (see Eq.~\eqref{eq:avemean} in Appendix~\ref{ap:theo1}). In this case, function $\widetilde{\bm{c}}_k$ is deterministic, and we can write: 
\begin{equation}
\widetilde{c}_k(h)=f_k(h)-\frac{\mu_k^+(f_k)+\mu_k^-(f_k)}{2}.
\end{equation}
Taking the conditional expectation of $\widetilde{c}_k(\bm{h}_{k,i})$, computed w.r.t. the prediction samples $\bm{h}_{k,i}$ given classes $+1$ and $-1$, yields:
\begin{align}
\E_{L_k(+1)}\widetilde{c}_k(\bm{h}_{k,i})&=\frac{\mu_k^+(f_k)-\mu_k^-(f_k)}{2},\label{eq:c+}\\
\E_{L_k(-1)}\widetilde{c}_k(\bm{h}_{k,i})&=-\frac{\mu_k^+(f_k)-\mu_k^-(f_k)}{2}.\label{eq:c-}
\end{align}
The approximation $\widetilde{c}_k$ satisfies the conditions for consistent learning in \eqref{eq:consist} if \eqref{eq:c+} is strictly positive and if \eqref{eq:c-} is strictly negative. Note that the debiasing operation introduces a symmetry to \eqref{eq:c+} and \eqref{eq:c-}. Therefore both consistent learning conditions in \eqref{eq:consist} are satisfied by ensuring that the weaker condition $\mu_k^+(f_k)>\mu_k^-(f_k)$ holds, regardless of the sign of the individual terms $\mu_k^+(f_k)$ and $\mu_k^-(f_k)$. Thus, even in the biased case, in which $\mu_k^+(f_k)>\mu_k^-(f_k)>0$, consistent learning using $\widetilde{c}_k$ can be achieved.

We proceed now to formally translate the consistent learning conditions for the SL algorithms seen in \eqref{eq:consist}, considering the approximation $\widetilde{\bm{c}}_k(\bm{h}_{k,i})$ in \eqref{eq:choiceofc}.
First, we define the network average of the conditional means in \eqref{eq:mupm}:
\begin{equation}
	\mu^{+}(f)\triangleq
	\sum_{k=1}^K \pi_{k}\mu^+_k(f_k),\quad
	\mu^{-}(f)\triangleq
	\sum_{k=1}^K \pi_{k}\mu^-_k(f_k),\label{eq:meannetpm}
\end{equation}
and the network average of the \emph{empirical} training mean:
\begin{equation}
\widetilde{\bm{\mu}}(f)=\sum_{k=1}^K\pi_k\widetilde{\bm{\mu}}_{k}(f_k).
\end{equation}
The training phase will generate the set of models $\{\widetilde{\bm{f}}_k\}$, which are random with respect to the training datasets. Given a particular training setup, we can ``freeze'' the randomness of the training set and work conditionally on a particular realization of learned models $\{\widetilde{f}_k\}$. 

We are now interested in ascertaining whether or not these particular learned models allow for consistent learning {\em during the prediction phase}. To this end, we can apply the condition for consistent learning seen in \eqref{eq:consist} to the functions $\{\widetilde{c}_k\}$ in \eqref{eq:choiceofc}, for a frozen set of trained models $\{\widetilde{f}_k\}$, resulting in the following two conditions:
\begin{align}
\sum_{k=1}^{K}\pi_k \E_{L_k(+1)} \widetilde{f}_k(\bm{h}_{k,i})> \sum_{k=1}^{K}\pi_k 	\widetilde{{\mu}}_{k}(\widetilde{f}_k), \label{eq:cond1}\\
\sum_{k=1}^{K}\pi_k \E_{L_k(-1)}\widetilde{f}_k(\bm{h}_{k,i})<  \sum_{k=1}^{K}\pi_k 	\widetilde{{\mu}}_{k}(\widetilde{f}_k).\label{eq:cond2}
\end{align}
where we recall that $\E_{L_k(\gamma)}$ is the expectation computed with respect to the \emph{prediction} samples $\bm{h}_{k,i}$ under the distribution $L_k(h|\gamma)$, and the prediction samples are independent of any random variable generated in the training phase. Finally, substituting the definitions in \eqref{eq:mupm} and \eqref{eq:meannetpm} respectively into \eqref{eq:cond1} and \eqref{eq:cond2}, yields the following necessary conditions for consistent learning within the SML paradigm, conditionally on a given set of trained models $\{\widetilde{f}_k\}$.
\begin{equation}
\boxed{	\mu^+(\widetilde{{f}})>\widetilde{{\mu}}(\widetilde{{f}})~~\text{ and }~~\mu^-(\widetilde{{f}})<\widetilde{{\mu}}(\widetilde{{f}})}
	\label{eq:log2exprisk}
\end{equation}
Since, the above description is given conditioned on a set of trained models $\{\widetilde{f}_k\}$, the conditions in \eqref{eq:log2exprisk} depend on the randomness stemming from the training phase. Therefore, characterizing the consistency of learning requires characterizing probabilistically the occurrence of both events described in \eqref{eq:log2exprisk}. More precisely, we can define the {\em probability of consistent learning}, namely,
\begin{equation}
P_{c}\triangleq \P\left( \mu^+(\widetilde{\bm{f}})>\widetilde{\bm{\mu}}(\widetilde{\bm{f}})\,,\,\mu^-(\widetilde{\bm{f}})<\widetilde{\bm{\mu}}(\widetilde{\bm{f}})\right),\label{eq:probcons}
\end{equation}
where boldface fonts now highlight the randomness in the training set. In the next section, we provide the characterization of \eqref{eq:probcons} for classifiers belonging to general classes of bounded real-valued functions $\mathcal{F}_k$. In this case, we assume that there exists some real value $\beta>0$ such that:
	\begin{equation}
		|f_k(h)|\leq\beta, \quad f_k\in\mathcal{F}_k, \,h\in\mathcal{H}_k.\label{eq:boundedf}
	\end{equation}
For example, consider the linear logistic regression case seen in \eqref{eq:logregression}. In practical applications, features belong to a bounded set $\mathcal{H}_k$, and thus condition \eqref{eq:boundedf} would be satisfied if the vector of weights $w$ is constrained according to $\|w\|_2\leq b$, where $b$ is some positive real value. Similarly, in the multilayer perceptron example seen in \eqref{eq:mlpcase}, the condition in \eqref{eq:boundedf} is satisfied for norm-constrained neural networks~\cite{neyshabur2015norm}, i.e., where the weight matrices are bounded in norm by a certain positive real value $b$. 

\section{Consistency of Social Machine Learning}
We will need to call upon well-established statistical learning paradigms (e.g., the Vapnik-Chervonenkis theory) and adapt them to the {\em distributed network} setting considered in this work~\cite{vapnik2015uniform,devroye2013probabilistic}. 
	More specifically, we will move along the path summarized below.
	\begin{itemize}
		\item
		We will assume that the individual agents minimize an {\em empirical risk}, producing a collection of $K$ learned models, namely, the functions $\{\widetilde{\bm{f}}_k\}$. As usual, these functions are random due to the randomness of the training samples.
		\item
		We will examine the prediction (i.e., classification) performance obtained with the learned models $\{\widetilde{\bm{f}}_k\}$. In particular, we will establish technical conditions for the social learning algorithm to predict reliably the correct label as the number of streaming data gathered during the {\em prediction phase} increases. 
		\item
		Since the learned models inherit the randomness of the training set, the consistency guarantees must be formulated in a probabilistic manner --- see \eqref{eq:probcons}. Specifically, we guarantee a high probability that the samples in the training set lead to models $\{\widetilde{\bm{f}}_k\}$ that enable correct classification.
		\item
		As it happens in classical statistical learning frameworks, the interplay between empirical and optimal risk will be critical to ascertain the learning and prediction ability of the classifiers. However, differently from what is obtained in classical statistical learning frameworks, our results will depend significantly on the {\em graph} properties. In particular, a major role will be played by {\em weighted combinations of the individual risk functions}. The combination weights are the entries of the Perron eigenvector reflecting the combination matrix that governs the social learning interactions among the agents. This property leads to novel and interesting phenomena, for example, consistent classification can be achieved even if some of the agents learn bad models, but the plurality of the agents is able to reach a satisfying {\em aggregate} risk value. 
\end{itemize}

Under the framework described above, the nontrivial interplay between the training and prediction phases might lead to some confusion. Therefore, it is useful to clarify the main path followed in the forthcoming analysis.  We will focus on the probability of consistent learning $P_c$ in \eqref{eq:probcons}, namely, the probability that the training set produces, at the end of the {\em training} phase, a consistent classifier. 
By ``consistent'', we mean that the classifier is able to mark the unlabeled data observed during the {\em prediction} phase correctly as $i\rightarrow\infty$. 
The probability $P_c$ will be shown to be close to $1$ if the training set size is large enough, namely, we will show that consistent learning is achievable provided that {\em sufficient training} is allowed. As Theorem~\ref{the:consist} will show, in order to quantify the qualification ``sufficient'', it is critical to introduce a formal way to characterize the classifier structure.

The complexity of the classifier structure is related to the complexity of the class of functions $\mathcal{F}_k$. The latter is quantified by using the concept of \emph{Rademacher complexity} (initially introduced as Rademacher penalty in \cite{koltchinskii2001rademacher}). We follow the definition in \cite{boucheron2005theory} and \cite{neyshabur2015norm} and consider a class of functions $\mathcal{F}$ and a set $x$ with $N$ training samples, namely, $x\triangleq \{x_1,x_2\dots,x_N\}$, 
where $x_n\in\mathcal{X}$ for all $n=1,2,\dots,N$. We also introduce the set of vectors $\mathcal{F}\left(x\right)$ defined as:
\begin{equation}
 \mathcal{F}\left(x\right) \triangleq \Big\{[f(x_1),f(x_2),\dots,f(x_N)]\,\Big|\, x_n\in\mathcal X\,, f\in\mathcal{F}\Big\}.
\end{equation}
Then, the (empirical) Rademacher complexity associated with $\mathcal{F}\left(x\right)$ is:
\begin{equation}
\mathcal{R}\left(\mathcal{F}\left(x\right)\right)\triangleq\E_{r}\sup_{f\in \mathcal{F}}\left|\frac{1}{N}\sum_{n=1}^N\bm{r}_nf(x_n)\right|,\label{eq:defradem}
\end{equation}
where $\bm{r}_n$ are independent and identically distributed Rademacher random variables, i.e., with $\P(\bm{r}_n=+1)=\P(\bm{r}_n=-1)=1/2$. This quantity can be seen as a measure of \emph{overfitting} during training over the class of functions $\mathcal{F}$~\cite{bartlett2002model}. In general, to avoid overfitting during training, and to ensure an improved generalization performance, we choose models with small classifier complexity.

Applying the above definition to our multi-agent case, we define the individual empirical Rademacher complexity of agent $k$ for samples $h^{(k)}\triangleq \{h_{k,1}, \dots, h_{k,N_k}\}$ as
	\begin{equation}
	\mathcal{R}(\mathcal{F}_k(h^{(k)}))=\E_r \sup_{f_k\in\mathcal{F}_k}\left|\frac{1}{N_k}\sum_{n=1}^{N_k}\bm{r}_nf_k(h_{k,n})\right|,
	\end{equation}
	and its expected Rademacher complexity, for features $\bm{h}_{k,1}, \bm{h}_{k,2}, \dots, \bm{h}_{k,N}$ as
	\begin{equation}
	\rho_k\triangleq \E_{h_k} \mathcal{R}(\mathcal{F}_k(\bm{h}^{(k)})),\label{eq:singrad}
	\end{equation}
	which represents the Rademacher complexity of the {$k$-th classifier structure, {\em averaged} over the feature distribution}.
	We also define the (expected) {\em network Rademacher complexity} according to:
	\begin{equation}
\rho \triangleq\sum_{k=1}^{K}\pi_k \rho_k,\label{eq:averad}
	\end{equation}
which represents an average complexity across all agents in the network, weighted by their centrality scores (given by the elements of the Perron eigenvector $\pi$).

\subsection{Learning Consistency}
In Theorem~\ref{the:consist}, we show that the SML strategy consistently learns the truth during the prediction phase, with high probability as the number of training samples grows and for a moderately complex classifier structure. Before introducing the theorem, we define the following two quantities (we assume $N_k>0$ for all $k$):
	\begin{equation}
	\alpha_k\triangleq \frac{N_{\sf max}}{N_k},\qquad \alpha\triangleq \sum_{k=1}^K\pi_k \alpha_k,\label{eq:defalph}
	\end{equation}
	with $N_{\sf max}\triangleq \max_kN_k$. The {\em individual imbalance penalty} $\alpha_k$ quantifies how distinct the number of training samples of agent $k$ is compared with $N_{\sf max}$. The {\em network imbalance penalty} $\alpha$ is the average of $\alpha_k$ over the network, and it quantifies how unequal the training samples are across different agents. For example, if all agents possess the same number of training samples, i.e., $N_k=N_{\sf max}$, for all $k=1,2,\dots,K$, then $\alpha$ assumes minimal value with $\alpha = 1$. The value of $\alpha$ tends to grow when agents have very different number of training samples, e.g., when, for some $k$, $N_k \ll N_{\sf max}$.

Moreover, we assume that the target risk $\ropt$ is strictly smaller than $\log 2$. To understand the meaning of such assumption, we first consider a single agent $k$, for which $\ropt_k<\log 2$. This assumption eliminates the case where the classifier makes {\em uninformed} decisions of the form:
\begin{equation}
\widehat{p}_k(\gamma|h)=\frac{1}{2},\text{ for any }h\in\mathcal{H}_k\text{ and }\gamma\in\Gamma,\label{eq:uninf}
\end{equation}
i.e., where the classification decision is independent of the input feature vector. In this case, from \eqref{eq:logit}, $f_k(h)=0$ for any $h\in\mathcal{H}_k$, which in view of \eqref{eq:exprisk} implies $R_k(f_k)=\log 2$. This situation arises, for example, when the classifier structure is not complex enough to address the classification task at hand.
Requiring $\ropt_k<\log 2$ guarantees that the classifier $k$ performs better than a classifier that randomly assigns labels $+1$ and $-1$ with equal probability. Requiring that the {\em network} target risk satisfies $\ropt<\log 2$ is an even weaker assumption, since it establishes this bound to the risk values averaged over the graph. For example, suppose that, in a $K-$agent network, $K-1$ classifiers yield uninformed decisions like in \eqref{eq:uninf}, for which $\ropt_{k}=\log 2$. To satisfy $\ropt<\log 2$ on a network level, it suffices that one classifier performs better than the uninformed ones.

The next theorem characterizes the consistency of the SML strategy during the prediction phase in terms of an exponential lower bound on the probability of consistent learning in \eqref{eq:probcons}. 
\begin{theorem}[\textbf{SML Consistency}]\label{the:consist} For the logistic risk, assume that ${\sf R}^{o}<\log 2$ and that $f_k(h)\leq\beta$ for every $h\in\mathcal{H}_k$, $f_k\in\mathcal{F}_k$ and $k=1,2,\dots,K$, with $\beta>0$.
Assume $\rho< \mathscr{E}(\ropt)$, where $\mathscr{E}(\ropt)$ is exactly computed in \eqref{eq:defeps} and can be approximated as (see Fig.~\ref{fig:function} in Appendix~\ref{ap:theo1}):
\begin{equation}
\mathscr{E}(\ropt)\approx 0.2812\left(1-\frac{\ropt}{\log 2}\right).\label{eq:approxeps}
\end{equation}
Then, we have the following bound for the probability of consistent learning, defined in \eqref{eq:probcons}:
\begin{align}
	P_c&\geq 1-2\exp\left\{
	-\frac{8 N_{\sf max}}{\alpha^2\beta^2} \Big(\mathscr{E}(\ropt) - \rho\Big)^2\right\}.\label{eq:theocons}
	\end{align}
\end{theorem}
\begin{IEEEproof}
See Appendix~\ref{ap:theo1}.
\end{IEEEproof}
Theorem 1 has at least two important implications. First, if the network-average Rademacher complexity $\rho$ is smaller than the function $\mathscr{E}({\sf R}^o)$,  then the probability of consistent learning is bounded in an exponential way. 
Now, the function $\mathscr{E}({\sf R}^o)$ is an error exponent that determines how fast the probability of consistent learning approaches $1$. It is a function of the optimal risk ${\sf R}^o$ --- see the  definition in \eqref{eq:defeps}.
An excellent approximation for $\mathscr{E}({\sf R^o})$ is \eqref{eq:approxeps}, showing that such exponent quantifies how close the target risk is to the $\log 2$ risk boundary. As already discussed, the $\log 2$ risk boundary corresponds to the risk associated with a binary classifier that randomly classifies samples with labels $+1$ and $-1$.  The closer the target model is to the $\log 2$ risk, the smaller the value of $\mathscr{E}({\sf R}^o)$. In other words, smaller values of $\mathscr{E}({\sf R}^o)$ are symptomatic of more difficult classification problems. Therefore, Eq. \eqref{eq:theocons} reveals a remarkable interplay between the inherent difficulty of the classification problem (quantified inversely by $\mathscr{E}({\sf R}^o)$) and the complexity of the classifier structure (quantified by $\rho$). Ideally, we would like to have simple classification problems (i.e., higher values of $\mathscr{E}({\sf R}^o)$) and low Rademacher complexity $\rho$.
Notably, both indices are {\em network} indices that embody the network structure inside them.  

Second, the exponent characterizing the bound in \eqref{eq:theocons} depends on the size of the training sets at the individual agents (through the network imbalance penalty $\alpha$ and the maximum training-set size), and the bounding constant $\beta$. In particular, we see from \eqref{eq:theocons} that the exponent (and, hence, the probability of consistent learning) increases if we have larger training sets (i.e., larger $N_{\sf max}$ and/or smaller $\alpha$) and more constrained class of functions (i.e., smaller $\beta$).

Note that by considering an agent-dependent bound $\beta_k$ on the logit function $f_k(h)$, under similar assumptions as in Theorem~\ref{the:consist} and adjusting Theorem~\ref{the:esterrorbound} in Appendix~\ref{ap:auxtheo}, we can achieve the following alternative bound for the probability of consistent learning:
\begin{equation}
P_c\geq 1-2\exp\left\{
-\frac{8 N_{\sf max}}{\alpha_\beta^2} \Big(\mathscr{E}(\ropt) - \rho\Big)^2\right\},\label{eq:theocons2}
\end{equation}
where we define the network average quantity
\begin{equation}
\alpha_\beta \triangleq \sum_{k =1 }^K\pi_k \alpha_k \beta_k.
\end{equation}
The expression in \eqref{eq:theocons2} is useful to encompass scenarios in which different agents have different bounding constants, in particular, the case in which some agent $k'$ has a dominant bound on its logit function with respect to other agents, such that $\beta_{k'}\gg \beta_k$ for all $k\neq k'$. More generally, when $\pi_{k'} \alpha_{k'} \beta_{k'} \gg \pi_{k} \alpha_{k} \beta_{k}$ for some agent $k'$, the probability of consistent learning is approximately lower bounded as:
\begin{equation}
	P_c\geq 1-2\exp\left\{
	-\frac{8 N_{\sf max}}{(\pi_{k'}\beta_{k'}\alpha_{k'})^2} \Big(\mathscr{E}(\ropt) - \rho\Big)^2\right\}.
\end{equation}
In summary, the bound in \eqref{eq:theocons} can be used to establish conditions under which the probability of consistent learning approaches $1$ exponentially fast as the training-set sizes increase. 
To this end, we must observe that the quantity $\rho$ itself depends on the training-set sizes. 
Accordingly, it is necessary to obtain an estimate (or a bound) for the network-average Rademacher complexity. 
Once this is done, we will be in the position of evaluating the sample complexity of the SML strategy, namely, of evaluating how many samples are necessary to achieve a target probability of consistency.  This analysis will be pursued in the next section.

\subsection{Sample Complexity}
Under typical classifier structures, the Rademacher complexity scales as $C_k/\sqrt{N_k}$, where $N_k$ is the number of training samples pertaining to agent $k$, and $C_k$ is a constant quantifying the inherent complexity of the $k-$th classifier structure~\cite{neyshabur2015norm}.
As an example, we will show in the next section how the Rademacher complexity behaves for the particular structure of multilayer perceptrons, and provide an upper bound for a given design of number of hidden layers and hidden units. 

Now, assuming that the Rademacher complexity of each classifier $k$ is bounded as $C_k/\sqrt{N_k}$, the {\em network} Rademacher complexity will be bounded as:
\begin{equation}
\rho\leq \sum_{\ell=1}^K\pi_k\frac{C_k}{\sqrt{N_k}} =\frac{1}{\sqrt{N_{\sf max}}}\,
\underbrace{
\sum_{k=1}^K \pi_k C_k\sqrt{\alpha_k}
}_{\triangleq {\sf C}},\label{eq:rhoboundk}
\end{equation}
where ${\sf C}$ is an average constant that mixes the individual complexity constants $C_k$, accounting for the Perron eigenvector entries $\pi_k$ and the individual imbalance penalties $\alpha_k$. In the case where \eqref{eq:rhoboundk} is satisfied, exploiting \eqref{eq:theocons} we obtain the bound:
\begin{equation}
P_c\geq 1-
2\exp\left\{
-\frac{8 N_{\sf max}}{\alpha^2 \beta^2} \left(\mathscr{E}(\ropt) - \frac{{\sf C}}{\sqrt{N_{\sf max}}}\right)^2
\right\}.\label{eq:worstcbound}
\end{equation}
Equation \eqref{eq:worstcbound} shows that when $N_{\sf max}$ scales to infinity (with the relative proportions between $N_{\sf max}$ and $N_k$ kept fixed, i.e., $\alpha$ kept constant),  the probability of consistent learning approaches $1$ exponentially fast. Moreover, Eq. \eqref{eq:worstcbound} can be used to carry out a sample-complexity analysis of the SML strategy, as stated in the forthcoming theorem.
\begin{theorem}[\textbf{SML Sample Complexity}]\label{cor:1}
	Assume $\rho_k\leq C_k/\sqrt{N_{k}}$ for some constant $C_k>0$ for all $k=1,2,\dots, K$, and let 
	\begin{equation}
	{\sf C}\triangleq  \sum_{k=1}^K \pi_k C_k\sqrt{\alpha_k}.
	\end{equation}
	Then, for the logistic risk, consistent learning takes place with probability at least $1-\varepsilon$, if the maximum number of training samples across the network satisfies:
\begin{equation}
	N_{\sf max} >\left(\frac{{\sf C}}{\mathscr{E}(\ropt)}\right)^2
	\left(1 + \frac{\alpha\beta}{2{\sf C}}\sqrt{\frac{1}{2}\log\left(\frac{2}{\varepsilon}\right)}\right)^2.\label{eq:cor1}
	\end{equation}
\end{theorem}
\begin{IEEEproof}
	See Appendix~\ref{ap:cor}.
\end{IEEEproof}
We now examine how the relevant system parameters appearing in \eqref{eq:cor1} influence the sample complexity.
\paragraph{Target performance} The desired probability of consistent learning, $1 - \varepsilon$, influences the bound in \eqref{eq:cor1} only logarithmically, and, hence, has a mild effect on the necessary number of training samples.
\paragraph{Term $\alpha$} Term $\alpha$ quantifies how unequal the number of training samples is across agents. Larger values of $\alpha$ imply that agents have a more uneven number of samples, and thus require that $N_{\sf max}$ be increased to compensate for the lack of data at some agents in the network.
\paragraph{Term $\beta$} Term $\beta$ corresponds to the bound of the output of the logit function $f_k$ and, hence, increasing $\beta$ corresponds to increasing the possible logit functions to choose from. Accordingly, from \eqref{eq:cor1} we see that the larger the value of $\beta$, the larger the number of training samples necessary to result in highly probable consistent learning.
\paragraph{Term ${\sf C}$}
The constant ${\sf C}$ quantifies the complexity of the chosen classifier structure. The necessary number of training samples grows quadratically with an increase in the classifiers' complexity.
\paragraph{Term $\mathscr{E}(\ropt)$} As explained before, the term $\mathscr{E}({\sf R}^o)$ quantifies (inversely) the difficulty of the classification problem. Smaller values of $\mathscr{E}({\sf R}^o)$ are representative of more difficult classification problems, and accordingly correspond to the necessity of acquiring more training  samples.  
\paragraph{Role of the network} Given the networked nature of our inference problem, described in the early Sec.~\ref{sec:ip}, and the fact that the conditions for consistent learning are given with respect to network average values as seen in \eqref{eq:log2exprisk}, it is expected that the network structure plays a significant role in the results of Theorems~\ref{the:consist} and \ref{cor:1}.
The network influence, as well as the graph topology, are captured by the presence of the Perron eigenvector $\pi$ in the probability expression for consistent learning, through the network terms $\alpha$, $\rho$ and ${\sf R}^o$, namely, the network imbalance penalty, the network Rademacher complexity and the network target risk. 

The Perron eigenvector represents the centrality or influence of each agent in determining the values of the pertinent network terms, e.g., a more influential agent $k$ has more power to steer the value of the network target risk ${\sf R}^{o}$ towards its own private target risk ${\sf R}^{o}_k$. For doubly-stochastic combination matrices, the vector $\pi$ is a vector with elements $1/K$~\cite{sayed2014adaptation}, thus influence is uniform across agents. While the dependence on the structure connecting the classifiers is not found in existing statistical bounds in the literature for ensembles of classifiers~\cite{boucheron2005theory,cortes2014deep}, similar network average dependences are key quantities in distributed estimation and social learning~\cite{sayed2014adaptation,lalitha2018social,bordignon2021adaptive}. For example, in social learning, convergence occurs around the hypothesis $\gamma\in\Gamma$ that minimizes the network average KL divergence, i.e., $\sum_{k=1}^K\pi_k D_{\sf KL}[L_{k}(\gamma^o)|L_{k}(\gamma)]$~\cite{lalitha2018social,matta2020interplay,bordignon2021adaptive}.
  
In summary, Eq. \eqref{eq:cor1} quantifies how the main system parameters act on the SML sample complexity. Specifically, we see that: $i)$ 
owing to the exponential bound, the dependence on the target error probability $\varepsilon$ is mild; $ii)$ the number of samples to achieve a prescribed performance increases with the ``size'' of the class of functions (higher $\beta$), the heterogeneity among classifiers (higher $\alpha$), the complexity of classifiers (higher ${\sf C}$), and the difficulty of the learning problem (lower $\mathscr{E}({\sf R}^o)$); and $iii)$ the network role is encoded in the Perron eigenvector that appears in the {\em network-averaged} values $\alpha$, ${\sf C}$, and ${\sf R}^o$. 

In the next section, we discuss in greater detail the expression of the classifier complexity $\rho$ for feedforward neural networks as a function of the classifier structure, i.e., number of hidden layers (depth of the neural network) and weight of hidden units (width of the neural network), and the size of the training dataset. 
\subsection{Neural Network Complexity}
In this section, we complement the result from Theorem~\ref{the:consist} by showing that the term $\rho$ in \eqref{eq:averad}, which depends on the Rademacher complexity of the classifier, vanishes with an increasing number of training samples in the case of the MultiLayer Perceptron (MLP). Assume that one classifier has the structure of a MLP with $L$ layers (excluding the input layer) and activation function $\sigma$. We drop index $k$ as we are referring to a single MLP. Each layer $\ell$ consists of $n_\ell$ nodes, equivalently the size of layer $\ell$ is given by $n_\ell$. 

At each node $m=1,2,\dots,n_\ell$ of layers $\ell=2,3,\dots,L$, the following function $g_m^{(\ell)}$ is implemented:
\begin{equation}
	g_m^{(\ell)}(h)=\sum_{j=1}^{n_{\ell-1}}w^{(\ell)}_{mj}\sigma\left(g^{(\ell-1)}_j(h)\right).\label{eq:ffnn1}
\end{equation}
The parameters $w^{(\ell)}_{mj}$ correspond to the elements of the weight matrix $W_\ell$ of dimension $n_{\ell}\times n_{\ell-1}$. For the first layer, the function implemented at node $m$ is of the form:
\begin{equation}
	g_m^{(1)}(h)=\sum_{j=1}^{n_0}w^{(1)}_{mj}h_j,\label{eq:ffnn2}
\end{equation}
where the input vector $h$ has dimension $n_0$. A bias parameter can be incorporated in \eqref{eq:ffnn2} by considering an additional input element $h_{n_0+1}=1$.

For a MLP whose purpose is to solve a binary classification problem, we denote the output at layer $L$ by $z\in\mathbb{R}^2$, where  $z_m=g_m^{(L)}(h)$ for $m=1,2$. The final output is given by applying the softmax function to $z$, that is,
\begin{equation}
\widehat{p}(+1|h)=\frac{e^{z_1}}{e^{z_1}+e^{z_2}},\quad \widehat{p}(-1|h)=\frac{e^{z_2}}{e^{z_1}+e^{z_2}}.
\end{equation}
 In this case, the logit function is given by:
	\begin{equation}
		f^{\sf NN}(h)=\log\frac{ \widehat{p}(+1|h)}{\widehat{p}(-1|h)}=z_1-z_2,\label{eq:ffnn3}
	\end{equation}
	where we say that $f^{\sf NN}$ belongs to a class of functions $\mathcal{F}^{\sf NN}$, which is parameterized by matrices $W_\ell$, for $\ell=1,2,\dots,L$, according to \eqref{eq:ffnn1}, \eqref{eq:ffnn2} and \eqref{eq:ffnn3}.
	
	The general evolution for the Rademacher complexity of class $\mathcal{F}^{\sf NN}$ described above is well known in the literature as scaling with  $C/\sqrt{N}$~\cite{neyshabur2015norm}. We would like nonetheless to obtain an expression for this complexity, which depends explicitly on the design choices for the MLP, i.e., depending on the depth and weights of the network. The objective is to provide the user with a general guideline on how to choose these parameters for a desired complexity value. With this purpose, a formal upper bound for this complexity is enunciated in Proposition~\ref{lem:radnn} inspired by results from \cite{bartlett2002rademacher,neyshabur2015norm}.

\begin{proposition}[\textbf{Rademacher Complexity of Norm-Constrained MLPs}] \label{lem:radnn}Consider an $L$-layered multilayer perceptron, satisfying\footnote{Note that $\|W\|_1$ corresponds to the maximum column sum matrix norm of matrix $W$.} $\|W_{\ell}\|_1\leq b$, for every layer $\ell=1,2,\dots,L$. Assume that the input vector $x\in\mathbb{R}^{n_0}$ satisfies $\max_i |x_i|\leq c$, and that the activation function $\sigma(x)$ is Lipschitz with constant $L_\sigma$ with $\sigma(0)=0$. Then the Rademacher complexity for the set of vectors $\mathcal{F}^{\sf NN}(x)$ is bounded as:
	\begin{align}
		& \mathcal{R}(\mathcal{F}^{\sf NN}(x))\leq \frac{4}{\sqrt{N}}\left[(2bL_{\sigma})^{L-1}bc\sqrt{\log(2n_0)}\right].
	\end{align}
\end{proposition}
\begin{IEEEproof}
See Appendix~\ref{ap:e}.
\end{IEEEproof}
Assume we have a network of $K$ classifiers, each with a MLP structure. Given Proposition~\ref{lem:radnn}, we can explicitly characterize the constant $C_k$ found in \eqref{eq:rhoboundk} as:
 \begin{equation}
C_k= 4\left[(2b^{(k)}L_{\sigma}^{(k)})^{(L^{(k)}-1)}b^{(k)}c^{(k)}\sqrt{\log(2n^{(k)}_0)}\right], \end{equation}
 where we introduce superscript $(k)$ to indicate that the classifier structural parameters can change across different agents. This characterization in association with Theorem~\ref{cor:1} can be used to design the MLP architecture, according to the available training samples, or yet to select the number of samples needed for a given set of previously fixed architectures.

\section{Simulation Results}
\subsection{MNIST Dataset}\label{sec:mnist}
In the simulations, we consider the MNIST dataset~\cite{lecun2010mnist}, building a binary classification problem aimed at distinguishing digits $0$ and $1$. We employ a network of 9 {spatially distributed} agents, where each agent observes only a part of the image (see Fig.~\ref{fig:img}). These agents wish to collaborate and discover which digit corresponds to the image they are collectively observing. 

\begin{figure}[t]
	\includegraphics[width=0.75\textwidth]{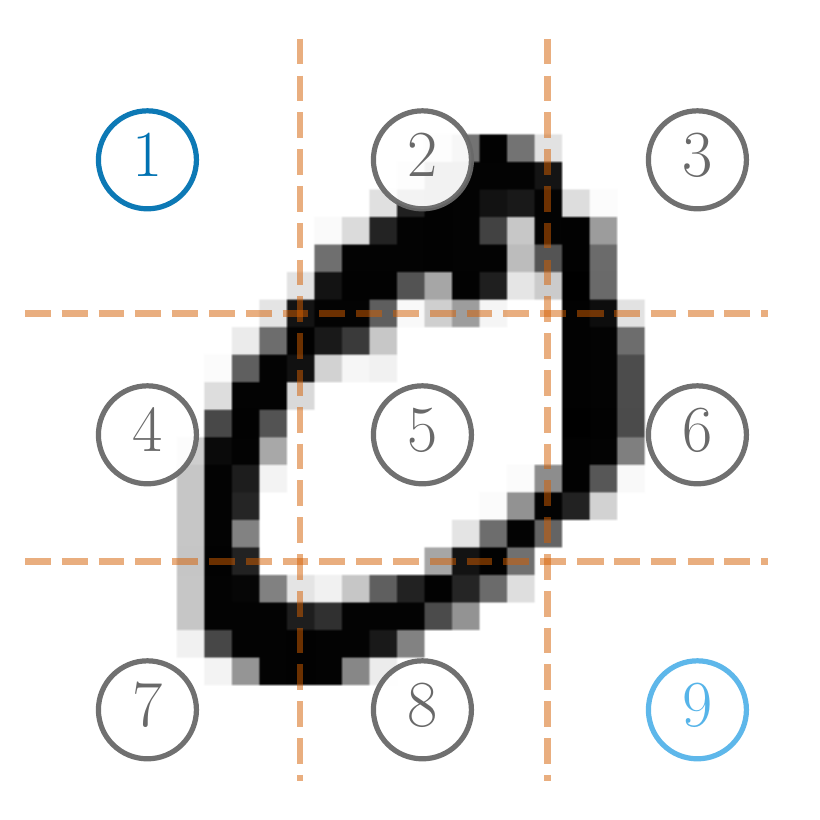}
	\caption{Each fraction of the image is observed by a different agent. Agents 1 and 9, highlighted in blue, correspond to the least informed agents.}
	\label{fig:img}
\end{figure}
\begin{figure}[t]
	\includegraphics[width=0.7\textwidth]{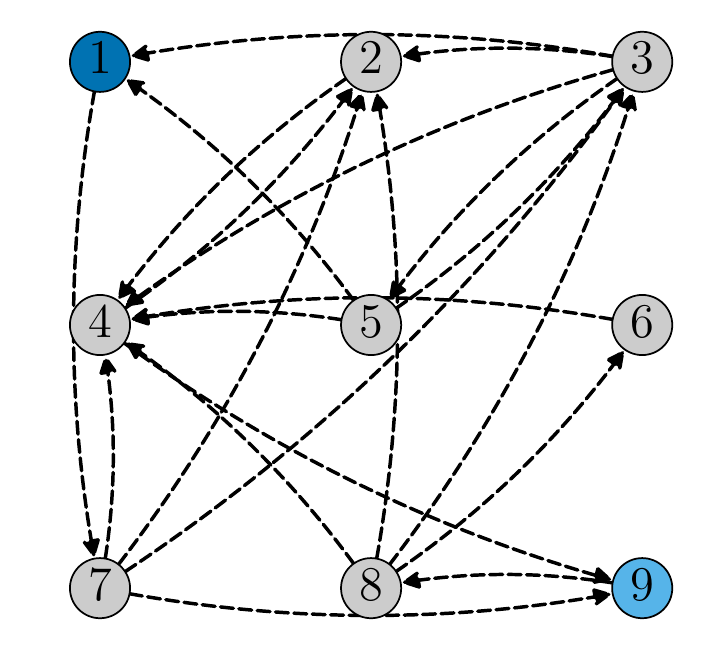}
	\caption{Topology of the network of agents.}
	\label{fig:net}
\end{figure}

As we can see in Fig.~\ref{fig:img}, different agents will observe data with different levels of informativeness, e.g., agents 1 and 9 will dispose of little or no information within their attributed image patch. To overcome this lack of local information, agents are connected through a strongly-connected network, whose combination matrix was generated using an averaging rule~\cite{sayed2014adaptation}. In Fig.~\ref{fig:net}, we show the network topology.

In the training phase, each agent is provided with a balanced set of $200$ labeled images. Using this set of examples, classifiers are independently trained using a MLP with $1$ hidden layer with $64$ hidden units and $\tanh$ activation function, over $30$ training epochs. The updates are performed using a batch size of $10$ with learning rate $0.001$. The training is repeated 5 times for each agent. The empirical training risk for each classifier over the training epochs can be seen in Fig.~\ref{fig:risk}, where the risk was averaged over the 5 training repetitions.

As expected, in Fig.~\ref{fig:risk} we see that classifiers 1 and 9 result in the least reliable training performances, i.e., their empirical risks exhibit the most variance across training. This could be problematic if these agents were to solve the classification problem on their own, but we will see that their individual poor classification performance is mitigated when collaborating within the network. 

In the prediction phase, agents observe unlabeled images over time. The nature of images switches at every \emph{prediction cycle}: In the cycle corresponding to interval $i\in[0,1000)$ agents start observing digits 0. In the following cycle, i.e., $i\in[1000, 2000)$, the nature of images changes to depict digits 1. Then, from instant $i=3000$ it switches back to digits 0, and so on. We implement the SML strategy based on the adaptive social learning algorithm described in Sec.~\ref{sec:SL}. In Fig.~\ref{fig:sl1}, we see the evolution of the decision variable $\bm{\lambda}_{1,i}$ for agent $1$ with $\delta = 0.01$.

\begin{figure}[t]
	\includegraphics[width=\textwidth]{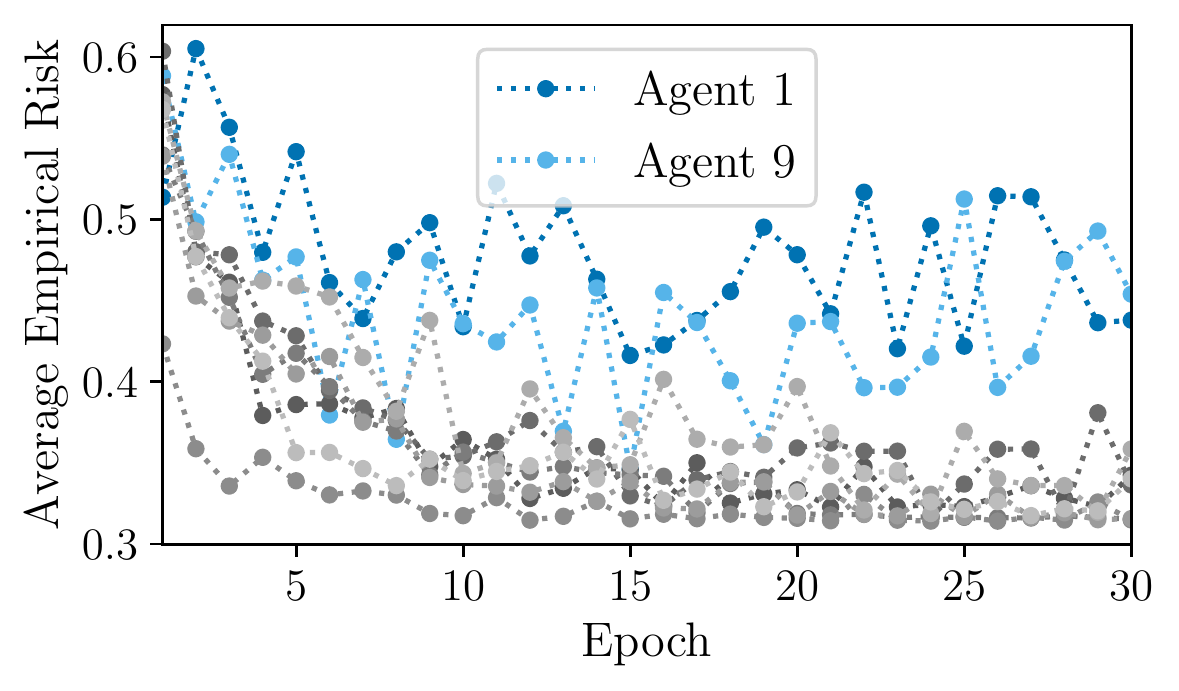}
	\caption{Empirical training risk averaged over 5 repetitions. The risks corresponding to agents 1 and 9 are highlighted in blue.}
	\label{fig:risk}
\end{figure}

\begin{figure}[t]
	\includegraphics[width=\textwidth]{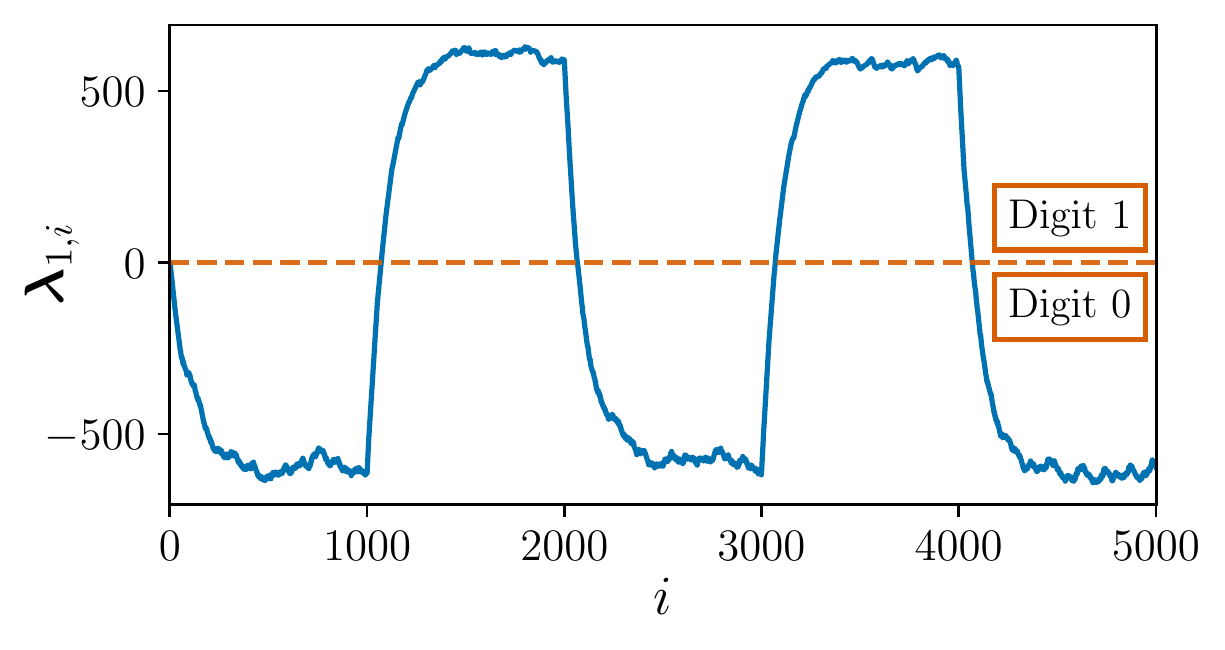}
		\caption{Evolution of the decision variable for agent 1 over the prediction phase. The observed digit is 0 within interval [0,1000), then it switches every 1000 time instants.}
	\label{fig:sl1}
\end{figure}

In Fig.~\ref{fig:sl1}, we see how, despite the limited information available during training, agent 1 is able to clearly distinguish digits 0 and 1. The instantaneous decision of agent 1 is given by the sign of the decision variable $\bm{\lambda}_{1,i}$ at any given time, i.e., whether the decision variable lies above or below the decision threshold (the orange dashed line in Fig.~\ref{fig:sl1}). Moreover, within the same prediction cycle, we can see in Fig.~\ref{fig:sl1} that the decision variable moves away from the decision threshold in the correct sense, i.e., it becomes more positive under digit 1 and more negative under digit 0. 

\subsection{Comparison with AdaBoost}
We compare the performance of the Social Machine Learning strategy with the classical AdaBoost strategy, as presented in~\cite{freund1999short}. In Boosting strategies, agents are trained sequentially, yielding a logit statistic $\widetilde{\bm{f}}^{\sf Boost}_k$ for each classifier $k$. The agents in this case are neural network classifiers, with the same architecture as described in the previous example. Once each agent is trained, its performance on the training dataset is evaluated and results in a boosting weight $a_k$ (see \cite{freund1999short} for further details on the implementation of AdaBoost). Larger values of $a_k$ indicate that agent $k$ has a better accuracy in the training dataset and makes less mistakes. 

During the prediction phase, as agents observe unlabeled data $\bm{h}_{k,i}$, the decision of an individual agent is given by:
\begin{equation}
\bm{\gamma}^{\sf Boost}_{k,i} = \text{sign}\left(\widetilde{\bm{f}}^{\sf Boost}_k(\bm{h}_{k,i})\right),\label{eq:hard}
\end{equation}
and the collective decision is performed using the boosting weights determined during training, according to:
\begin{equation}
\bm{\gamma}^{\sf Boost}_i= \text{sign}\left(\sum_{\ell=1}^Ka_\ell \bm{\gamma}^{\sf Boost}_{\ell,i}\right).
\end{equation}
Note that computing $\bm{\gamma}^{\sf Boost}_i$ requires centralized information, i.e., knowledge of the instantaneous decisions of all agents. We compare this centralized boosting decision with the individual instantaneous decision of agent $1$ from the SML strategy, whose decision variable $\bm{\lambda}_{1,i}$ was seen in Fig.~\ref{fig:sl1}. In a strongly-connected network, when agents repeatedly interact with their neighbors, information eventually propagates across all agents, and thus the decision variables $\bm{\lambda}_{k,i}$ tend to be similar over the network as time grows. The same holds for the decision $\bm{\gamma}^{\sf SML}_{k,i}$, defined in \eqref{eq:decsml}. Therefore it suffices to compare the decisions under Adaboost with the decisions of agent $1$ under the SML strategy.

The comparison can be seen in Fig.~\ref{fig:boost1}, for a similar prediction setup as previously described. As a result from training AdaBoost, the lowest boosting weights were obtained for agents $1, 3, 4, 9$. This result is expected since these agents are observing less relevant information (see Fig.~\ref{fig:img}) and can be regarded as the weakest agents.
\begin{figure}[ht]
	\includegraphics[width=\textwidth]{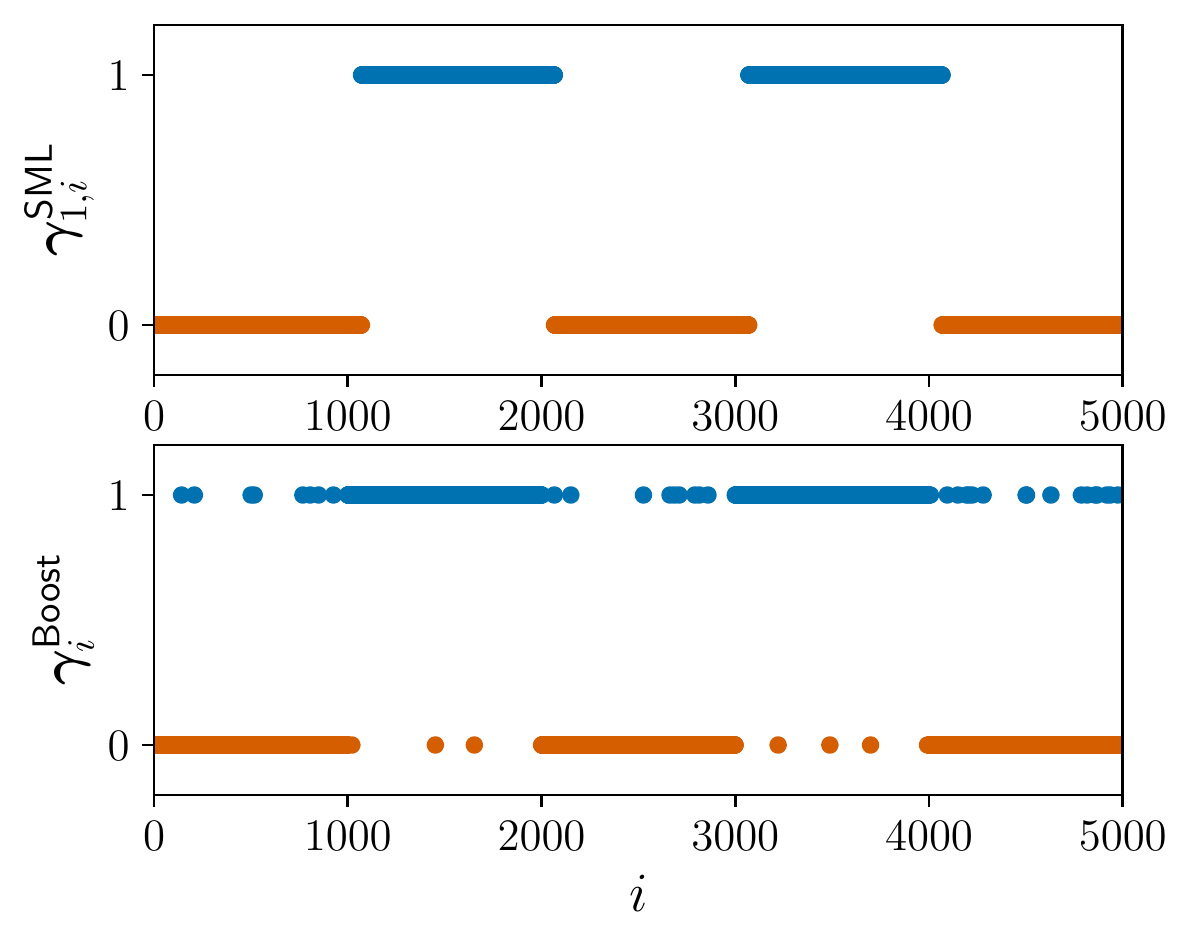}
	\caption{Comparison of the individual decision of agent $1$ within the SML framework and the collective AdaBoost decision. The observed digit is 0 within interval [0,1000), then it switches every 1000 time instants.}
	\label{fig:boost1}
\end{figure}

In Fig.~\ref{fig:boost1}, we see how the SML strategy results in misclassified samples only during short periods after the state transitions occur, whereas the AdaBoost solution makes mistakes throughout the prediction phase. This difference can be explained by the fact that the SML strategy uses the ASL protocol in the prediction phase:
\begin{align}
	\bm{\lambda}_{k,i}=\underbrace{\sum_{\ell\in\mathcal{N}_k}a_{\ell k}}_{\text{network}}\Big((1-\delta) \underbrace{\bm{\lambda}_{\ell,i-1}}_{
	\text{past}}+\underbrace{\widetilde{\bm{c}}_\ell (\bm{h}_{\ell,i})}_{\text{new}}\Big),\label{eq:asl}
\end{align} 
which introduces a combination {\em over time}, where the past information is combined with new data, and a combination {\em over the network}, where information is exchanged locally with neighbors. The combination over time introduces an adaptation time necessary for the algorithm to reach steady-state performance. The adaptation time scales as $1/\delta$ --- see~\cite{bordignon2021adaptive}. Furthermore, due to the decentralized combination over the network, information takes some iterations to propagate throughout the network, as we will further discuss in the example shown in Fig.~\ref{fig:boost2}. In contrast, Adaboost does not perform any kind of combination over time (since it does not aggregate information sequentially) or over the network (since the solution is centralized) and therefore there is no adaptation time associated with its behavior. Note that the fact that Adaboost makes instantaneous decisions without aggregating information over time results into a worse performance in steady state (higher error probability) see Fig.~\ref{fig:boost2}. In other words, the proposed social machine learning strategy takes significant advantage from integrating information over time, at the price of a small adaptation delay.

In the second simulation setup, we can observe how the SML strategy improves its learning performance (i.e., the error probability decreases) over time during the prediction phase. To emphasize this behavior, we reduce the number of hidden units in the neural network structure to 10, and the number of training samples available at each agent to 40. The SML strategy and AdaBoost are trained for the new setup, and deployed in two prediction scenarios: a {\em stationary} scenario, in which the true underlying class corresponds to digit 0 throughout the simulation period; and a {\em nonstationary} scenario, when the true underlying class starts at digit 0 and at instant $i=20$ switches to digit 1.

The SML strategy is implemented considering distinct social learning approaches. In the stationary scenario, we use traditional social learning (SL), implemented with the Bayesian update in \eqref{eq:bayest} and combination rule in \eqref{eq:combt}. In the nonstationary scenario, we use adaptive social learning (ASL), implemented with the adaptive Bayesian update in \eqref{eq:bayes} and combination rule in \eqref{eq:combt}. In Fig.~\ref{fig:boost2}, we depict the probability of error of the centralized Boosting algorithm and the SML strategy at agent $1$ (now with the choice of parameter $\delta=0.1$) for the two scenarios. The probability is empirically estimated from 1000 Monte Carlo runs. 

\begin{figure}[t]
	\includegraphics[width=\textwidth]{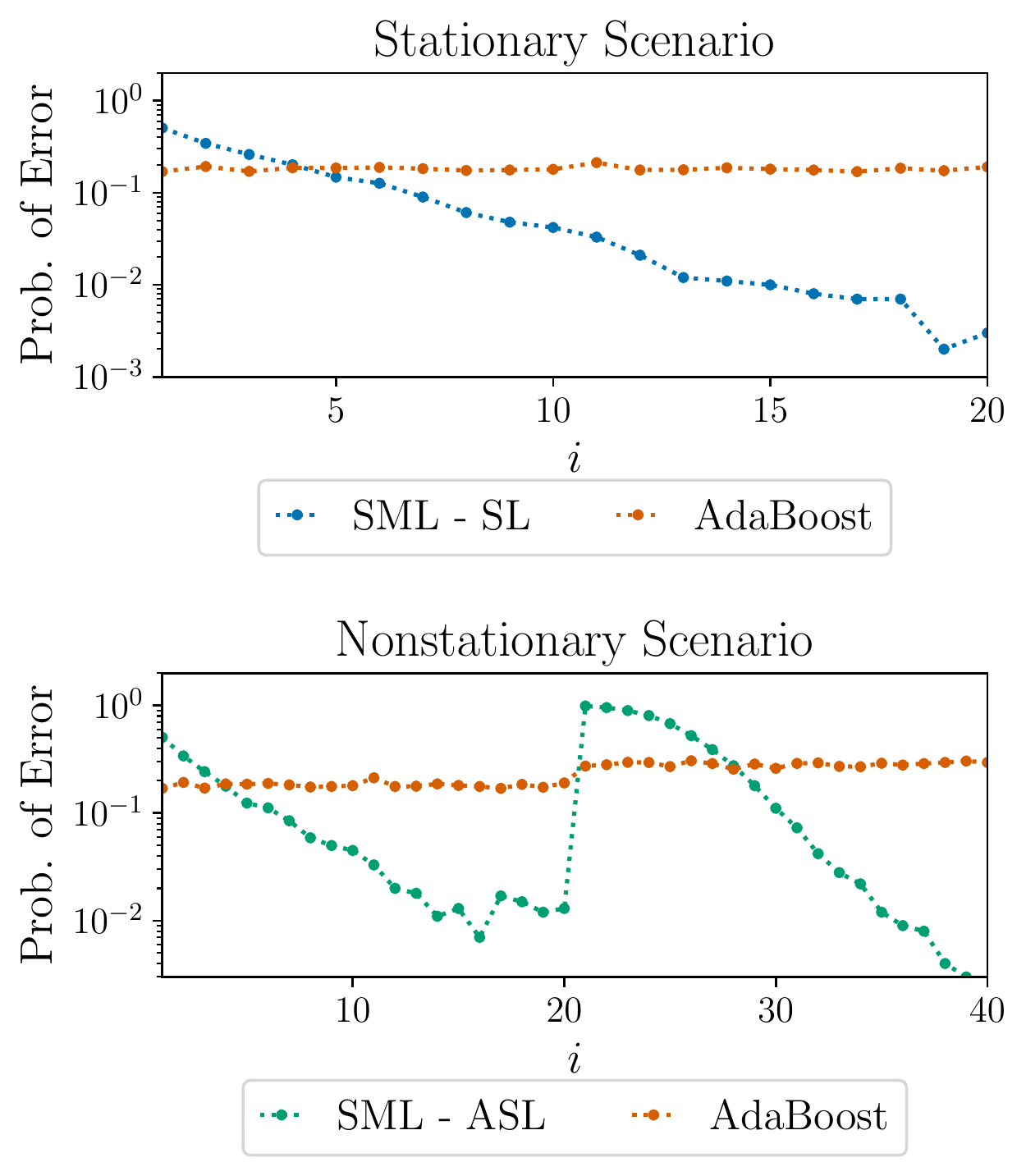}
	\caption{Evolution of the probability of error for the SML strategy and AdaBoost (centralized decision) estimated from 1000 Monte Carlo runs. {\em Upper panel}: SML is run with the traditional social learning rule (SL). The true state corresponds to digit $0$. {\em Bottom panel}: SML is run with the adaptive social learning rule (ASL). Until instant $i=20$, the true state corresponds to digit 0, after which the true state is digit 1. }
	\label{fig:boost2}
\end{figure}

In the upper panel of Fig.~\ref{fig:boost2}, we note that the SML strategy, associated with the SL protocol, quickly surpasses AdaBoost's performance and attains a significantly improved accuracy over time. Notably this improvement in accuracy exhibits a linear behavior as times progresses.  The traditional social learning strategy, although powerful, is not suitable to operate under nonstationary conditions, as discussed in~\cite{bordignon2021adaptive}. This is why, in the bottom panel of Fig.~\ref{fig:boost2}, we consider instead the SML strategy with the ASL protocol. In this scenario, we can clearly distinguish two prediction cycles, corresponding to the period under different underlying classes of digits. Compared with AdaBoost, SML yields the best performance as time passes. It is able to adapt its predictive behavior in view of the change in the observed digit, and it eventually surpasses the performance of AdaBoost with an adaptation time that scales with the chosen step-size $1/\delta$.

This improved performance can be explained by noticing the following aspect. The SML strategy leverages not only information distributed across agents, but also knowledge accumulated over time. We see that by considering, for example, the SML strategy associated with the SL protocol in \eqref{eq:SLrecursion}, the evolution of the decision variable of agent $k$ over the prediction phase is governed by the recursion:
\begin{align}
\bm{\lambda}_{k,i}&=\sum_{\ell\in\mathcal{N}_k}a_{\ell k}\Big(\bm{\lambda}_{\ell,i-1}+\widetilde{\bm{c}}_\ell(\bm{h}_{\ell,i})\Big).\label{eq:simaux}
\end{align} 
As we can see in \eqref{eq:simaux}, at every instant $i$, $\bm{\lambda}_{k,i}$ aggregates information from the past through the term $\bm{\lambda}_{\ell,i-1}$. The aggregation of past decision variables leverages the fact that the underlying class changes slowly during the prediction phase, thus allowing the classifiers to grow in confidence over time. 

We should also note that, in face of a single observation, i.e., at instant $i=1$ in Fig.~\ref{fig:boost2}, AdaBoost outperforms SML in its classification accuracy. As already mentioned, this can be explained by the fact that SML is a decentralized algorithm, i.e., at each iteration, agent $k$ communicates only with its one-hop neighbors. If the agent's neighbors happen to be poorly informed classifiers, then their $1-$iteration decision will also be unreliable. As the time passes, that is, as $i$ grows, this challenge is overcome due to the strong-connectivity of the graph topology, which enables the diffusion of information across all agents. In the example above, this is accomplished around instant $i=4$, when SML surpasses AdaBoost in performance.

\subsection{Multiple-hypothesis SML}
Consider now the multiple-hypothesis setting, where instead of a binary set we have a set of $M$ hypotheses $\Gamma\triangleq \{0,1,\dots, M-1\}$. If likelihood models $L_k(h|\gamma)$ are perfectly known, agents can use the social learning strategy \eqref{eq:bayest}--\eqref{eq:combt} to update their beliefs. Similarly to \eqref{eq:def_lc1} and \eqref{eq:def_lc}, let us make the following change of variables:
\begin{equation}
\bm{\lambda}_{k,i}(\gamma)\triangleq \log \frac{\bm{\varphi}_{k,i}(0)}{\bm{\varphi}_{k,i}(\gamma)},\quad 
c_{k}(\bm{h}_{k,i},\gamma)\triangleq \log \frac{L_{k}(\bm{h}_{k,i}|0)}{L_{k}(\bm{h}_{k,i}|\gamma)},\label{eq:defs2}
\end{equation}
for $\gamma\in\Gamma\setminus\{0\}$. Similarly to \eqref{eq:SLrecursion}, we can use \eqref{eq:defs2} to represent \eqref{eq:bayest}--\eqref{eq:combt} as the following linear recursion:
\begin{equation}
\bm{\lambda}_{k,i}(\gamma)=\sum_{\ell\in\mathcal{N}_k}a_{\ell k}\Big(\bm{\lambda}_{\ell,i-1}(\gamma) + c_{\ell}(\bm{h}_{\ell,i},\gamma)\Big), \quad \gamma\in\Gamma\setminus\{0\}.\label{eq:rec3}
\end{equation}
Likewise, we can write the adaptive social learning recursion, based on steps \eqref{eq:bayes} and \eqref{eq:combt}, as
\begin{equation}
	\bm{\lambda}_{k,i}(\gamma)=\sum_{\ell\in\mathcal{N}_k}a_{\ell k}\Big((1-\delta)\bm{\lambda}_{\ell,i-1}(\gamma) + c_{\ell}(\bm{h}_{\ell,i},\gamma)\Big),\label{eq:rec4}
\end{equation}
for $\gamma\in\Gamma\setminus\{0\}$. Note that, in both cases, from knowledge of $\bm{\lambda}_{k,i}(\gamma)$, we can recover the complete belief vector $\bm{\varphi}_{k,i}$ by considering that its entries must add up to $1$.

An instantaneous classification decision can be made by taking the hypothesis that maximizes the belief $\bm{\varphi}_{k,i}$:
\begin{equation}
\bm{\gamma}^{\sf SML}_{k,i}\triangleq \arg\max_{\gamma\in\Gamma} \bm{\varphi}_{k,i}(\gamma).
\end{equation}
The models $c_k(h,\gamma)$ are in practice unknown, therefore we extend next the SML approach to the multiple-hypothesis case.

\paragraph{Training phase}
Assume that each agent $k$ possesses a balanced dataset $\{\widetilde{\bm{h}}_{k,n},\widetilde{\bm{\gamma}}_{k,n}\}_{n=1}^{N_k}$ over the $M$ hypotheses in $\Gamma$, then under the assumption of uniform priors over the classes, and using Bayes’ rule, we can write:
\begin{equation}
c_k(h,\gamma)=\log \frac{L_{k}(h|0)}{L_{k}(h|\gamma) }= \log \frac{\widetilde{p}(0|h)}{\widetilde{p}(\gamma|h)}, \quad\gamma\in\Gamma\setminus\{0\},
\end{equation}
where $\widetilde{p}(\gamma|h)$ is the posterior probability of class $\gamma$ given the observation $h$.
In discriminative machine learning strategies, classifiers yield approximate posterior probabilities
\begin{equation}
\widehat{p}_k(\gamma|h; W), \quad \gamma\in\Gamma,
\end{equation}
where the parameters $W$ (e.g., $W=\{W_\ell\}$ can be the weight matrices of an $L$-layered neural network) are found by minimizing the cross-entropy risk:
\begin{equation}
R_k(W)\triangleq -\E_{\widetilde{h}_k,\widetilde{\gamma}_k}\log \widehat{p}_k(\widetilde{\bm{\gamma}}_{k,n}|\widetilde{\bm{h}}_{k,n};W).
\end{equation}
Using its training dataset, each agent $k$ can approximate the cross-entropy risk by its empirical counterpart:
\begin{equation}
\widetilde{\bm{R}}_k(W)\triangleq -\frac{1}{N_k}\sum_{n=1}^{N_k}\log \widehat{p}_k(\widetilde{\bm{\gamma}}_{k,n}|\widetilde{\bm{h}}_{k,n};W),
\end{equation}
whose minimizer is given by:
\begin{equation}
\widetilde{\bm{W}}_{k}\triangleq \argmin_{W\in\mathcal{W}_k}\widetilde{\bm{R}}_k(W),
\end{equation}
where the parameter space $\mathcal{W}_k$ depends on the chosen machine architecture (e.g., for a neural network the parameter space depends on the number of layers and neurons per layer).

\begin{figure*}[ht]
	\centering
	\includegraphics[width=\textwidth]{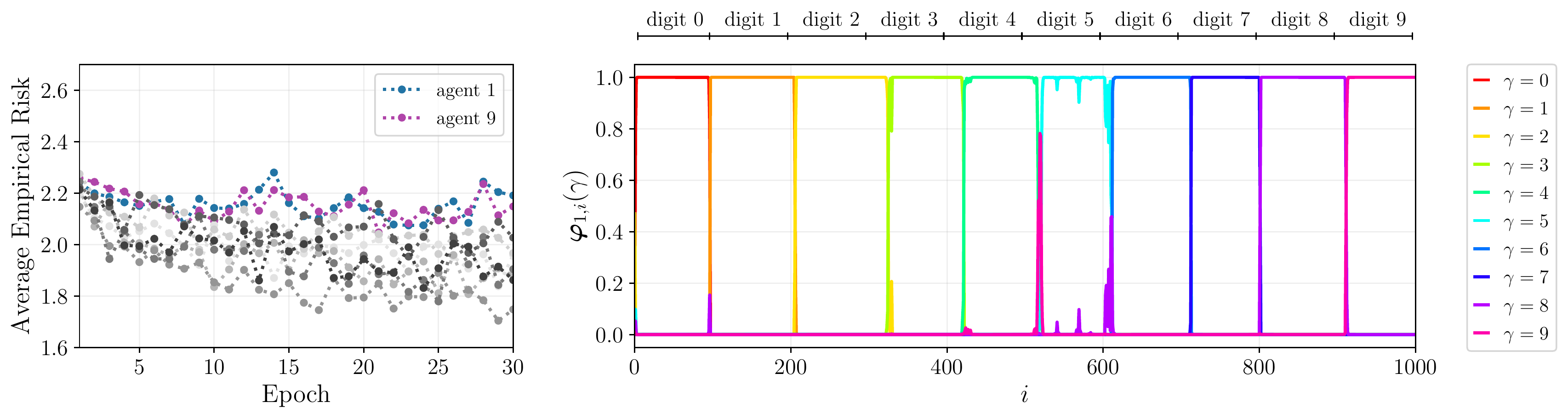}
	\caption{Social machine learning applied to the MNIST example with classes $\Gamma\triangleq \{0,1,\dots, 9\}$. ({\em Left}) Empirical training risk averaged over $5$ repetitions for all agents over training epochs. The risk corresponding to agents 1 and 9, the least informed agents, are highlighted in color. ({\em Right}) Belief evolution over time for agent 1 as the underlying true state changes at every $100$ time instants, i.e., $\gamma_0=0$ for $i\in[1,100)$, then $\gamma_0=1$ for $i\in[100,200)$, and so on.}\label{fig:multifig}
\end{figure*}

\paragraph{Prediction phase}
Upon training the above classifiers, agents can use the learned models to generate the approximate posterior probabilities given an observation $\bm{h}_{k,i}$:
\begin{equation}
\widehat{p}_k(\gamma|\bm{h}_{k,i}; \widetilde{\bm{W}}_k),\quad  \gamma \in\Gamma.
\end{equation}
During prediction, agents can implement the social learning scheme in \eqref{eq:rec3} by replacing the unknown statistic $c_k(\bm{h}_{k,i},\gamma)$ by:
\begin{equation}
\widetilde{\bm{c}}_{k}(\bm{h}_{k,i},\gamma)=\underbrace{\log \frac{\widehat{p}_k(0|\bm{h}_{k,i}; \widetilde{\bm{W}}_k)}{\widehat{p}_k(\gamma|\bm{h}_{k,i}; \widetilde{\bm{W}}_k)}}_{\triangleq \widetilde{\bm{f}}_k(\bm{h}_{k,i},\gamma)}-\widetilde{\bm{\mu}}_{k}(\widetilde{\bm{f}}_{k},\gamma), \quad\gamma\in\Gamma\setminus\{0\},\label{eq:cstat}
\end{equation}
where the second term on the RHS is the empirical training mean conditioned on classes $0$ and $\gamma$, defined as
\begin{equation}
\widetilde{\bm{\mu}}_{k}(\widetilde{\bm{f}}_{k},\gamma)\triangleq \frac{1}{|\mathcal{N}_{k,\gamma}|}\sum_{n\in \mathcal{N}_{k,\gamma}}\widetilde{\bm{f}}_k(\widetilde{\bm{h}}_{k,n},\gamma),
\end{equation}
where $\mathcal{N}_{k,\gamma}\triangleq \{n:\, \widetilde{\bm{\gamma}}_{k,n}\in\{0,\gamma\}\}$, which corresponds to the training samples associated with labels $0$ and $\gamma$. Note that if we reduce the multiple-hypothesis problem to a binary-hypothesis problem, i.e., $\Gamma=\{0,1\}$, then \eqref{eq:cstat} reduces back to \eqref{eq:choiceofc}, as we show next. When $\Gamma=\{0,1\}$, we deal with a single statistic $\widetilde{\bm{c}}_{k}(\bm{h}_{k,i},1)\triangleq \widetilde{\bm{c}}_{k}(\bm{h}_{k,i})$ and a single logit model $\widetilde{\bm{f}}_{k}(\bm{h}_{k,i},1)\triangleq \widetilde{\bm{f}}_{k}(\bm{h}_{k,i})$, while the empirical training mean conditioned on classes $0$ and $1$ reduces to:
\begin{equation}
\widetilde{\bm{\mu}}_{k}(\widetilde{\bm{f}}_{k},1)\triangleq\widetilde{\bm{\mu}}_{k}(\widetilde{\bm{f}}_{k})=\frac{1}{N_k}\sum_{n=1}^{N_k}\widetilde{\bm{f}}_{k}(\bm{h}_{k,i}),
\end{equation}
which is equivalent to the empirical training mean in \eqref{eq:emptrainmean} for the binary case.

We present next an experimental example, showing that the social machine learning strategy can be applied to the multiple-hypothesis classification problem, wherein a single machine is trained using data coming from multiple classes/ hypotheses.
\paragraph{Multi-class MNIST} Let us consider a similar setup as the one presented in Sec.~\ref{sec:mnist}, except that now we take into account all classes contained in the MNIST dataset, that is, $M=10$ where classes represent digits $0,1,2,\dots,9$. We consider the same network shown in Fig.~\ref{fig:net}, where each agent sees a patch of the handwritten image according to Fig.~\ref{fig:img}.

In the training phase, each agent is given a balanced set of $1000$ labeled images ($100$ images per digit). Classifiers are then independently trained using a MLP with $1$ hidden layer containing $64$ hidden units and using $\tanh$ activation function, over $30$ training epochs. The updates are performed using a batch size of $10$ with learning rate $0.001$. The empirical training risk for each classifier over the training epochs can be seen in the left panel of Fig.~\ref{fig:multifig}, where the risk was averaged over $5$ training repetitions.

In the prediction phase, agents observe unlabeled images over time. The nature of images changes at every $100$ samples: In $i\in[0,100)$ agents start observing digits $0$; In $i\in[100,200)$, agents observe digits $1$;  In $i\in[200,300)$, agents observe digits $2$, and so on. We implement the SML strategy based on the adaptive social learning algorithm in \eqref{eq:rec4} with the choice $\delta=0.1$, and with the statistic $\widetilde{\bm{c}}_{k}(h,\gamma)$ given in \eqref{eq:cstat}. In the right panel of Fig.~\ref{fig:multifig}, we see the evolution of the belief $\bm{\varphi}_{1,i}(\gamma)$ for agent $1$ over time $i$.

Note that, as the digit being observed by the agent changes, the algorithm is able to track these changes in such a way that the belief is maximized at the true state. In this example, the belief is not only maximized at the changing true state, but it is also able to concentrate its full confidence around the truth during most of the prediction phase.



\section{Concluding Remarks}
In this work, we focused on the following classification problem. A network of spatially distributed agents observes an event and all agents wish to determine the underlying class {exploiting a growing number of streaming observations collected over time}. Such problem has been thoroughly studied within the \emph{social learning} literature, where agents possess a set of possible models to explain their observations. By cooperating with neighbors, these agents are able to overcome local limitations and achieve collective consistent learning of the true underlying class.

These methods, however powerful, depend on the prior knowledge of the set of possible models, or \emph{likelihoods}, which characterize the distribution of observations given different underlying classes. In this work, we provided a fully data-driven solution to the aforementioned problem. We introduced Social Machine Learning, which is a two-step framework that allows aggregating the information perceived by heterogeneous classifiers to improve their decision performance over time, as the classifiers observe streaming data. { The classifiers are heterogeneous in the sense that their private observations originate from different distributions.} In our approach, we introduce a training phase that, with a finite training dataset, results in approximate models for the unknown data logit statistics. These models are deployed in a prediction phase, where one of the available social learning algorithms can be used.

We show that consistent learning in the prediction phase can be achieved with high probability, and we describe how the number of training samples should scale to yield the desired consistency. Furthermore, the decentralized collaboration among agents results in an increased robustness in face of poorly informed agents, as seen in the simulation results. Simulations also show that our solution continually improves performance over time, leveraging past acquired knowledge to make better informed decisions in the present.

\appendices

\section{Comparison between SML and~\cite{hare2021general}}\label{ap:compare}

Both~\cite{hare2021general} and our work address the issue of unspecified or unknown likelihood models, which arises naturally in practical applications where only data is available for inference. However, our approach is significantly different from the approach proposed in~\cite{hare2021general}, providing completely original contributions. We highlight next the main differences between the two works.
	
Consider the approach in~\cite{hare2021general}. The first step of the approach consists in choosing a proper family of parametric distributions. These distributions should satisfy two requirements. On one hand, they must have an analytical form that makes the subsequent steps viable. On the other hand, to guarantee good performance, they must match  the underlying physics of the observed phenomenon~\cite{hare2021general}. They should also satisfy a series of regularity conditions stated in Assumption 7 found in~\cite{hare2021general}. In contrast, in the SML strategy we make no assumption about the underlying distributions and rely instead on a model-independent and data-oriented approach by using a machine learning architecture, which can be as general as a neural network structure. One practical example is a residual neural network, e.g., the RESNET18 architecture~\cite{he2016deep} which has millions of parameters. These architectures are particularly suitable when the designer has very little preliminary knowledge concerning the nature of the dataset and would like to choose a structure that can {\em learn} the ``shape'' of the underlying distribution using only data.

In the SML strategy, we do not need to constrain our likelihood models to belong to any usual distribution family, such as Gaussian, multinomial, or Poisson. In contrast, the choice of the family of distributions is important in~\cite{hare2021general} to avoid the difficulties that arise in other steps of the procedure. For instance, in the second step the system designer must be able to derive a closed-form expression for the conjugate prior function, which is only well defined for a limited family of distributions~\cite{degroot2005optimal}. Without a conjugate prior expression, we also cannot determine a function to update its hyperparameters. Finally, the authors in~\cite{hare2021general} explain that even after performing the aforementioned steps it is possible that the uncertain likelihood update does not have a closed-form solution, in which case agents must use numerical methods to compute the belief update. 

The approach in~\cite{hare2021general} is valuable and is applicable in many scenarios of interest. 
However, our approach introduces significant novelty and is especially suited to strongly data-driven settings where little or no prior knowledge is available. In fact, the SML approach contemplates a wide variety of applications, including those envisioned in~\cite{hare2021general}, due to the general architecture used in training. 
For example, consider a realistic learning problem where agents are trying to detect the underlying class of a stream of images consisting of handwritten digits. We have shown in this work that the SML is particularly suitable for such applications. In our example, each agent trains a multilayer perceptron using their training datasets, by solving an empirical minimization problem. The samples from the MNIST dataset consist of $786$ pixels, which are partly observed by agents in such a way that each agent sees an image patch with $\sim 90$ pixels. It is unclear how the strategy proposed in~\cite{hare2021general} could be used in this case, first, due to the more diverse statistics of the data samples compared to the distribution families considered in~\cite{hare2021general}, and second, due to the high-dimensionality of the samples.

We show next that, with reference to an example proposed in~\cite{hare2021general}, where features follow a 2-dimensional Gaussian distribution, the SML strategy delivers better performance than the strategy in~\cite{hare2021general}, with the advantage of not requiring knowledge about the parameterization of the true distributions.

\subsection{2-dimensional Gaussian Example}
 Inspired by the 2-dimensional Gaussian example in~\cite{hare2021general}, we consider the following simulation setup with binary classes, i.e., $\gamma\in\{-1,+1\}$. We consider the same 4-agent network used in~\cite{hare2021general}, where the combination matrix is given by:
\begin{equation}
A=\begin{bmatrix}
	0.5&0.5&0&0\\
	0&0.5&0.5&0\\
	0&0&0.5&0.5\\
	0.5&0&0&0.5
\end{bmatrix}.
\end{equation}
Define the following distributions:
\begin{align}
g_1\triangleq \mathcal{G}\left({\sf m}_1,\Sigma_1 \right),\qquad g_2\triangleq \mathcal{G}\left({\sf m}_2,\Sigma_2 \right)
\end{align}
where $\mathcal{G}({\sf m}, \Sigma )$ represents a multivariate Gaussian distribution with mean vector ${\sf m}$ and covariance matrix $\Sigma$. In this example, we choose
\begin{equation}
{\sf m}_1={\sf m}_2\triangleq \begin{bmatrix}
	0&0
\end{bmatrix},
\end{equation}
and
\begin{equation}
\Sigma_1\triangleq \begin{bmatrix}
	1&0\\0&1
\end{bmatrix},\qquad \Sigma_2\triangleq \begin{bmatrix}
	1.5&0\\0&1.5
\end{bmatrix}.
\end{equation}
Table~\ref{tab:likes} shows the correct likelihood models for each agent, i.e., the true distribution of their observations $h$ given different hypotheses $\gamma$. Note that agent $2$ is the only informative agent, whose likelihoods are different given different classes.
\begin{table}[ht]
	\centering
	\begin{tabular}{c|c|c|c|c}
		&agent $1$&agent $2$&agent $3$&agent $4$\\
		\hline
		$L_k(h|+1)$&$g_1$&$g_1$&$g_1$&$g_1$\\
		\hline
		$L_k(h|-1)$&$g_1$&$g_2$&$g_1$&$g_1$
	\end{tabular}
	\caption{Likelihood models setup.}\label{tab:likes}
\end{table}

We implement the strategy with uncertain likelihoods proposed in~\cite{hare2021general}, assuming that agents use multivariate Gaussian distributions as the parameterized family of distributions. We assume that all agents possess the same number of training samples, i.e., $N_1=N_2=N_3=N_4\triangleq 2N$, where $N$ corresponds to the number of training samples per class. During prediction, we assume the true state is $+1$. The details in implementation follow the steps in~\cite{hare2021general}.

\begin{figure*}[t]
	\centering
	\includegraphics[width=\textwidth]{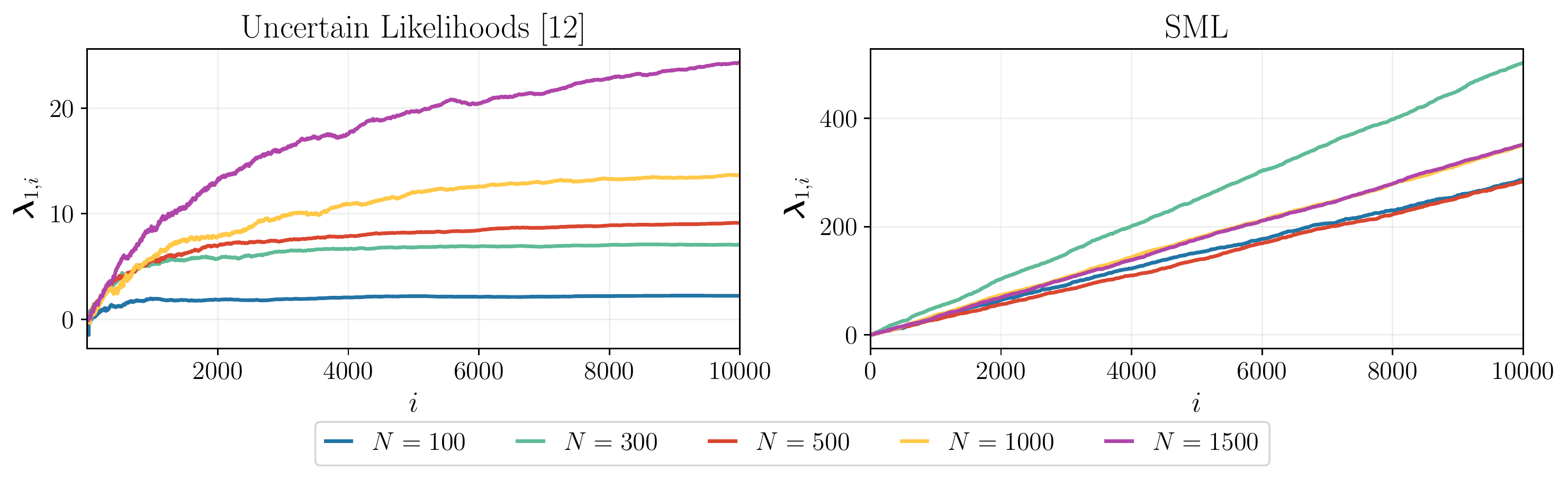}
	\caption{Evolution over time of the log-ratio of beliefs for agent $1$, averaged across $10$ experiments regarding the $2$-D Gaussian example using two different approaches. ({\em Left}) Social learning with uncertain models~\cite{hare2021general}. ({\em Right}) Social machine learning (SML).}\label{fig:compare}
\end{figure*}	

For the sake of comparison with the SML strategy, we choose to display the evolution of the log-ratio of beliefs, namely
\begin{equation}
\bm{\lambda}_{k,i}\triangleq \log \frac{\bm{\varphi}_{k,i}(+1)}{\bm{\varphi}_{k,i}(-1)}
\end{equation}
which can be seen in the left panel of Fig.~\ref{fig:compare} for agent $1$ and for different sizes of training sets, i.e., for different $N$. Note that for all training scenarios, the curve tends to converge to a fixed positive value, which implies that, on one hand, the belief is maximized at the true hypothesis $+1$, i.e., $\bm{\varphi}_{1,i}(+1)>\bm{\varphi}_{1,i}(-1)$ for large $i$, and on the other hand, that the belief at the wrong hypothesis $\bm{\varphi}_{1,i}(-1)$ does not vanish, i.e., $\bm{\varphi}_{k,i}(-1)>0$ for large $i$.

We apply the SML approach to the aforementioned setup. We consider that each agent can train a feedforward neural network, with $2$ hidden layers with $10$ hidden units each and $\tanh$ activation function. The training is performed over $1000$ epochs with learning rate $0.0001$. We see in the right panel of Fig.~\ref{fig:compare} the evolution of the log-ratio of beliefs over time for different sizes of training sets. For all training setups, the curves grow linearly with increasing time $i$. This implies not only that the belief is maximized at the true hypothesis, but also that for large $i$ the belief corresponding to the wrong hypothesis $-1$ vanishes, i.e., $\bm{\varphi}_{k,i}(-1)\rightarrow 0$, which means that $\bm{\varphi}_{k,i}(+1)\rightarrow 1$ (since the entries of the belief add up to $1$).

Note that the profound difference between the behavior of the log-ratio of beliefs implies a significant difference in performance between the two strategies. In the SML strategy, this diverges linearly, as it happens in traditional social learning with {\em known} likelihood models. This linear growth implies an exponentially fast convergence of the belief vector toward the right hypothesis, with a probability of correct learning that improves steadily (and exponentially) as time elapses. These benefits are lost with the approach in~\cite{hare2021general}, since the log-ratio of beliefs is no longer diverging exponentially.

\section{Proof of Theorem~\ref{the:consist}}\label{ap:theo1} 
Before detailing the proof of Theorem~\ref{the:consist}, we provide a roadmap of the proof. The proof starts from Lemma~\ref{lem:prob} in this same Appendix, resulting in the lower bound \eqref{eq:lemprob}. On the RHS of \eqref{eq:lemprob}, there are two probability terms that should be upper bounded. The first one is bounded using Eq.~\eqref{eq:theombound} in Theorem~\ref{the:esterrorbound} (found in Appendix~\ref{ap:auxtheo}), while the second one is bounded using Lemma~\ref{lem:boundest} (found in Appendix~\ref{ap:aux}) and then Eq.~\eqref{eq:theobound} in Theorem~\ref{the:esterrorbound}. After some algebraic manipulations, we reach the final result of Theorem~\ref{the:consist}.

We present next Lemma~\ref{lem:prob}, which provides a lower bound on the probability of consistent learning. We denote the total expected value of $f_k(\widetilde{\bm{h}}_{k,n})$ by:
\begin{align}
\mu_k(f_k)&\triangleq {\E_{\tilde{h}_k} f_k(\widetilde{\bm{h}}_{k,n})=\frac{\mu_{k}^+(f_k)+\mu_{k}^-(f_k)}{2},\label{eq:avemean}}
\end{align}
where we considered equal priors over the two classes $+1$ and $-1$}. We also denote its average across the network by:
\begin{align}
\mu(f)&\triangleq \sum_{\ell=1}^K\pi_k \mu_k(f_k)\stackrel{\text{(a)}}{=}{\frac{\mu^+(f)+\mu^-(f)}{2},\label{eq:netavemean}}
\end{align}
where (a) follows from using \eqref{eq:avemean} {and the definition of $\mu^+(f)$ and $\mu^-(f)$ found in \eqref{eq:meannetpm}.
\begin{lemma}[\textbf{Probability Bound for Consistent Learning}]\label{lem:prob}
For any $d>0$, we have that:
	\begin{align}
	P_c&\geq 1
		-\P\left(\left|\widetilde{\bm{\mu}}(\widetilde{\bm{f}})-\mu(\widetilde{\bm{f}})\right|\geq d\right)-\P\left(R(\widetilde{\bm{f}})\geq \Delta\right),\label{eq:lemprob}
	\end{align}
where $\Delta\triangleq \log (1+e^{-d})$ and $P_c$ is the probability of consistent learning defined in \eqref{eq:probcons}.
\end{lemma}
\begin{IEEEproof}[Proof]
	Define the following events, which will be used in the proof:
	\begin{align}
	&\mathcal{A}\triangleq \left\{\Big|\mu(\widetilde{\bm{f}})- \widetilde{\bm{\mu}}(\widetilde{\bm{f}})\Big|\geq\frac{\mu^+(\widetilde{\bm{f}})-\mu^-(\widetilde{\bm{f}})}{2}\right\},\label{eq:defa}\\
	&\mathcal{B}\triangleq\left\{\frac{\mu^+(\widetilde{\bm{f}})-\mu^-(\widetilde{\bm{f}})}{2}> d\right\}.\label{eq:defb}
	\end{align}
	First, in view of \eqref{eq:probcons} and using de Morgan's law~\cite{billingsley2008probability}, we can write:
\begin{align}
	&1-P_c=\P\Big(\Big\{\mu^+(\widetilde{\bm{f}})\leq\widetilde{\bm{\mu}}(\widetilde{\bm{f}})\Big\}\medcup\Big\{ \mu^-(\widetilde{\bm{f}})\geq\widetilde{\bm{\mu}}(\widetilde{\bm{f}})\Big\}\Big)\nonumber\\
	&\stackrel{\text{(a)}}{=}\P\Big(\Big\{\mu^+(\widetilde{\bm{f}})-\mu(\widetilde{\bm{f}})\leq\widetilde{\bm{\mu}}(\widetilde{\bm{f}})-\mu(\widetilde{\bm{f}})\Big\}\nonumber\\
	&\qquad\medcup\Big\{\mu^-(\widetilde{\bm{f}}) -\mu(\widetilde{\bm{f}})\geq\widetilde{\bm{\mu}}(\widetilde{\bm{f}})-\mu(\widetilde{\bm{f}})\Big\}\Big)\nonumber\\
	&\stackrel{\text{(b)}}{=}\P \Bigg(\Bigg\{ \mu^+(\widetilde{\bm{f}})-\frac{\mu^+(\widetilde{\bm{f}})+\mu^-(\widetilde{\bm{f}})}{2}\leq\widetilde{\bm{\mu}}(\widetilde{\bm{f}})-\mu(\widetilde{\bm{f}})\Bigg\} \nonumber\\
	&\qquad \medcup\Bigg\{\mu^-(\widetilde{\bm{f}}) -\frac{\mu^+(\widetilde{\bm{f}})+\mu^-(\widetilde{\bm{f}})}{2}\geq\widetilde{\bm{\mu}}(\widetilde{\bm{f}})-\mu(\widetilde{\bm{f}})\Bigg\}\Bigg)\nonumber\\
	&=\P \Bigg(\Bigg\{ \frac{\mu^+(\widetilde{\bm{f}})-\mu^-(\widetilde{\bm{f}})}{2}\leq-\Big(\mu(\widetilde{\bm{f}})-\widetilde{\bm{\mu}}(\widetilde{\bm{f}})\Big)\Bigg\} \nonumber\\
	&\qquad  \medcup\,\Bigg\{\frac{\mu^+(\widetilde{\bm{f}})-\mu^-(\widetilde{\bm{f}})}{2}\leq\mu(\widetilde{\bm{f}})- \widetilde{\bm{\mu}}(\widetilde{\bm{f}})\Bigg\}\Bigg)\nonumber\\
	&\stackrel{\text{(c)}}{=}\P(\mathcal{A})\nonumber\\
	& \stackrel{\text{(d)}}{=}\P\Big(\mathcal{A}\,,\,\mathcal{B})+\P\Big(\mathcal{A}\,,\,\overline{\mathcal{B}}\Big)\nonumber\\
	&\stackrel{\text{(e)}}{\leq} \P\left(\Big|\mu(\widetilde{\bm{f}})-\widetilde{\bm{\mu}}(\widetilde{\bm{f}})\Big|\geq d\right)+\P\left(\frac{\mu^+(\widetilde{\bm{f}})-\mu^-(\widetilde{\bm{f}})}{2}\leq d\right),\label{eq:lemma1a}
	\end{align}
	where in (a) we subtract $\mu(\widetilde{\bm{f}})$ from both terms within the probability operator, in (b) we replace $\mu(\widetilde{\bm{f}})$ with \eqref{eq:netavemean}, and (c) follows from the following relation:
\begin{align}
&\Bigg\{ \frac{\mu^+(\widetilde{\bm{f}})-\mu^-(\widetilde{\bm{f}})}{2}\leq-\Big(\mu(\widetilde{\bm{f}})-\widetilde{\bm{\mu}}(\widetilde{\bm{f}})\Big)\Bigg\} \nonumber\\&\medcup\, \Bigg\{\frac{\mu^+(\widetilde{\bm{f}})-\mu^-(\widetilde{\bm{f}})}{2}\leq\mu(\widetilde{\bm{f}})- \widetilde{\bm{\mu}}(\widetilde{\bm{f}})\Bigg\}\nonumber\\
& \Leftrightarrow \Bigg\{\Big|\mu(\widetilde{\bm{f}})-\widetilde{\bm{\mu}}(\widetilde{\bm{f}})\Big|\geq \frac{\mu^+(\widetilde{\bm{f}})-\mu^-(\widetilde{\bm{f}})}{2}\Bigg\} \triangleq \mathcal{A}.
\end{align}
In (d), we used the law of total probability, and (e) follows from:
\begin{align}
	&\mathcal{A}\,\medcap \mathcal{B}\Rightarrow \Big\{\Big|\mu(\widetilde{\bm{f}})-\widetilde{\bm{\mu}}(\widetilde{\bm{f}})\Big|\geq d\Big\},
	\end{align}
	where $\mathcal{B}$ is defined in \eqref{eq:defb}, and also from the fact that the probability of intersection of two events is upper bounded by the probability of one of the events.
	
	Let us address the second probability term on the RHS of \eqref{eq:lemma1a}. Consider the average network risk evaluated on the training samples $(\widetilde{\bm{h}}_{k,n}, \widetilde{\bm{\gamma}}_{k,n})$, computed for a given function $f_k\in\mathcal{F}_k$:
	\begin{align}
	&\sum_{k=1}^K\pi_kR_k(f_k)\nonumber\\
	&=\sum_{k=1}^K\pi_k\E_{\tilde{h}_k,\tilde{\gamma}_k} \log \Bigg(1+\exp\Big(- \widetilde{\bm{\gamma}}_{k,n}f_k(\widetilde{\bm{h}}_{k,n})\Big)\Bigg)\nonumber\\
	&\stackrel{\text{(a)}}{\geq} \sum_{k=1}^K\pi_k\log \Big( 1+\exp\Big(-\E_{\tilde{h}_k, \tilde{\gamma}_k} \widetilde{\bm{\gamma}}_{k,n}f_k(\widetilde{\bm{h}}_{k,n}) \Big)\Big)\nonumber\\
	&\stackrel{\text{(b)}}{\geq} \log \Big( 1+\exp\Big(-\sum_{k=1}^K\pi_k\E_{\tilde{h}_k, \tilde{\gamma}_k}\widetilde{\bm{\gamma}}_{k,n}f_k(\widetilde{\bm{h}}_{k,n}) \Big)\Big)\nonumber\\
	&\stackrel{\text{(c)}}{\geq} \log \Big( 1+\exp\Big(\frac{1}{2}\sum_{k=1}^K\pi_k\E_{L_k(-1)}f_k(\widetilde{\bm{h}}_{k,n})\nonumber\\
	&\quad-\frac{1}{2}\sum_{k=1}^K\pi_k\E_{L_k(+1)}f_k(\widetilde{\bm{h}}_{k,n}) \Big)\Big)\nonumber\\
	&\stackrel{\text{(d)}}{=}\log \Bigg( 1+\exp\Bigg(-\frac{\mu^+(f)-\mu^-(f) }{2}\Bigg)\Bigg),\label{eq:lem1aux}
	\end{align}
	where in (a) and (b) we used Jensen's inequality with the convexity of function $\log (1+e^x)$. In (c), we used the assumption of uniform priors during training, and in (d) we used the definition of the conditional means averaged over the network found in \eqref{eq:mupm} and \eqref{eq:meannetpm}. From \eqref{eq:lem1aux}, we have the following implication for a given $f_k\in\mathcal{F}_k$ for $k=1,2,\dots, K$:
	\begin{equation}
	\frac{\mu^+(f)-\mu^-(f)}{2}\leq d\Rightarrow \hspace{-3pt} \sum_{k=1}^N\pi_kR_k(f_k)\geq \log\left(1+e^{-d} \right).\label{eq:riskp}
	\end{equation}
Replace $f_k$ in \eqref{eq:riskp} by $\widetilde{\bm{f}}_k$ (i.e., the models obtained in the training phase). Then \eqref{eq:riskp} results in:
	\begin{align}
	&\P\Big(\sum_{k=1}^N\pi_kR_k(\widetilde{\bm{f}}_k)\geq \log \left(1+e^{-d}\right)\Big)\nonumber\\&\geq \P\left(\frac{\mu^+(\widetilde{\bm{f}})-\mu^-(\widetilde{\bm{f}})}{2}\leq d\right)=\P\left(\overline{\mathcal{B}}\right).\label{eq:lemma1b}
	\end{align}	
	Using \eqref{eq:lemma1b} in \eqref{eq:lemma1a} and defining $\Delta\triangleq \log (1+e^{-d})$ yields the bound in \eqref{eq:lemprob}.
\end{IEEEproof}
\begin{IEEEproof}[Proof of Theorem~\ref{the:consist}]
	From Lemma~\ref{lem:prob}, we obtain the lower bound in \eqref{eq:lemprob} for the probability of consistent learning. Next, we need to examine each of the terms on the RHS of \eqref{eq:lemprob}.
	
Regarding the first term, we can write:
		\begin{equation}
		\left|\widetilde{\bm{\mu}}(\widetilde{\bm{f}})-\mu(\widetilde{\bm{f}})\right|\leq \sup_{f\in \mathcal{F}}\left|\widetilde{\bm{\mu}}(f)-\mu(f)\right|,
		\end{equation}
which implies that
	\begin{equation}
	\P\left(\left|\widetilde{\bm{\mu}}(\widetilde{\bm{f}})-\mu(\widetilde{\bm{f}})\right|\geq d\right)\leq \P\left(\sup_{f\in\mathcal{F}}\left|\widetilde{\bm{\mu}}(f)-\mu(f)\right|\geq d\right),\label{eq:bounda}
		\end{equation}
	providing us with a \emph{uniform} bound for the first term on the RHS of \eqref{eq:lemprob}. We will now call upon Theorem~\ref{the:esterrorbound} (Appendix~\ref{ap:auxtheo}) to obtain an exponential upper bound on the RHS of \eqref{eq:bounda}. Using \eqref{eq:theombound} with the choice $x=d$ in \eqref{eq:bounda} yields:
	\begin{equation}
		\P\left(\left|\widetilde{\bm{\mu}}(\widetilde{\bm{f}})-\mu(\widetilde{\bm{f}})\right|\geq d\right)
		\leq \exp\left\{\frac{-(d-4\rho)^2N_{\sf max}}{2\alpha^2\beta^2} \right\},\label{eq:bounda1}
		\end{equation}
	for any positive $d$ such that $d> 4\rho$.

	Next, we examine the second term on the RHS of \eqref{eq:lemprob}. Using Lemma~\ref{lem:boundest} (Appendix~\ref{ap:aux}), with the choice $x=\Delta -{\sf R}^o$, we can derive the following uniform upper bound:
	\begin{equation}
		\P\left(R(\widetilde{\bm{f}})\geq \Delta\right)\leq\P\left(\sup_{f\in\mathcal{F}}
		\left|
		\widetilde{\bm{R}}(f) - R(f)
		\right|\geq\frac{\Delta-{\sf R}^{o}}{2}\right),\label{eq:boundc}
		\end{equation}
for any positive $d$ such that $\Delta=\log (1+e^{-d})> {\sf R}^{o}$. Such $d$ exists since, by assumption, $\ropt<\log 2$.
	
	Next, consider Eq.~\eqref{eq:theobound} of Theorem~\ref{the:esterrorbound} (Appendix~\ref{ap:auxtheo}), with the choice $x=(\Delta-{\sf R}^{o})/2$ and function $\phi(x) = \log(1+e^{x})$, which is a function with Lipschitz constant $L_\phi = 1$.
	Replacing \eqref{eq:theobound} with these choices into \eqref{eq:boundc} results in the bound:
	\begin{align}
		&\P\left(R(\widetilde{\bm{f}})\geq \Delta\right)
		\leq \exp\left\{\frac{-(\frac{\Delta-{\sf R}^{o}}{2}-4\rho)^2N_{\sf max}}{2\alpha^2\beta^2} \right\},\label{eq:boundc1}
		\end{align}
	for any $d$ such that $(\Delta -{\sf R}^{o})/2> 4\rho$.
Using \eqref{eq:bounda1} and \eqref{eq:boundc1} in \eqref{eq:lemprob} results in the following bound on the probability of consistent learning
	\begin{align}
	P_c&\geq 1-\exp\left\{\frac{-8(\frac{d}{4}-\rho)^2 N_{\sf max}}{\alpha^2\beta^2} \right\}\nonumber\\
	&-\exp\left\{\frac{-8(\frac{\Delta -{\sf R}^{o}}{8}-\rho)^2 N_{\sf max}}{\alpha^2\beta^2} \right\},\label{eq:boundpc}
	\end{align}
	for any $d$ satisfying
	\begin{equation}
\frac{d}{4} -\rho> 0~~\text{ and }~~
		\frac{\Delta -{\sf R}^{o}}{8}- \rho> 0,
		\end{equation}
	i.e., for any $d$ contained in the following interval:
	\begin{equation}
d\in (  4\rho, -\log(e^{8 \rho+{\sf R}^{o}}-1)).
		\end{equation}
For simplicity, we can rewrite \eqref{eq:boundpc} in the following manner:
\begin{align}
P_c&\geq 1-\exp\left\{\frac{-8 E^2_1(x) N_{\sf max}}{\alpha^2\beta^2} \right\}-\exp\left\{\frac{-8 E^2_2(x) N_{\sf max}}{\alpha^2\beta^2} \right\},
\end{align}
where we introduced the auxiliary functions:
\begin{align}
E_1(x)&\triangleq x-\rho,\\
E_2(x)&\triangleq 
\frac{\log(1+e^{-4 x})-\ropt}{8}-\rho,
\end{align}
and the free variable $d$ was replaced by
\begin{equation}
x\triangleq  \frac{d}{4}.
\end{equation}
We can now maximize the minimum exponent, i.e., the slowest decay rate, over the free parameter $x$. To this end, let us consider the value $x^{\star}$ that solves the equation: 
\begin{equation}
E_1(x^{\star})=E_2(x^{\star}),
\end{equation}
which corresponds to:
\begin{equation}
x^{\star}=\frac{\log(1+e^{-4 x^{\star}})-\ropt}{8}
\Leftrightarrow
e^{\ropt}\,e^{12x^{\star}}-e^{4x^{\star}}-1=0.
\end{equation}
Setting $e^{4x^{\star}}=y$, we have to solve the third-order equation:
\begin{equation}
e^{\ropt}\,y^3-y-1=0,
\end{equation}
whose unique real-valued solution $y^{\star}$ is available in closed form. Within the range $\ropt\in[0,\log 2]$, $y^{\star}$ is strictly greater than $1$, yielding:
\begin{align}
x^{\star}&=\frac 1 4 \log(y^{\star})\nonumber\\
&=\frac{1}{4}\log\left(\frac{
	2 \times 3^{\frac 1 3} + 2^{\frac 1 3} e^{-\ropt} [Z(\ropt)]^{\frac 2 3}}
{6^{\frac 2 3} [Z(\ropt)]^{\frac 1 3}}\right)\triangleq\mathscr{E}(\ropt),\label{eq:defeps}
\end{align}
where
\begin{equation}
Z(\ropt)=9 e^{2\ropt} + \sqrt{3e^{3\ropt}(-4+27 e^{\ropt})}.
\end{equation}
A good approximation for the function $\mathscr{E}(\ropt)$ is the linear fit --- see Fig.~\ref{fig:function}:
\begin{equation}
\mathscr{E}(\ropt)\approx 4\mathscr{E}(0)\left(1-\frac{\ropt}{\log 2}\right),\label{eq:approxE}
\end{equation}
where the maximum allowed complexity corresponding to a zero risk is:
\begin{equation}
4\mathscr{E}(0)=0.2812,
\end{equation}
which is related to the solution of the third-order equation:
\begin{equation}
	y^3 - y -1 =0.
\end{equation}
Fig.~\ref{fig:function} shows how accurate the linear approximation in \eqref{eq:approxE} is with respect to the exact expression for $\mathscr{E}(\ropt)$ in \eqref{eq:defeps} within the interval $\ropt\in[0, \log 2]$. 
\begin{figure}[t]
	\centering
	\includegraphics[width=3in]{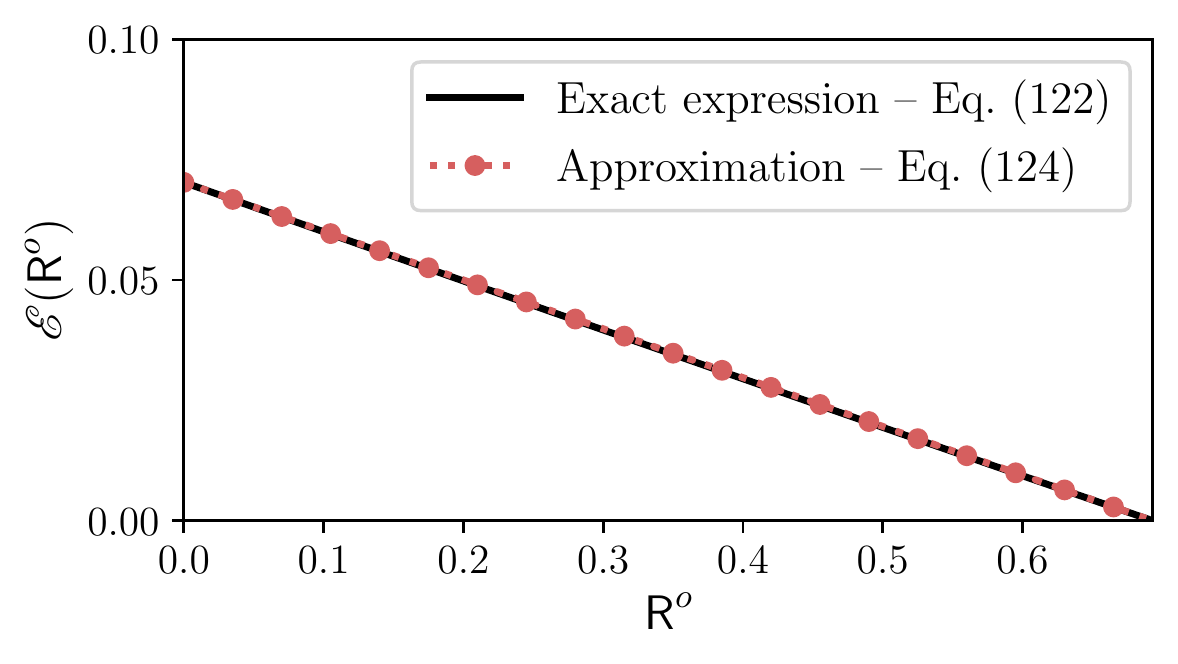}
	\caption{Comparison between the exact expression in \eqref{eq:defeps} and the approximation in \eqref{eq:approxE}.}\label{fig:function}
\end{figure}

Now, since $E_1(x)$ is an increasing function of $x$, while $E_2(x)$ is a decreasing function of $x$, we conclude that if we choose a value $x\neq x^{\star}$ the minimum exponent necessarily decreases. Accordingly, the minimum exponent is maximized at the value $x^{\star}=\mathscr{E}(\ropt)$. 

Finally, letting
\begin{equation}
\rho<\mathscr{E}(\ropt),
\end{equation}
we end up with the following bound:
\begin{equation}
P_c\geq 1- 2\exp\left\{
-\frac{8 N_{\sf max}}{\alpha^2\beta^2} \Big(\mathscr{E}(\ropt) - \rho\Big)^2
\right\},
\end{equation}
and the proof is complete.
\end{IEEEproof}

\section{Auxiliary Theorem}\label{ap:auxtheo}
To develop the forthcoming result, we consider a $L_\phi$-Lipschitz loss function $\phi:\mathbb{R}\mapsto \mathbb{R}_+$. The individual expected and empirical risks are written accordingly as:
\begin{align}
R_k(f_k)&=\E_{h_k,\gamma_k}\phi(-\bm{\gamma}_{k,n}f_k(\bm{h}_{k,n})),\\
\widetilde{\bm{R}}_k(f_k)&=\frac{1}{N_k}\sum_{n=1}^{N_k}\phi(-\bm{\gamma}_{k,n}f_k(\bm{h}_{k,n})),
\end{align}
where we removed the symbol $\sim$ from the top of random variables $\bm{\gamma}_{k,n}$ and $\bm{h}_{k,n}$ for simplicity of notation. Their network averages $R(f)$ and $\widetilde{\bm{R}}(f)$ are defined as shown in \eqref{eq:netrisk}.

\begin{theorem}[\textbf{Uniform Law of Large Numbers}]\label{the:esterrorbound}
Assume that the loss function $\phi:\mathbb{R}\mapsto\mathbb{R}_+$ is $L_\phi-$Lipschitz and that there exists $\beta>0$ such that $f_k(h)\leq \beta$ for every $h\in\mathcal{H}_k$, and $f_k\in\mathcal{F}_k$ and $k=1,2,\dots,K$. Then we have the following two results. First,
	\begin{align}
	&\P\left(\hspace{-1pt}\sup_{f\in\mathcal{F}}\left| \widetilde{\bm{R}}(f)-R(f)\right|\geq x\hspace{-1pt}\right)\hspace{-2pt}\leq \hspace{-1pt}\exp\left\{\hspace{-1pt}\frac{\hspace{-1pt}-N_{\sf max}\hspace{-1pt}\left(x-4L_\phi\rho\right)^2}{2\alpha^2L_\phi^2\beta^2} \hspace{-1pt}\right\},\label{eq:theobound}
	\end{align}
	for any $x> 4L_\phi\rho$. Second, 
	\begin{align}
	&\P\left(\hspace{-1pt}\sup_{f\in\mathcal{F}}\left| \widetilde{\bm{\mu}}(f)-\mu(f)\right|\geq x\hspace{-1pt}\right)\hspace{-2pt}\leq \hspace{-1pt}\exp\left\{\hspace{-1pt}\frac{-N_{\sf max}\left(x-4\rho\right)^2}{2\alpha^2\beta^2} \hspace{-1pt}\right\},\label{eq:theombound}
	\end{align}
	for any $x> 4\rho$, where $N_{\sf max}\triangleq\max_kN_k$, $\rho$ is the network Rademacher complexity defined in \eqref{eq:averad}, and $\alpha$ is defined as \eqref{eq:defalph}.
\end{theorem}
\begin{IEEEproof}
	In the proof, we use the known \emph{independent bounded differences inequality}, which is also known as \emph{McDiarmid's inequality}~\cite{mcdiarmid1989method}. The inequality is reproduced here without proof to facilitate its reference in the forthcoming results.
	\begin{mcdiar}\label{lem:McDiarmid}
		Let $\bm{x}$ represent a sequence of independent random variables $\bm{x}_n$, with $n=1,2,\dots,N$ and $\bm{x}_n\in\mathcal{X}_n$ for all $n$. Suppose that the function $g:\prod_{n=1}^N\mathcal{X}_n\mapsto \mathbb{R}$ satisfies for every $j=1,2,\dots,N$:
		\begin{equation}
		|g(x)-g(\check{x})|\leq c_j
		\end{equation}
		whenever the sequences $x$ and $\check{x}$ differ only in the $j-$th component. Then we have for $t>0$:
		\begin{align}
		&\P\Big(g(\bm{x})-\E g(\bm{x})\geq t\Big)\leq e^{-2t^2/\sum\limits_{j=1}^Nc_j^2},\label{eq:mcd1}\\
		&\P\Big(g(\bm{x})-\E g(\bm{x})\leq -t\Big)\leq e^{-2t^2/\sum\limits_{j=1}^Nc_j^2}.
		\end{align}\QED
	\end{mcdiar}	
	
	Before introducing the proof for Theorem~\ref{the:esterrorbound}, we provide a roadmap of the results used in the proof. Both \eqref{eq:theobound} and \eqref{eq:theombound} are proven using McDiarmid's inequality, followed by the auxiliary Lemma~\ref{lem:auxres} (found in Appendix~\ref{ap:aux}) and finally using Lemma~\ref{lem:lipsch} (also found in Appendix~\ref{ap:aux}). 
		
	We now develop the proof of \eqref{eq:theobound} and \eqref{eq:theombound} in Theorem~\ref{the:esterrorbound} separately as follows.

	\paragraph{Proof of \eqref{eq:theobound}}
	Consider that the sequence of samples $\bm{x}_n$ is replaced by a sequence of random pairs $(\bm{h}_n,\bm{\gamma}_n)$, with $n=1,2,\dots, N_{\sf max}$, where $N_{\sf max}\triangleq \max_k N_k$.
	The quantity $\bm{h}_n$ is a sequence collecting random variables (or vectors) $\bm{h}_{k,n}$ for $k=1,2,\dots,K$:
	\begin{align}
	\bm{h}_n&\triangleq \{\bm{h}_{1,n}, \bm{h}_{2,n},\dots, \bm{h}_{K,n}\},\label{eq:seqh}
	\end{align}
	and $\bm{\gamma}_n$ is a sequence of random variables $\bm{\gamma}_{k,n}$ for $k=1,2,\dots,K$:
	\begin{align}
	\bm{\gamma}_n&\triangleq \{\bm{\gamma}_{1,n}, \bm{\gamma}_{2,n},\dots, \bm{\gamma}_{K,n}\}.\label{eq:seqg}
	\end{align}
	The pairs $(\bm{h}_n,\bm{\gamma}_n)$ are independent and identically distributed over time, i.e., for all $n$.
	
Define the following auxiliary quantity:
	\begin{equation}
	\chi_k(f_k)\triangleq \E_{h_k,\gamma_k}\phi (-\bm{\gamma_}{k,n}f_k(\bm{h}_{k,n})),
	\end{equation}
where we recall that $\E_{h_k, \gamma_k}$ is the expectation computed according to the joint distribution of $\bm{h}_{k,n}$ and $\bm{\gamma}_{k,n}$.
	Our function of interest is the following:
\begin{align}
	g(h,\gamma) = &\sup_{f\in\mathcal{F}}\Bigg|\sum_{k=1}^{K}\pi_k\Big[ \chi_k(f_k)
	-\frac{1}{{N_k}}\sum_{n=1}^{N_k}\phi (-\gamma_{k,n}f_k(h_{k,n}))\Big] \Bigg|,\label{eq:ghy}
	\end{align}
where, to keep a concise notation, the arguments $h,\gamma$ indicate that the function $g(\cdot)$ depends on the collection of sequences $h_{n}$ (defined in \eqref{eq:seqh}) and $ \gamma_{n}$ (defined in \eqref{eq:seqg}) for $n=1,2,\dots,N_k$. The argument $f$ represents the ensemble of functions $\{f_k\}$, where $f_k\in\mathcal{F}_k$, and we define the global space of functions:
	\begin{equation}
	\mathcal{F}\triangleq \mathcal{F}_1\times  \mathcal{F}_2\times\dots\times \mathcal{F}_K.\label{eq:globalf}
	\end{equation}
	From the collections $h$ and $\gamma$, we can construct collections $\check{h}$ and $\check{\gamma}$, by replacing $h_{k,j}$ and $\gamma_{k,j}$ respectively with the distinct samples $\check{h}_{k,j}$ and $\check{\gamma}_{k,j}$ for all $k=1,2,\dots,K$. If $j>N_k$, the inner summand in \eqref{eq:ghy} is not altered, then, using the indicator function, we can write 
\begin{align}
	&g(\check{h},\check{\gamma})\nonumber\\&=  \sup_{f\in\mathcal{F}}\Bigg|\sum_{k=1}^{K}\pi_k\Big[ \chi_k(f_k) 
-\frac{1}{N_k}	\sum_{\substack{n=1\\n\neq j}}^{N_k}\phi (-\gamma_{k,n}f_k(h_{k,n})) \nonumber\\&\quad-\frac{\mathbbm{1}[j\leq N_k]}{N_k}\phi (-\check{\gamma}_{k,j}f_k(\check{h}_{k,j}))\nonumber\\&\quad-\frac{ \mathbbm{1}[j> N_k]}{N_k}\phi (-\gamma_{k,j}f_k(h_{k,j}))\Big]\Bigg|\nonumber\\
	&=\sup_{f\in\mathcal{F}}\Bigg|\sum_{k=1}^{K}\pi_k\Bigg[  \chi_k(f_k) 
	-\frac{1}{N_k}\sum_{n=1}^{N_k}\phi (-\gamma_{k,n}f_k(h_{k,n}))\nonumber\\&\quad +\frac{\mathbbm{1}[j\leq N_k]}{N_k}\Big(\phi (-\gamma_{k,j}f_k(h_{k,j}))-\phi(-\check{\gamma}_{k,j}f_k(\check{h}_{k,j}))\Big)\Bigg]\Bigg|,\label{eq:ghyp}
	\end{align}
where $\mathbbm{1}[E]$ is indicator function defined as: $\mathbbm{1}[E]=1$, if event $E$ takes place, $\mathbbm{1}[E]=0$ otherwise.
	It is convenient to introduce the following quantities:
	\begin{align}
u_k(f_k) &\triangleq  \chi_k(f_k) -\frac{1}{N_k}\sum_{n=1}^{N_k}\phi (-\gamma_{k,n}f_k(h_{k,n})),\label{eq:defx}\\
v_k(f_k)&\triangleq \frac{\mathbbm{1}[j\leq N_k]}{N_k}\Big[\phi (-\gamma_{k,j}f_k(h_{k,j}))-\phi(-\check{\gamma}_{k,j}f_k(\check{h}_{k,j}))\Big],\label{eq:defy}
	\end{align}
where the dependence of $u_k(\cdot)$ upon $(h,\gamma)$ and of $v_k(\cdot)$ upon $(\check{h},\check{\gamma})$ has been skipped for ease of notation.
In view of the definitions in \eqref{eq:defx} and \eqref{eq:defy}, we can rewrite \eqref{eq:ghy} and \eqref{eq:ghyp} as:
	\begin{align}
	g(h,\gamma)&=\sup_{f\in\mathcal{F}}\Bigg|\sum_{k=1}^{K}\pi_k u_k(f_k) \Bigg|\\
	g(\check{h},\check{\gamma})&=\sup_{f\in\mathcal{F}}\Bigg|\sum_{k=1}^{K}\pi_k u_k(f_k) +\sum_{k=1}^{K}\pi_kv_k(f_k)\Bigg|.
	\end{align}
Applying Lemma~\ref{lem:auxres} (Appendix~\ref{ap:aux}) with the choices $s_1=g(h,\gamma)$, $s_2=g(\check{h},\check{\gamma})$, and
	\begin{equation}
	S(f)=\sum_{k=1}^{K}\pi_k u_k(f_k),\quad T(f)=\sum_{k=1}^{K}\pi_k v_k(f_k),
	\end{equation}
	we obtain:
\begin{align}
|g(h,\gamma)-g(\check{h},\check{\gamma})|&\leq \sup_{f\in\mathcal{F}}\Big| \sum_{k=1}^{K}\pi_k v_k(f_k)\Big|\nonumber\\&\stackrel{\text{(a)}}{\leq} \sum_{k=1}^{K}\pi_k\sup_{f_{k}\in\mathcal{F}_k}\Big|  v_k(f_k)\Big|,\label{eq:theo3aux}
\end{align}
where (a) follows from the triangle inequality and the subadditive property of the supremum operator. Replacing \eqref{eq:defy} into \eqref{eq:theo3aux} yields
\begin{align}
&|g(h,\gamma)-g(\check{h},\check{\gamma})|\nonumber\\
&\leq\sum_{k=1}^K\pi_k \sup_{f_k\in\mathcal{F}_k}\Bigg|\frac{ \mathbbm{1}[j\leq N_k]}{N_k}\Big[\phi (-\gamma_{k,j}f_k(h_{k,j}))\nonumber\\
&\quad-\phi(-\check{\gamma}_{k,j}f_k(\check{h}_{k,j}))\Big]\Bigg|\nonumber\\
&\leq \sum_{k=1}^K\pi_k\hspace{-2pt} \sup_{f_k\in\mathcal{F}_k}\Bigg| \frac{1}{N_k}\Big[\phi (-\gamma_{k,j}f_k(h_{k,j}))-\phi(-\check{\gamma}_{k,j}f_k(\check{h}_{k,j}))\Big]\Bigg|\nonumber\\
&\stackrel{\text{(a)}}{\leq}L_\phi\sum_{k=1}^K\frac{\pi_k}{N_k} \sup_{f_k\in\mathcal{F}_k}\Bigg|\gamma_{k,j}f_k(h_{k,j})-\check{\gamma}_{k,j}f_k(\check{h}_{k,j})\Bigg|\nonumber\\
&\stackrel{\text{(b)}}{\leq}L_\phi\sum_{k=1}^K\frac{\pi_k}{N_k} \sup_{f_k\in\mathcal{F}_k}\left\{\Big|\gamma_{k,j}\Big|\,\Big|f_k(h_{k,j})\Big|+\Big|\check{\gamma}_{k,j}\Big|\,\Big|f_k(\check{h}_{k,j})\Big|\right\}\nonumber\\
&\stackrel{\text{(c)}}{\leq }2L_\phi \beta\sum_{k=1}^K\frac{\pi_k}{N_k}\stackrel{\text{(d)}}{=}\frac{2\alpha L_\phi \beta}{N_{\sf max}}.
\end{align}
	where (a) follows from the Lipschitz property of $\phi$, (b) follows from the triangle inequality, (c) follows from the boundedness assumption $f_k(h)\leq \beta$ and the fact that $|\gamma_{k,n}|=1$ for all $k$ and $i$. Finally, in (d) we used the definition in \eqref{eq:defalph}, namely,
\begin{equation}
 \alpha\triangleq \sum_{k=1}^K\pi_k \frac{N_{\sf max}}{N_k}.
\end{equation}
Applying McDiarmid's Inequality in \eqref{eq:mcd1} with $c_j=2\alpha L_\phi \beta/N_{\sf max}$, we obtain the following deviation bound:
	\begin{align}
	&\P\Big(\sup_{f\in\mathcal{F}}|R(f)-\widetilde{\bm{R}}(f)|-\E\sup_{f\in\mathcal{F}}|R(f)-\widetilde{\bm{R}}(f)|\geq t\Big)\nonumber\\
	&\leq { e^{-t^2 N_{\sf max}/(2\alpha^2L_\phi^2\beta^2)}},\label{eq:boundplus}
	\end{align}
holding for all $t>0$. To conclude the proof, we seek to upper bound the second term inside the probability operator in \eqref{eq:boundplus}. 	The result from Lemma~\ref{lem:lipsch} (Appendix \ref{ap:aux}) can be directly employed to conclude that:
	\begin{align}
	\E\sup_{f\in\mathcal{F}}|R(f)-\widetilde{\bm{R}}(f)|\leq 4 L_\phi\rho.\label{eq:auxbounds}
	\end{align}
In view of \eqref{eq:auxbounds}, we have that
	\begin{align}
	 &\sup_{f\in\mathcal{F}}|R(f)-\widetilde{\bm{R}}(f)|\geq t+4 L_\phi\rho
	 \nonumber\\ &\Rightarrow\sup_{f\in\mathcal{F}}|R(f)-\widetilde{\bm{R}}(f)|-\E\sup_{f\in\mathcal{F}}|R(f)-\widetilde{\bm{R}}(f)|)\geq t. \label{eq:auxbounds2}
	\end{align}
	From \eqref{eq:auxbounds2} and \eqref{eq:boundplus}, we can conclude that
	\begin{equation}
	\P\Big(\sup_{f\in\mathcal{F}}|R(f)-\widetilde{\bm{R}}(f)|\geq t+4L_\phi\rho\Big)\leq { e^{-t^2 N_{\sf max}/(2\alpha^2L_\phi^2\beta^2)}.}
	\end{equation}
	Defining $x=t+4L_\phi \rho$, and noting that $x> 4L_\phi \rho$ since $t> 0$, completes the proof of \eqref{eq:theobound}.
	
	\paragraph{Proof of \eqref{eq:theombound}} The proof for the uniform bound in \eqref{eq:theombound} follows similar arguments and will be thus presented in a concise manner. We start by using McDiarmid's Inequality with the following choice of function $g$:
	\begin{equation}
	g(h)=\sup_{f\in\mathcal{F}}\left|\sum_{k=1}^K\pi_k\Big[\nu_k(f_k)-\frac{1}{N_k}\sum_{n=1}^{N_k}f_k(h_{k,n})\Big]\right|,
	\end{equation}
where we define the auxiliary quantity: 
	\begin{equation}
	\nu_k(f_k)\triangleq \E_{h_k}f_k(\bm{h}_{k,n}).
	\end{equation}
	We follow similar steps as the ones used to prove \eqref{eq:theobound}, which results in the following bound:
	\begin{align}
	&\P\Big(\sup_{f\in\mathcal{F}}|\mu(f)-\widetilde{\bm{\mu}}(f)|-\E\sup_{f\in\mathcal{F}}|\mu(f)-\widetilde{\bm{\mu}}(f)|)\geq t\Big)\nonumber\\
	&\leq { e^{-t^2 N_{\sf max}/(2\alpha^2\beta^2)}}.\label{eq:mbound}
	\end{align}
	We use again Lemma~\ref{lem:lipsch} (Appendix~\ref{ap:aux}) to bound the second term inside the probability operation in \eqref{eq:mbound}. For this we take $L_\phi=1$ and we take $\bm{\gamma}_n=1$ as a deterministic variable, which allows us to derive the result:
	\begin{equation}
	\E\sup_{f\in\mathcal{F}}|\mu(f)-\widetilde{\bm{\mu}}(f)|\leq 4\rho.
	\end{equation}
	Replacing this bound in \eqref{eq:mbound}, defining $x=t+4\rho$, with $x> 4\rho$, yields the final result.
\end{IEEEproof}

\section{Proof of Theorem~\ref{cor:1}}\label{ap:cor}
Assuming $\rho_k\leq C_k/\sqrt{N_k}$, it follows that \eqref{eq:rhoboundk} holds, i.e.,
\begin{equation}
\rho\leq \frac{{\sf C}}{\sqrt{N_{\sf max}}}.\label{eq:rhoboundk2}
\end{equation}
For the bound in Theorem~\ref{the:consist} to hold, the Rademacher complexity must satisfy
\begin{equation}
\rho\leq \mathscr{E}(\ropt).\label{eq:condtheo2a}
\end{equation}
 In view of \eqref{eq:rhoboundk2}, \eqref{eq:condtheo2a} is met if we choose:
\begin{equation}
\frac{{\sf C}}{\sqrt{N_{\sf max}}}<\mathscr{E}(\ropt)\Leftarrow N_{\sf max}>\left(\frac{{\sf C}}{\mathscr{E}(\ropt)}\right)^2.\label{eq:firstboundonn}
\end{equation}
Next, for a desired minimum probability of consistent learning we should consider the bound found in \eqref{eq:worstcbound}. We have that:
\begin{align}
&P_c\geq 1-\varepsilon\nonumber\\
&\Leftrightarrow 2\exp\left\{
-\frac{8 N_{\sf max}}{\alpha^2 \beta^2} \left(\mathscr{E}(\ropt) - \frac{{\sf C}}{\sqrt{N_{\sf max}}}\right)^2\right\}
\leq \varepsilon\nonumber\\
&\Leftrightarrow N_{\sf max} \left(\mathscr{E}(\ropt) - \frac{{\sf C}}{\sqrt{N_{\sf max}}}\right)^2\geq \frac{\alpha^2\beta^2}{8}\log\left(\frac{2}{\varepsilon}\right).\label{eq:coraux}
\end{align}
We can develop the quadratic term in the LHS of \eqref{eq:coraux} as
\begin{align}
&N_{\sf max} \left(\mathscr{E}(\ropt) - \frac{{\sf C}}{\sqrt{N_{\sf max}}}\right)^2\nonumber\\&=
N_{\sf max} \,[\mathscr{E}(\ropt)]^2 - 2 \sqrt{N_{\sf max}} \,{\sf C} \,\mathscr{E}(\ropt) + {\sf C}^2.\label{eq:corineq}
\end{align}
Let 
\begin{equation}
z=\sqrt{N_{\sf max}}\,\mathscr{E}(\ropt),~~~
b={\sf C}^2-\frac{\alpha^2\beta^2}{8}\log\left(\frac{2}{\varepsilon}\right).
\end{equation}
To solve the inequality in \eqref{eq:coraux}, we must study the following quadratic equality: 
\begin{equation}
z^2 - 2 {\sf C} z  + b =0,
\end{equation}
whose positive solution is:
\begin{equation}
z={\sf C} + \sqrt{\frac{\alpha^2\beta^2}{8}\log\left(\frac{2}{\varepsilon}\right)}.
\end{equation}
Thus the inequality in \eqref{eq:coraux} is satisfied whenever:
\begin{equation}
\sqrt{N_{\sf max}}>\frac{1}{\mathscr{E}(\ropt)}\left({\sf C} + \sqrt{\frac{\alpha^2\beta^2}{8}\log\left(\frac{2}{\varepsilon}\right)}\right),
\end{equation}
or yet when:
\begin{equation}
N_{\sf max}>\left(\frac{{\sf C}}{\mathscr{E}(\ropt)}\right)^2
\left(1 + \frac{\alpha\beta}{2{\sf C}}\sqrt{\frac{1}{2}\log\left(\frac{2}{\varepsilon}\right)}\right)^2.\label{eq:corboundnmin}
\end{equation}
The final result of the theorem is established, since the bound in \eqref{eq:corboundnmin} is more stringent than \eqref{eq:firstboundonn}.

\section{Proof of Proposition~\ref{lem:radnn}}\label{ap:e}
Before introducing the proof, in order to establish the complexity of class $\mathcal{F}^{\sf NN}$, we will resort to a set of known inequalities involving the Rademacher complexity operator~\cite{bartlett2002rademacher,ledoux2013probability}, summarized in Property~\ref{prop:rad} (Appendix~\ref{ap:aux}). The proof follows an inductive argument similar to the one used in \cite{neyshabur2015norm}, where we establish an upper bound for the Rademacher complexity of the output of one layer with respect to the output of the previous layer, then this bound is iterated over the depth of the Multilayer Perceptron (MLP). 

We wish to analyze the complexity of the class of functions $\mathcal{F}^{\sf NN}$, which is defined in \eqref{eq:ffnn3} as the difference between  the outputs of the neural network $z_1$ and $z_2$, for an input vector $x\in\mathbb{R}^{n_0}$.
That is, function $f^{\sf NN}$ has the following form (as seen in \eqref{eq:ffnn3}):
	\begin{equation}
	f^{\sf NN}(x)=\log \frac{p(+1|x;f)}{p(-1|x;f)}=z_1-z_2,
	\end{equation} 
	where $z_1,z_2$ implement functions $g^{(L)}\in\mathcal{G}^{(L)}$ as defined in \eqref{eq:ffnn1} for $\ell=L$. We thus say that $f^{\sf NN}\in\mathcal{F}^{\sf NN}$, with
	\begin{equation}
	f^{\sf NN}(x)=g_1^{(L)}(x)- g_2^{(L)}(x),
	\end{equation}
	where $g_1^{(L)}, g_2^{(L)}\in\mathcal{G}^{(L)}$. 
	
	From items 1 and 2 in Property~\ref{prop:rad} (Appendix~\ref{ap:aux}). choosing $c=-1$, the empirical Rademacher complexity of $\mathcal{F}^{\sf NN}(x)$ will satisfy:
	\begin{align}
	\mathcal{R}\left(\mathcal{F}^{\sf NN}\left(x \right)\right)&\leq \mathcal{R}\left(\mathcal{G}^{(L)}\left(x\right)\right)+\mathcal{R}\left(\mathcal{G}^{(L)}\left(x\right)\right)\nonumber\\
	&= 2 \mathcal{R}\left(\mathcal{G}^{(L)}\left(x\right)\right).\label{eq:fnnbound}
	\end{align}
From \eqref{eq:ffnn1}, the Rademacher complexity of $\mathcal{G}^{(\ell)}(x)$ can be expressed as: 
\begin{align}
&\mathcal{R}\left(\mathcal{G}^{(\ell)}\left(x\right)\right)=\E_r\hspace{-8pt} \sup_{ w_j, g^{(\ell-1)}_j}\left|\frac{1}{N}\sum_{i=1}^N\bm{r}_i\sum_{j=1}^m w_j \sigma \left(g^{(\ell-1)}_j(x_i)\right) \right|,\label{eq:wthetarad}
\end{align}
where $\bm{r}_i$ are independent and identically distributed Rademacher random variables, with $\P(\bm{r}_i=+1)=\P(\bm{r}_i=-1)=1/2$.
The term on the RHS of \eqref{eq:wthetarad} can be rewritten as:
\begin{align}
&\E_r \sup_{w, g^{(\ell-1)}}\left|\frac{1}{N}\sum_{j=1}^m w_j\sum_{i=1}^N\bm{r}_i \sigma \left(g^{(\ell-1)}_j(x_i)\right) \right| \nonumber\\
&{ \stackrel{\text{(a)}}{\leq} \E_r \sup_{w, g^{(\ell-1)}}\|w\|_1\max_j\left|\frac{1}{N}\sum_{i=1}^N\bm{r}_i \sigma \left(g^{(\ell-1)}_j(x_i)\right) \right|\nonumber}\\
&\stackrel{\text{(b)}}{\leq} b\,\E_r\sup_{ g^{(\ell-1)}}\max_j\left|\frac{1}{N}\sum_{i=1}^N\bm{r}_i \sigma \left(g^{(\ell-1)}_j(x_i)\right) \right|\nonumber\\
&\stackrel{\text{(c)}}{=} b\,\mathcal{R}\left(\sigma \circ\mathcal{G}^{(\ell-1)}(x)\right) \stackrel{\text{(d)}}{\leq} 2bL_\sigma \mathcal{R}\left(\mathcal{G}^{(\ell-1)}(x)\right),
	\label{eq:recursionrad}
\end{align}
where (a) follows from triangle inequality and taking the maximum w.r.t. $j$, (b) follows from the assumption that $\|w\|_1\leq b$, (c) follows from the fact that $g_j^{(\ell -1)}\in\mathcal{G}^{(\ell-1)}$ for all $j=1,2,\dots,m$, (c)  and (d) follows from the contraction principle (item 3 in Property~\ref{prop:rad}, Appendix~\ref{ap:aux}) in association with the assumption that $\sigma$ is a Lipschitz function with constant $L_\sigma$.

Replacing \eqref{eq:recursionrad} into \eqref{eq:wthetarad}, we have the following recursion:
\begin{equation}
\mathcal{R}\left(\mathcal{G}^{(\ell)}\left(x\right)\right)\leq 2bL_{\sigma}\mathcal{R}\left(\mathcal{G}^{(\ell-1)}\left(x\right)\right).
\end{equation}
We can develop the recursion above across all layers up to $\ell$:
\begin{equation}
\mathcal{R}\left(\mathcal{G}^{(\ell)}(x)\right)\leq (2bL_{\sigma})^{\ell-1}\mathcal{R}\left(\mathcal{G}^{(1)}\left(x\right)\right).\label{eq:boundradL}
\end{equation}
It remains to bound the Rademacher complexity relative to $\mathcal{G}^{(1)}(x)$ of the first layer, whose functions have the form of $g^{(1)}_m$ defined in \eqref{eq:ffnn2}. For this purpose, we can directly use the result in Lemma 15 of \cite{neyshabur2015norm}, which bounds the Rademacher complexity of a linear separator with bounded $\ell_p$ norm. Applying this lemma with $p=1$, $\gamma = b$ and $\|x\|_\infty=\max_i |x_i|\leq c$, we have:
\begin{align}
\mathcal{R}\left(\mathcal{G}^{(1)}(x)\right)&=\E_r \sup_{ w_j}\left|\frac{1}{N}\sum_{i=1}^N\bm{r}_i\sum_{j=1}^d w_j x_{i,j}\right|\nonumber\\
&\leq \frac{2bc\sqrt{\log (2n_0)}}{\sqrt{N}}.\label{eq:f1bound}
\end{align}
Replacing \eqref{eq:boundradL} with $\ell=L$ and \eqref{eq:f1bound} into \eqref{eq:fnnbound} yields the final result.

\section{Auxiliary Results}\label{ap:aux}
We list three key properties of Rademacher complexity, which are used in some of our results. These properties are well known and therefore are reported here without proof, which can be found in~\cite{bartlett2002rademacher,ledoux2013probability}. 
	\begin{property}[\textbf{Inequalities involving Rademacher Complexity}~\cite{bartlett2002rademacher}] \label{prop:rad}
		Let $\mathcal{F},\mathcal{F}_1,\dots, \mathcal{F}_K$ be classes of real-valued functions, and $x$ a sequence of samples  $\{x_1,x_2,\dots,x_N\}$. The Rademacher complexity defined in \eqref{eq:defradem} satisfies the following properties:
		\begin{enumerate}
			\item Subadditivity: \begin{equation}
			\mathcal{R}\left(\textstyle\sum_{k=1}^{K}\mathcal{F}_{k}(x)\right)\leq\textstyle\sum_{k=1}^{K}\mathcal{R}(\mathcal{F}_{k}(x)),
			\end{equation}
			with $\mathcal{F}_1(x)+\mathcal{F}_2(x)\triangleq 
			\{[f_1(x_{1})+f_2(x_1), f_1(x_2)+f_2(x_2), \dots, f_1(x_N)+f_2(x_N)]: f_1\in\mathcal{F}_1, f_2\in\mathcal{F}_2\}.$
			\item Scaling: For every $c\in\mathbb{R}$, 
			\begin{equation}
			\mathcal{R}(c\,\mathcal{F}(x))\leq |c|\mathcal{R}(\mathcal{F}(x)),
			\end{equation}
			where $c\,\mathcal{F}(x)\triangleq\{[c f(x_1), c f(x_2),\dots,c f(x_N)]: f\in\mathcal{F}\}$.
			\item Contraction principle: Let $\phi:\mathbb{R}\mapsto \mathbb{R}_+$ be $Lipschitz$ with constant $L_\phi$ and $\phi(0)=0$. Then:
			\begin{equation}
			\mathcal{R}(\phi \circ \mathcal{F}(x))\leq 2L_\phi \mathcal{R}(\mathcal{F}(x)),
			\end{equation}
			with $\phi \circ \mathcal{F}(x)\triangleq\{[\phi( f(x_1)), \phi( f(x_2)), \dots, \phi( f(x_N))
			]: f \in\mathcal{F}\}$.
		\end{enumerate}\QED
\end{property}

The next three lemmas are important auxiliary results used in the proofs of Theorem~\ref{the:consist} and Theorem~\ref{the:esterrorbound}.

\begin{lemma}[\textbf{Upper Bound on the Estimation Error for the Empirical Risk}]\label{lem:boundest}
	From the definitions in \eqref{eq:targetrisk}--\eqref{eq:emprisk} and \eqref{eq:netrisk}, for $x>0$ we have that:
		\begin{equation}
		\P\left(R(\widetilde{\bm{f}})-{\sf R}^{o}\geq x \right)\leq\P\left(\sup_{f\in\mathcal{F}}
		\left|
		\widetilde{\bm{R}}(f) - R(f)
		\right|\geq \frac{x}{2}\right).
		\label{eq:overall}
		\end{equation}
\end{lemma}
\begin{IEEEproof}
	From~\eqref{eq:targetrisk} and~\eqref{eq:emprisk}, we can verify that for all $k=1,2,\dots,K$:
\begin{equation}
 \widetilde{\bm{R}}_k(\widetilde{\bm{f}}_k)\leq \widetilde{\bm{R}}(f_k), \text{ for all }f_k\in\mathcal{F}_k,
\end{equation}
which imply, from the definitions in \eqref{eq:netrisk}, that
\begin{align}
 \widetilde{\bm{R}}(\widetilde{\bm{f}})&\leq \widetilde{\bm{R}}(f), \text{ for all }f\in\mathcal{F},\label{eq:minRNineq}
\end{align}
where $\mathcal{F}$ is the global class of functions defined in \eqref{eq:globalf}. We can develop the expression of the estimation error to obtain the following uniform bound:
	\begin{align}
	R(\widetilde{\bm{f}})-\ropt&\stackrel{\text{(a)}}{=}R(\widetilde{\bm{f}})-\inf_{f\in\mathcal{F}}R(f)\nonumber\\
	&=R(\widetilde{\bm{f}})-\widetilde{\bm{R}}(\widetilde{\bm{f}})+\widetilde{\bm{R}}(\widetilde{\bm{f}})-\inf_{f\in\mathcal{F}}R(f)\nonumber\\
	&=R(\widetilde{\bm{f}})-\widetilde{\bm{R}}(\widetilde{\bm{f}})+\sup_{f\in\mathcal{F}}\left(\widetilde{\bm{R}}(\widetilde{\bm{f}})-R(f)\right)\nonumber\\
	&\stackrel{\text{(b)}}{\leq} R(\widetilde{\bm{f}})-\widetilde{\bm{R}}(\widetilde{\bm{f}})+\sup_{f\in\mathcal{F}}\left(\widetilde{\bm{R}}(f)-R(f) \right)\nonumber\\
	&\leq 2\sup_{f\in\mathcal{F}}\left|\widetilde{\bm{R}}(f)-R(f) \right|,\label{eq:boundrisk}
	\end{align}
	where (a) follows from the definition in \eqref{eq:targetrisk} and (b) follows from~\eqref{eq:minRNineq}.
		
		Finally, from \eqref{eq:boundrisk} and using the target risk notation in \eqref{eq:targetrisk}, we note that
		\begin{equation}
		R(\widetilde{\bm{f}})-{\sf R}^{o}\geq x\Rightarrow \sup_{f\in\mathcal{F}}\left|\widetilde{\bm{R}}(f)-R(f) \right|\geq x/2,
		\end{equation}
		for any $x>0$, thus concluding the proof.
\end{IEEEproof}

\begin{lemma}[\textbf{Uniform Upper Bound for Lipschitz Cost Functions}] \label{lem:lipsch}
		Assume that the pair of sequences $(\bm{h}_{n},\bm{\gamma}_n)$ is sampled independently from the same joint distribution for all $n=1,2,\dots,N_{\sf max}$. Let $f_k: \mathcal{H}_k\mapsto\mathbb{R}$ be a function belonging to class $\mathcal{F}_k$, and let $\phi:\mathbb{R}\mapsto\mathbb{R}_+$ be a $L_\phi-$Lipschitz function. Then it follows that
\begin{align}
		&\E_{h,\gamma} \sup_{f\in \mathcal{F}}\Bigg| \sum_{k=1}^{K}\pi_k\Big[ \chi_k(f_k)-\frac{1}{N_k}\sum_{n=1}^{N_k}\phi(-\bm{\gamma}_{k,n}f_k(\bm{h}_{k,n}))\Big]\Bigg|\nonumber\\&\leq 4L_\phi \rho,\label{eq:lemma2eq}
		\end{align}
		with 
	\begin{equation}	
		\chi_k(f_k)\triangleq \E_{h_k,\gamma_k} \phi(-\bm{\gamma}_{k,n}f_k(\bm{h}_{k,n})).
		\end{equation}
\end{lemma}
\begin{IEEEproof}
	Introduce the artificial pair $\bm{h}_n',\bm{\gamma}'_n$, sampled independently with the same joint distribution of $\bm{h}_n,\bm{\gamma}_n$. We develop the following symmetrization argument, inspired by the ones used in \cite{boucheron2005theory,bartlett2002rademacher}. 
	
	First, we use the triangle inequality and the subadditive property of the supremum operator:
	\begin{align}
	&\E_{h,\gamma} \sup_{f\in \mathcal{F}}\Bigg| \sum_{k=1}^{K}\pi_k\Big[ \chi_k(f_k)-\frac{1}{N_k}\sum_{n=1}^{N_k}\phi(-\bm{\gamma}_{k,n}f_k(\bm{h}_{k,n}))\Big]\Bigg|\nonumber\\
	&\leq\hspace{-1pt}\sum_{k=1}^{K}\pi_k\E_{h_k,\gamma_k}\hspace{-2pt}\sup_{f_k\in \mathcal{F}_k}\Bigg|\chi_k(f_k)\hspace{-1pt}-\hspace{-1pt}\frac{1}{N_k}\sum_{n=1}^{N_k}\phi(-\bm{\gamma}_{k,n}f_k(\bm{h}_{k,n}))\Bigg|,\label{eq:auxlemma2}
	\end{align}
where we recall that the argument $f$ represents the ensemble of functions $\{f_k\}$, with $f_k\in\mathcal{F}_k$, and $\mathcal{F}$ denotes the global space of functions defined in \eqref{eq:globalf}. 

We focus on the individual elements of the summation on the RHS of \eqref{eq:auxlemma2}, that is, on each term indexed by $k$. We drop subscript $k$ everywhere to simplify the notation.
	\begin{align}
	&\E_{h,\gamma}  \sup_{f\in \mathcal{F}}\Bigg|\chi(f)
	-\frac{1}{N}\sum_{n=1}^N\phi(-\bm{\gamma}_{n}f(\bm{h}_{n}))\Bigg|\nonumber\\
	&\stackrel{\text{(a)}}{=}\E_{h,\gamma} \sup_{f\in \mathcal{F}}\Bigg| \E_{h',\gamma'}\hspace{-1pt} \frac{1}{N}\hspace{-2pt}\sum_{n=1}^N\Big[\phi(-\bm{\gamma}'_{n}f(\bm{h}'_{n}))\hspace{-1pt}-\hspace{-2pt}\sum_{n=1}^N\phi(-\bm{\gamma}_{n}f(\bm{h}_{n}))\Big]\Bigg|\nonumber\\
	&\stackrel{\text{(b)}}{\leq} \E_{h,\gamma} \E_{h',\gamma'}\sup_{f\in \mathcal{F}}\left|  \frac{1}{N}\sum_{n=1}^N\Big[\phi(-\bm{\gamma}'_nf(\bm{h}'_n))-\phi(-\bm{\gamma}_nf(\bm{h}_n))\Big]\right|\nonumber\\
	&\stackrel{\text{(c)}}{=}\E_{h,\gamma}\E_{h',\gamma'}\E_r\sup_{f\in \mathcal{F}}\Bigg|  \frac{1}{N}\hspace{-2pt}\sum_{n=1}^N\hspace{-2pt}\bm{r}_n\Big[\phi(-\bm{\gamma}_n'f(\bm{h}'_n))\hspace{-2pt}-\hspace{-2pt}\phi(-\bm{\gamma}_nf(\bm{h}_n))\Big]\Bigg|\nonumber\\
	&\stackrel{\text{(d)}}{\leq}2\E_{h,\gamma}\E_r\sup_{f\in \mathcal{F}}\Bigg|  \frac{1}{N}\sum_{n=1}^N\bm{r}_n\phi(-\bm{\gamma}_nf(\bm{h}_n))\Bigg|\nonumber\\
	&\stackrel{\text{(e)}}{\leq}4 L_\phi\E_{h,\gamma}\E_r\sup_{f\in \mathcal{F}}\left|  \frac{1}{N}\sum_{n=1}^N\bm{r}_n\bm{\gamma}_nf(\bm{h}_n)\right|\nonumber\\
	&\stackrel{\text{(f)}}{\leq} 4 L_\phi\E_{h}\E_r\sup_{f\in \mathcal{F}}\left|  \frac{1}{N}\sum_{n=1}^N\bm{r}_nf(\bm{h}_n)\right|\nonumber\\&=4L_\phi \E_h\mathcal{R}\left(\mathcal{F}(\bm{h})\right).\label{eq:boundsym}
	\end{align}
	We explain now each of the steps (a)--(f) performed in \eqref{eq:boundsym}. In (a) we used the i.i.d. property of the artificial samples $(\bm{h}'_n,\bm{\gamma}'_n)$, (b) follows from the following two properties: $i)$ $|\E\bm{x}|\leq \E |\bm{x}|$;
		$ii)$ $\sup_{f\in\mathcal{F}}\E |\bm{y}(f)|\leq \E \sup_{f\in\mathcal{F}} |\bm{y}(f)|$.
	
	In (c) we introduced the i.i.d. Rademacher random variables, i.e., $\bm{r}_n\in\{-1,+1\}$ with uniform probability, which are independent of samples $(\bm{h}_n,\bm{\gamma}_n)$ and $(\bm{h}'_n,\bm{\gamma}'_n)$. Since  $(\bm{h}_n,\bm{\gamma}_n)$ and $(\bm{h}'_n,\bm{\gamma}_n')$ are identically distributed and independently sampled, exchanging $(\bm{h}_n,\bm{\gamma}_n)$ and $(\bm{h}'_n,\bm{\gamma}'_n)$ is immaterial and therefore we can safely introduce the Rademacher random variables $\bm{r}_n$ in the summation.
	
	In (d), we used the triangle inequality for the absolute value and the fact that $(\bm{h}_n,\bm{\gamma}_n)$ and $(\bm{h}'_n,\bm{\gamma}'_n)$ are identically distributed. In (e), we use the Lipschitz property of $\phi$ associated with the contraction principle of the Rademacher complexity (item 3 in Property~\ref{prop:rad}) to conclude that:
	\begin{align}
&\E_r\sup_{f\in \mathcal{F}}\Bigg|  \frac{1}{N}\sum_{n=1}^N\bm{r}_n\phi(-\bm{\gamma}_nf(\bm{h}_n))\Bigg|
\nonumber\\&\leq 2L_\phi\E_r\sup_{f\in \mathcal{F}}\left|  \frac{1}{N}\sum_{n=1}^N\bm{r}_n\bm{\gamma}_nf(\bm{h}_n)\right|.
	\end{align}
Step (f) follows from similar symmetrization arguments, considering that $\bm{\gamma}_n$ assumes values $\pm 1$ and that $\bm{r}_n$ and $-\bm{r}_n$ are equally distributed and independent form the samples and over $i$. Finally replacing \eqref{eq:boundsym} into \eqref{eq:auxlemma2} for each of the summands indexed by $k$, for all terms indexed by $k$, and recalling the definition of $\rho$ in \eqref{eq:averad}, we obtain \eqref{eq:lemma2eq}.
\end{IEEEproof}

\begin{lemma}[\textbf{Auxiliary Result for Bounded Differences}]\label{lem:auxres}
		Assume $S(f)$ and $T(f)$ are operators dependent on a real-valued function $f\in\mathcal{F}$, and consider the following quantities:
	\begin{align}
		s_1 = \sup_{f\in\mathcal{F}}\Big|S(f) \Big|,\quad
		s_2 = \sup_{f\in\mathcal{F}}\Big|S(f)+T(f)\Big|.
		\end{align}
		 Then, we have that:
		\begin{equation}
		|s_1-s_2|\leq \sup_{f\in\mathcal{F}}\Big|T(f)\Big|.\label{eq:auxres}
		\end{equation}
	\end{lemma}
	\begin{IEEEproof}
		The proof is split in two cases. 
		
		\emph{a)} Case $s_2\geq s_1$:
	\begin{align}
		s_2-s_1&= \sup_{f\in\mathcal{F}}\Big|S(f)+T(f)\Big|- \sup_{f\in\mathcal{F}}\Big|S(f) \Big|\nonumber\\
		&\leq \sup_{f\in\mathcal{F}}\Big|S(f)\Big|+\sup_{f\in\mathcal{F}}\Big|T(f)\Big|- \sup_{f\in\mathcal{F}}\Big|S(f)\Big|\nonumber\\
		&=\sup_{f\in\mathcal{F}}\Big|T(f)\Big|,\label{eq:supa}
		\end{align}
		where the inequality follows from the triangle inequality and the subadditive property of the supremum operator, i.e., $\sup_{f\in\mathcal{F}}[a(f)+b(f)]\leq \sup_{f\in\mathcal{F}}a(f)+ \sup_{f\in\mathcal{F}}b(f)$.
		
		\emph{b) Case} $s_2< s_1$:
	\begin{align}
		s_1-s_2
		&=  \sup_{f\in\mathcal{F}}\Big|S(f) \Big|-\sup_{f\in\mathcal{F}}\Big|S(f)+T(f)\Big|\nonumber\\
		&=\sup_{f\in \mathcal{F}}\left(\Big|S(f) \Big|-s_2\right)\nonumber\\&
		\stackrel{\text{(a)}}{\leq}\sup_{f\in \mathcal{F}}\left(\Big|S(f)\Big|-\Big|S(f)+T(f)\Big|\right)\nonumber\\
		&\leq\sup_{f\in \mathcal{F}}\left|\,\Big|S(f)\Big|-\Big|S(f)+T(f)\Big|\,\right|\nonumber\\
		&\stackrel{\text{(b)}}{\leq}\sup_{f\in \mathcal{F}}\Big|S(f)-S(f)+T(f)\Big|=\sup_{f\in \mathcal{F}}\Big|T(f)\Big|,\label{eq:supb}
		\end{align}
		where (a) follows from the definition of $s_2$, and (b) from the reverse triangle inequality, i.e., $|a-b|\geq |\, |a| - |b| \,|$.
		
		Using \eqref{eq:supa} and \eqref{eq:supb}, we obtain the desired result in \eqref{eq:auxres}.
\end{IEEEproof}
\ifCLASSOPTIONcaptionsoff
  \newpage
\fi

\bibliographystyle{IEEEtran}


\begin{IEEEbiographynophoto}{Virginia Bordignon}
	received engineering degrees from Ecole Centrale de Lyon, France, and the Federal University of Rio Grande do Sul (UFRGS), Brazil, the M.S. degree in electrical engineering from UFRGS, and the Ph.D. degree in electrical engineering from \'Ecole Polytechnique F\'ed\'erale de Lausanne (EPFL), Switzerland. She is currently a Postdoctoral Researcher with the Adaptive Systems Laboratory, EPFL, Switzerland. Her research interests include statistical inference, distributed learning and information processing over networks.
\end{IEEEbiographynophoto}

\begin{IEEEbiographynophoto}{Stefan Vlaski} received the B.Sc. degree in Electrical Engineering and Information Technology from Technical University Darmstadt, Darmstadt, Germany, in 2013, and the M.S. degree in Electrical Engineering and Ph.D. degree in Electrical and Computer Engineering from the University of California, Los Angeles, CA, USA, in 2014 and 2019, respectively. He is currently a Lecturer with Imperial College, London, U.K., where he conducts research at the intersection of machine learning, network science, and optimization. From 2019 to 2021, he was a Postdoctoral Researcher with the Adaptive Systems Laboratory, \'Ecole Polytechnique F\'ed\'erale de Lausanne, Lausanne, Switzerland.
\end{IEEEbiographynophoto}

\begin{IEEEbiographynophoto}{Vincenzo Matta} is currently a Full Professor in telecommunications at the Department of Information and Electrical Engineering and Applied Mathematics (DIEM), University of Salerno, Italy. He is the author of more than 130 articles published on international journals and proceedings of international conferences. His research interests include adaptation and learning over networks, social learning, statistical inference on graphs, and security in communication networks. He serves as an Associate Editor for the IEEE Open Journal of Signal Processing. He served as an Associate Editor for the IEEE Transactions on Aerospace and Electronic Systems, the IEEE Signal Processing Letters, and the IEEE Transactions on Signal and Information Processing over Networks, and a Senior Area Editor for the IEEE Signal Processing Letters. He is a member of the Sensor Array and Multichannel Technical Committee of the Signal Processing Society (SPS), and served as an IEEE SPS Representative to the IEEE Transactions on Signal and Information Processing over Networks.
\end{IEEEbiographynophoto}

\begin{IEEEbiographynophoto}{Ali H. Sayed} is Dean of Engineering at EPFL, Switzerland, where he also leads the Adaptive Systems Laboratory. He has served before as distinguished professor and chairman of electrical engineering at UCLA. He is a member of the US National Academy of Engineering (NAE) and The World Academy of Sciences (TWAS). He served as President of the IEEE Signal Processing Society in 2018 and 2019. His work has been recognized with several awards including more recently the 2022 IEEE Fourier Award, the 2020 IEEE Norbert Wiener Society Award, and several Best Paper Awards. He is a Fellow of IEEE, EURASIP, and AAAS.
\end{IEEEbiographynophoto}

\end{document}